%% file: main.tex
\documentclass{article} 
\usepackage{iclr2027_conference,times}

\input{math_commands.tex}

\usepackage{hyperref}
\usepackage{url}
\usepackage{wrapfig}
\usepackage{booktabs}
\usepackage{tabularx}

\title{How VLMs Plan What's Next}

\iclrfinalcopy

\author{Rohit Gandikota\thanks{Correspondence to \href{mailto:gandikota.ro@northeastern.edu}{gandikota.ro@northeastern.edu}. Code and data are available at \href{https://plan.baulab.info}{plan.baulab.info}.} \quad Arnab Sen Sharma \quad Adam Belfki \quad  David Bau\\
Northeastern University \\
}
\input{task_commands}
\input{color_task_commands}
\begin{document}

\maketitle
\lhead{Preprint. Under review.}

\input{secs/0_abstract}
\input{secs/1_introduction}
\input{secs/2_related}
\input{secs/3_setup}
\input{secs/4_binding}
\input{secs/5_progress}
\input{secs/6_readers}
\input{secs/7_conclusion}

\input{secs/misc}

\bibliography{main}
\bibliographystyle{iclr2027_conference}

\newpage
\input{secs/appendix}

\end{document}

%% file: math_commands.tex
\usepackage{amsmath,amsfonts,bm}

\def\eqref#1{equation~\ref{#1}}

\def\1{\bm{1}}

\DeclareMathAlphabet{\mathsfit}{\encodingdefault}{\sfdefault}{m}{sl}
\SetMathAlphabet{\mathsfit}{bold}{\encodingdefault}{\sfdefault}{bx}{n}



%% file: task_commands.tex
\usepackage{tikz}
\usetikzlibrary{shapes.geometric}
\tikzset{panel mark/.style={draw,line width=0.4pt,inner sep=0pt,font=\sffamily\bfseries\tiny}}

\newcommand{\strippanels}[1]{%
  \foreach[count=\i] \s/\l in {#1}{%
    \draw[line width=0.3pt,gray] (\i*1.6em-0.8em,-0.8em) rectangle ++(1.6em,1.6em);
    \begin{scope}[shift={(\i*1.6em,0)}]\csname strip\s\endcsname{\l}\end{scope}}}
\newcommand{\stripscale}{0.85}
\newcommand{\strip}[1]{\tikz[baseline=-0.6ex,scale=\stripscale,every node/.append style={transform shape}]{\strippanels{#1}}}
\newcommand{\inp}[2]{\tikz[baseline=-0.6ex,scale=\stripscale,every node/.append style={transform shape}]{%
  \draw[line width=0.5pt] (0.55em,0.8em) .. controls (0.25em,0.4em) and (0.25em,-0.4em) .. (0.55em,-0.8em);
  \strippanels{#1}
  \node[anchor=base west,inner sep=0pt,font=\sffamily,transform shape=false] (q) at ([xshift=0.15em]current bounding box.east |- 0,-0.6ex) {,\,#2};
  \draw[line width=0.5pt] ([xshift=0.15em]q.east |- 0,0.8em) .. controls +(0.3em,-0.4em) and +(0.3em,0.4em) .. ([xshift=0.15em]q.east |- 0,-0.8em);}}
\newcommand{\pmv}[1]{{\scriptsize\,$\pm$\,#1}}

%% file: color_task_commands.tex
\usepackage{CJKutf8}
\tikzset{colour mark/.style={draw,line width=0.4pt,inner sep=0pt}}

\newcommand{\cstrippanels}[1]{%
  \foreach[count=\i] \s/\c in {#1}{%
    \draw[line width=0.3pt,gray] (\i*1.6em-0.8em,-0.8em) rectangle ++(1.6em,1.6em);
    \begin{scope}[shift={(\i*1.6em,0)}]\csname cstrip\s\endcsname{\c}\end{scope}}}
\newcommand{\cstrip}[1]{\begingroup\let\strippanels\cstrippanels\strip{#1}\endgroup}
\NewDocumentCommand{\cinp}{m m o}{\begingroup\let\strippanels\cstrippanels
  \IfNoValueTF{#3}{\inp{#1}{#2}}{\inp{#1}{#2\textsubscript{#3}}}\endgroup}

%% file: secs/0_abstract.tex
\begin{abstract}
When a vision-language model describes an image, it names one object at a time and it rarely repeats itself, yet how the model plans this order is far from obvious. 
We characterize the mechanism a VLM uses to \emph{decide where to look} before it looks. 
The mechanism has four main parts. First, \emph{binding heads} match the relevant text tokens against the image and resolve them into pointers to the matched image region. Second, the same binding heads bind each written answer to the region it came from, and the model turns it into a \emph{progress} signal: a cursor, the position in the image up to which the list has been written. Third, a \emph{select} step chooses the nearest unwritten region after the cursor, or stops. Finally, \emph{gaze heads} execute the chosen pointer and return a concept of what is drawn there, which \emph{language heads} and the late MLPs turn into a word in the language of the question. We validate each part of this mechanism through causal patching in Qwen3-VL models on a synthetic visual retrieval task, and test our mechanisms against alternative hypotheses. We further show that the mechanism holds across input formats, natural photographs, and languages, and that its binding, relay, and gaze heads appear at every model size from 2B to 32B. This work allows us to investigate how a VLM turns a question and an image into a sequence of answers, layer by layer and token by token. More broadly, it opens a way to study how models plan over complex multi-step tasks with long rollouts. Our project is available at \href{https://plan.baulab.info/}{\textcolor[rgb]{0.21,0.49,0.74}{plan.baulab.info}}.
\end{abstract}

%% file: secs/1_introduction.tex
\section{Introduction}
\label{sec:introduction}

How does a vision-language model decide which part of an image to describe next? Unlike pure language models, VLMs take two modalities of input, with an inherent mismatch between them. Nothing in a two-dimensional image imposes a one-dimensional order on its contents. Somewhere inside the model, a plan must turn the image into a sequence of words. When a VLM describes an image, it names several objects one after another and rarely names any of them twice, so this plan must both choose what comes next and keep track of what has been said. In this work, we characterize the mechanism inside a VLM that makes this plan, in a controlled task where each step can be tested causally.

Prior work uncovered parts of this mechanism through the lens of binding. In language models, an entity can be retrieved in two ways. First, through an abstract positional code: each entity in a prompt is tagged with a positional code, and the model later follows that code to retrieve the entity~\citep{feng2024binding,prakash2026lookback}. Second, through content: filter heads~\citep{sensharma2026filter} attend to the items of a list that match a rule carried in their query vectors. Similar mechanisms were discovered in VLMs: when they process the words of a prompt, they build spatial indexes to the matching image regions~\citep{assouel2026visual}, and a small set of Gaze Heads attends to the region being described, so that redirecting them redirects the description~\citep{gandikota2026gaze}. Together, these works explain how a model finds a region in the image that answers a question. 

It is still unclear how the model solves a task with multiple answers corresponding to several image regions. This requires keeping track of the regions it has already described. In an autoregressive transformer, this progress cannot be kept as a running state that is updated after each answer. What one token computes at layer $l$ reaches later tokens only at layer $l+1$ or above, so a state handed from answer to answer would soon run out of layers. The model must instead recalculate its progress at every token, using the previously generated tokens. Our work discovers how progress is computed and shows how the prior mechanisms play a part in solving the task.

An open-ended description is difficult to study causally since it is hard to control which objects the model describes or in what order. To find the plan, we instead design a simple synthetic retrieval task (Fig.~\ref{fig:overview}a) that lets us control each part of the mechanism on its own and see what changes. We then test how much of the mechanism carries over to other layouts, other prompt wordings, and list questions on natural photographs (App.~\ref{app:other-images}). Each image is a strip of panels, where each panel holds a shape and a letter. We query the model to output the letter inside a named shape (``\texttt{What is the letter inside the star?}''). We then design causal patching experiments with paired source and target runs, where patching activations that causally carry different information produces different answers in the target run. The target output therefore tells us exactly what the patched state carried.

We find a structured planning mechanism inside VLMs that also connects several of the prior mechanisms (Fig.~\ref{fig:overview}b). First, at the shape token in the user prompt, a few \emph{binding heads} match the requested shape against the image and resolve a pointer to every matching region. The model largely reuses the same heads to track progress: when it writes an answer, the binding heads bind that answer to the image region it came from, and the comma token turns this region into a cursor, the place in the image up to which the list has been written. Next, \emph{relay heads} carry the pointers to the token where the answer is being generated, and the model \emph{selects} the nearest unwritten region after the cursor, or stops. Then \emph{gaze heads} execute the selected pointer, following it back to its region, and bring the visual information. Finally, a few \emph{language heads} in the late layers identify the language of the prompt and inform the last MLP layers, which translate the visual information into the language of the user prompt.

In addition to finding new mechanisms, our work connects several previously discovered mechanisms, found in different models and tasks, into a single plan. We show that the same components solve more complex tasks than the ones they were found on. We also find that the model rebuilds its plan at every token from the tokens it has already generated: we can read the plan and edit it. This view may help explain why long descriptions repeat or skip objects, and it opens a way to study how models track progress in long, multi-step tasks.

\input{figures/overview}

%% file: figures/overview.tex
\begin{figure}[t]
\centering
\includegraphics[width=0.92\textwidth]{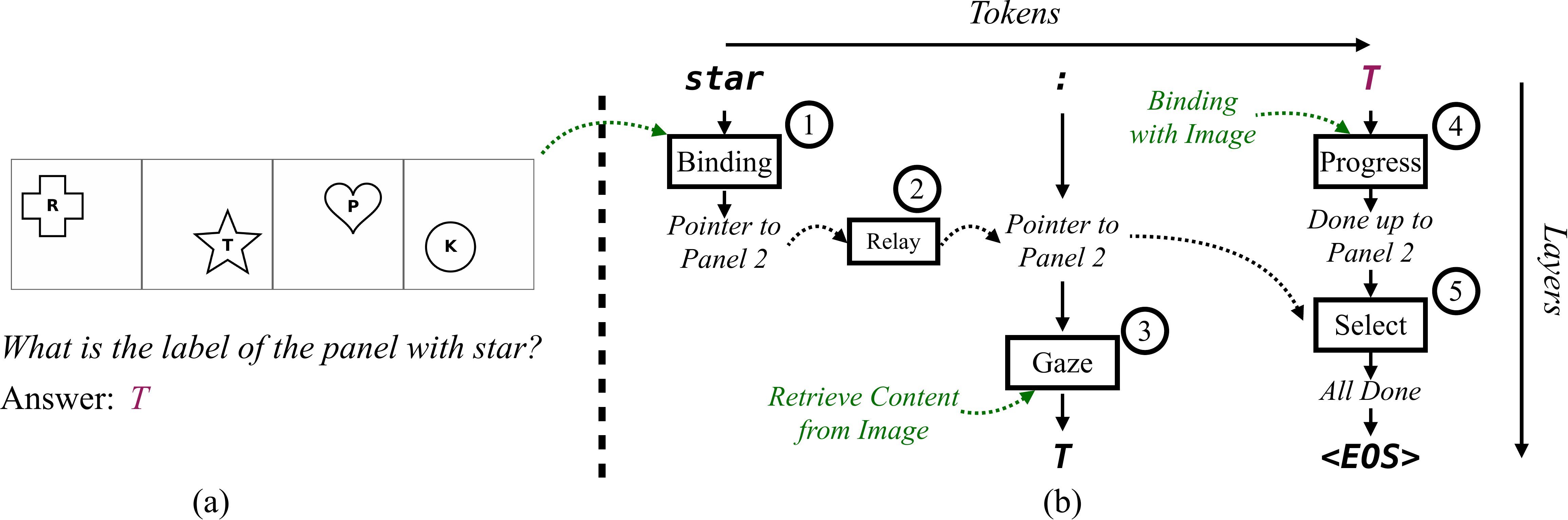}
\caption{(a)~Task setup: We provide an image strip with each panel containing a shape and a letter. We ask the model to retrieve the label of the panel that contains the queried shape. (b)~The mechanism, with tokens across and layers down. The model solves the task early, at the shape token \texttt{star}, well before it answers: (1)~binding heads match the request against the image and resolve a pointer to the matching panel. (2)~Relay heads carry that pointer to the ``\texttt{:}'' token, where the answer has to be produced, and (3)~gaze heads execute it to retrieve the letter from that panel. (4)~Once \texttt{T} is written, binding heads bind it to its panel, and the model calculates its progress and (5)~selects what answer to mention next (by repeating steps 3-5) or to stop if none is left.}
\label{fig:overview}
\end{figure}

%% file: secs/2_related.tex
\section{Related Work}
\label{sec:related}

\paragraph{Binding and pointers:} When a prompt introduces several entities, language models tag each one with an abstract code tied to the order of mention, and a later attention step follows that code back to the entity it needs \citep{feng2024binding,dai2024binding,prakash2026lookback, prakash2024finetuning}. LLMs also contain filter heads that carry, in their queries, the rule that selects the matching items \citep{sensharma2026filter}. VLMs often fail to keep the features of several objects apart \citep{campbell2024binding}. Image information reaches the answer either through the question tokens \citep{zhang2025crossmodal,basu2024information} or directly from the image tokens \citep{neo2025visual,kaduri2025whats, gandikota2026gaze}, and the route depends on the task \citep{salazar2026pathways}. On a synthetic task, several works have found that attention heads retrieve a pointer through binding, which is later used by another set of heads to retrieve the information~\citep{saravanan2025binding,kang2026linear,wang2026knowing,cui2026dual,assouel2026visual}. \citet{cui2026dual} show that VLMs bind objects to their spatial relations through two routes: content-independent spatial codes in the middle layers of the language backbone, and a stronger spatial signal from the vision encoder that the backbone uses directly. We find similar binding mechanisms, and extend this study to how a VLM keeps track of its answers and how it chooses the next object when the question matches several regions.

\paragraph{Keeping track of progress:} Language models plan beyond the next token \citep{pal2023future,lindsey2025biology,hanna2026latent} and keep a running state of their own output: a character count \citep{gurnee2025manifolds}, a count of list items at the separators \citep{hasani2026counting}, and the suppression of answers already given \citep{yan2025promote,merullo2024talking}. VLMs seem to work serially only through the text they generate \citep{budny2025serial} and judge when to stop by comparing that text with the image \citep{yue2024less}. They list the objects of a grid in raster order \citep{assouel2026visual}. A VLM trained to point keeps an internal record of the objects it has counted \citep{haputhanthri2026pointing}. We find such a record in a general VLM that writes only letters. It is neither a count nor a set of the letters written, but a place in the image, rebuilt from the written answers, from which the next region is chosen.

\paragraph{Image reading heads:} Several studies find small sets of heads that attend from the text to the relevant part of the image. Some select heads by where they look \citep{kang2025localization, gandikota2026gaze}, while others rank heads by their causal effect \citep{golovanevsky2025notice,schaumloeffel2026localization,liu2026grounding}. OCR heads read the text written in an image~\citep{baek2025ocr}, and \citet{feucht2026ocr} find that heads needed for OCR also read out the meaning of image regions without text, such as ``feathers'' at a bird's wing. We find that heads chosen by causal effect read the pointer better than heads chosen by attention and are more causally sufficient.

\paragraph{Concept and output language:} Multilingual language models hold concepts in their middle layers in a form close to their dominant language, often English \citep{wendler2024llamas,schut2025english,wu2025semantichub}. Concept and output language can be patched apart \citep{dumas2025tongue} with a small set of attention heads \citep{zhang2025translation, lasnier2026translation,liu2026focusing}. We find that when the concept comes from an image, the heads that read the image return a concept that does not depend on the question's language, and separate heads and the last MLPs translate it into that language.

%% file: secs/3_setup.tex
\section{Setup and Notation}
\label{sec:setup}
\paragraph{Model:}
We primarily study the language backbone of the Qwen3-VL-8B-Instruct model \citep{bai2025qwen3vl}. Its language backbone has 36 layers with 32 attention heads per layer. We do not intervene on the vision encoder; the image inputs are encoded once and cached. We test other model sizes and model families in Appendices~\ref{app:model_size} and~\ref{app:model-families}.

\paragraph{Dataset:} We build a synthetic visual retrieval task where each image is a strip of blank panels. Each panel consists of a shape and a letter label, \strip{plus/R, star/T, heart/P, circle/K}, as shown in Figure~\ref{fig:overview}a. The user input contains the image followed by the prompt ``\texttt{Answer with the letter only. What is the letter inside the star?}''. We prefill the assistant turn with ``\texttt{Answer:}'' to force the next generated token to be the answer token, \texttt{T}. In this setup, four types of tokens are relevant to our experiments. The \emph{shape token} is the word that names the shape (e.g. \texttt{star}). The \emph{answer token} is the letter the model generates. The \emph{colon token} and \emph{comma token} are where the LLM generates the answer token. Throughout the paper, we use the notation \inp{plus/R, star/T, heart/P, circle/K}{star} to define the input image and prompt. The shapes are placed randomly on a panel with the letters center-aligned inside them. The question asks for a letter based on a shape instead of a panel number. We do this to separate the binding signal from knowledge of panel numbers. The 8B model answers our synthetic task correctly in 98.9\% of questions (Appendix~\ref{app:task-native}).

\paragraph{Intervention:} Every experiment pairs two runs that differ in both the image and the question:

\begin{center}
\begin{tabular}{@{}r@{\ }l@{\quad}l@{\quad}l@{}}
Source:    & \inp{star/G, circle/C, plus/Z, heart/D}{heart} & prefilled ``\texttt{Answer:}'' & $\to$ \texttt{D} \\
Target: & \inp{plus/R, star/T, heart/P, circle/K}{star}  & prefilled ``\texttt{Answer:}'' & $\to$ \texttt{T} \\
\end{tabular}
\end{center}

To answer ``\texttt{What is the letter inside the star?}'', prior work~\citep{assouel2026visual} suggests that the model must do two things. First, it must \emph{resolve} a pointer by matching the requested shape against the layout of the image to decide which panel holds the answer. Then, it must \emph{execute} the pointer by reading the letter in that panel. To separate these steps, we record the internal activations of the source run and copy them into the target run, one component at a time, a method known as activation patching~\citep{meng2022rome}. We build each pair such that every kind of information a patch could move leads the target to a different answer token:
\begin{itemize}
\item \texttt{T}, no effect: the patch changed nothing.
\item \texttt{D}, shallow copy: the patch copied the source's answer.
\item \texttt{K}, resolved pointer: the patch moved the pointer resolved in the source.
\item \texttt{P}, request: the patch moved the source's request, which was resolved on the target's layout.
\item \texttt{R}, layout: the patch moved the source's layout, on which the target's request was resolved.
\end{itemize} 
In the \texttt{K}, \texttt{P}, and \texttt{R} cases the resolved pointer is still executed on the target's image.

Every patch replaces one activation of the target run with that of the source run, at one token and one layer, and leaves everything else in the target run unchanged. We patch at three activation levels: the whole residual vector at a layer, the output of an MLP, and the output of an attention head. For attention heads, we also patch the queries, the keys, and the values separately: a query patch changes what the head looks for, a key patch lets it read the source's image keys, and a value patch changes what it brings back from the tokens it attends to. Instead of aiming to identify a single head, we search for the top-$K$ heads. To find them, we start by patching one head at a time, for all 1,152 heads, on 200 discovery pairs, and rank the heads by how much the patch raises the logit of the expected letter over the target's own letter. We then patch the top $K$ heads together for $K$ in 1, 2, 4, \ldots, 128 on 100 validation pairs to choose $K$, and report every result on 500 held-out pairs that played no part in either choice, against the same number of random heads from the same layers (Appendix~\ref{app:task-implementation}). App.~\ref{app:overview} lists every experiment of the paper with the token it is applied at, the activation it replaces, and the answer it predicts.

\paragraph{Task types:} In our dataset, some inputs have one answer and some have a list. In the list task the asked shape sits in multiple panels. We ask ``\texttt{What is the letter inside each \{shape\}?}'', and the model lists the letters from left to right, for example \inp{heart/E, diamond/M, heart/U, diamond/V, heart/P, heart/A}{heart} $\to$ \texttt{E, U, P, A}. We use the list task to study how the model tracks its progress through the list and selects the next panel. 

%% file: secs/4_binding.tex
\section{Binding: VLMs Resolve Pointers to Solve a Task}
\label{sec:binding}
Similar to language models~\citep{prakash2024finetuning,feng2024binding,dai2024binding,prakash2026lookback}, we find that vision-language models first resolve a pointer~\citep{assouel2026visual,cui2026dual} to the image region containing the answer and later execute that pointer to read the answer. To show this, we use the source and target pair setup from Section~\ref{sec:setup}. We start by patching the entire residual vector at the colon token from the source into the target, one layer at a time, and track the logits of the five candidate letters from Section~\ref{sec:setup} in the patched target run (Figure~\ref{fig:binding}a). Up to layer 20 the target's answer is unchanged in 99.6\% of cases or more. From layers 23-29 the target answers with \texttt{K}: the patch transfers the resolved pointer from the source run, but the execution is still done in the target run. From layer 30 on, the patch is a shallow copy of the source's answer token, \texttt{D}, in 95\% of cases or more. The request \texttt{P} and the layout \texttt{R} appear in at most 6.2\% of cases at any layer. This shows that the colon token at layers 23-29 holds a resolved pointer. The model then executes this pointer to retrieve the letter from the image, and writes the final answer into the residual stream after these layers. In this section, we investigate where and how this pointer is resolved.

\input{figures/binding}

\subsection{Pointers Are Resolved as Early as the User Prompt Tokens}
\label{subsec:pointer}
To localize the token where the pointer is resolved, we patch the residual vector at every token in the user prompt and the assistant prompt, at every layer, and track the logit of \texttt{K} in Figure~\ref{fig:binding}b. In addition to the colon token ``\texttt{:}'' causally carrying the pointer at layers 23-29, we find that the shape token also carries the pointer, although weakly, at the earlier layers 16-20 (App.~\ref{app:one-token-map}). Since no other earlier token causally carries the pointer, we investigate whether the shape token is where the model resolves the pointer.

We start by patching the output of one attention head at a time at the shape token. We rank the heads by how much each patch raises the logit of \texttt{K}. We find that the top 16 attention heads, at layers 15-21, causally resolve the pointer (Figure~\ref{fig:binding}c). Patching the outputs of these 16 heads moves the target to \texttt{K} in 85.9\% of cases, while the same number of random heads from the same layers does so in at most 0.1\%. When the task is changed from shape-based querying to color-based querying (App.~\ref{app:other-bindings}), the same heads resolve the pointer. We call them \emph{binding heads}. No single MLP at the shape token carries the pointer signal (App.~\ref{app:one-mlp}).

We further dissect how attention heads are used by the model to resolve the pointer by patching the queries and the keys of all the attention heads at the shape token separately (Appendix Figure~\ref{fig:appendix_binding_qk}). We then rank the heads based on the logit differences across the five candidate letters from Section~\ref{sec:setup}. The queried request is carried through the queries of the attention heads: patching the queries of the top 8 query heads makes the target answer with \texttt{P}, the letter inside its own heart, in 81.0\% of cases. The layout is carried through the image keys being read at the shape token: letting the top 8 key heads read the source's image keys makes the target answer with \texttt{R} in 83.1\% of cases, where the pointer is resolved with the target's request for the star but with the source's layout. Patching both the queries and the keys of these heads together moves the pointer itself, \texttt{K}, in 95.0\% of cases. In other words, the queries act as a proxy for the language instruction ``I need to look for stars'', and the keys of the image tokens act like ``I am the panel with the star''. Most of the query and key heads we find separately are binding heads (App.~\ref{app:one-screens}). When the binding heads read the source's image values instead, the target's answer is unchanged in 95.4\% of cases (Appendix~\ref{app:binding-checks}): the queries and keys still select the target's panel, and the patched values carry only the location, which is the same in both images. The letter is still read from the target's own image. So the pointer resolved at the shape token is executed later, at the colon token. How does it get there?

\subsection{Relay Heads Carry the Resolved Pointer to the Colon Token}
\label{subsec:relay}
The pointer is resolved at the shape token but executed at the colon token. Since attention is the only component that moves information between tokens, some heads at the colon token must fetch the pointer from the shape token (Figure~\ref{fig:relay_progress}a). We start by patching the values that each head at the colon token reads from each of the previous tokens, one head at a time, while its query and keys remain unchanged. We rank the heads by how much each patch raises the logit of \texttt{K}. We find that the top 16 attention heads, mostly in layers 19-23, carry the resolved pointer from the shape token to the colon token: patching their values moves the target to \texttt{K} in 87.5\% of cases (Appendix Figure~\ref{fig:appendix_relay}a). We call them \emph{relay heads}.

To localize the token the relay heads bring the pointer from, we now patch the values that all 16 relay heads at the colon token read, one token at a time~(Figure~\ref{fig:relay_progress}a). Reading the shape token alone moves the target to \texttt{K} in 47.3\% of cases, against 9.9\% for the question mark after it and at most 0.2\% for any other token (App.~\ref{app:one-sources}). The relay heads read the resolved pointer mostly from the shape token, where the binding heads wrote it.

To show that the relay heads carry the pointer resolved by the binding heads at the shape token to the colon token, we test whether they are necessary and sufficient. First, when we block the relay heads' attention from the colon token to the text tokens, the target keeps its own answer in only 7.8\% of cases, against 99.8\% when the same number of random heads from the same layers is blocked (App.~\ref{app:one-knockouts}).  Second, we take the binding head patching experiment of Section~\ref{subsec:pointer}, which patches the source's resolved pointer into the target at the shape token, but let only the relay heads at the colon token of an unedited target read the values from the shape token of this patched run. Here nothing else of the patched run reaches the target, not even its residual vectors, yet the target answers \texttt{K} in 89.0\% of cases, as often as in the binding experiment. Conversely, if we keep the binding patch but let the relay heads read the target's original values, the target's answer \texttt{T} comes back in 99.8\% of cases (App.~\ref{app:one-sources}). These collectively show that the binding heads resolve the pointer at the shape token, and the relay heads alone carry it to the colon token, where the model executes it to read the letter from the image.

%% file: figures/binding.tex
\begin{figure*}[t]
\centering
\includegraphics[width=\textwidth]{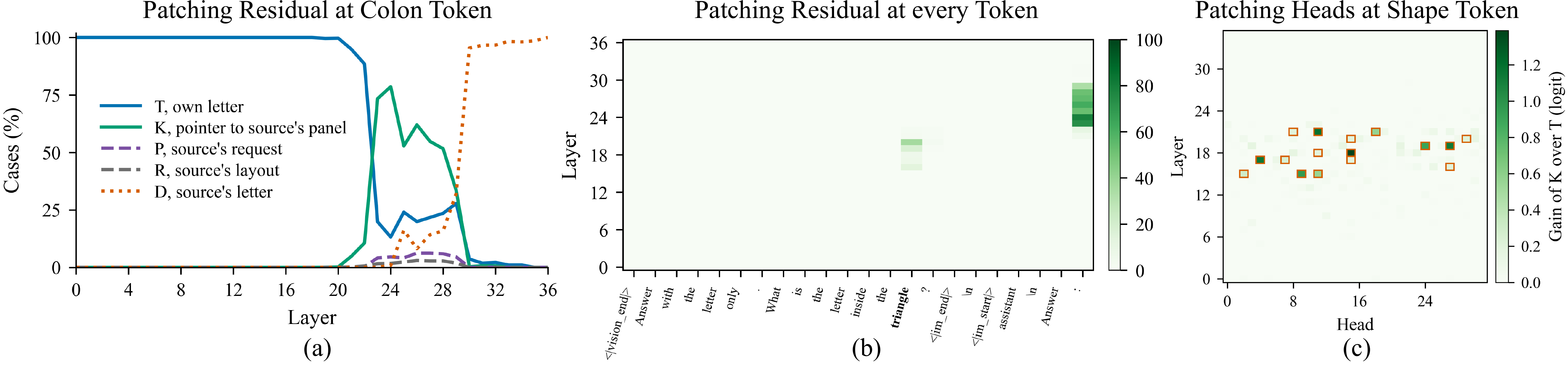}
\caption{\textbf{The pointer is resolved at the shape token by a few attention heads.} (a)~Patching the source's residual at the colon token into the target, one layer at a time. Up to layer 20 the target keeps its own letter \texttt{T}; from layers 23-29 it answers \texttt{K}, the target's letter at the source's answer location, so the colon token holds a pointer to a panel; from layer 30 it copies the source's letter \texttt{D}. (b)~The same patch at every text token and every layer, tracking the logit of \texttt{K}: the pointer appears only at the shape token, around layer 20, and at the colon token, from layer 23. (c)~The output of one attention head at the shape token replaced with the source's, for every head; each cell is the mean gain of \texttt{K} over \texttt{T}, and the 16 heads selected by the search sit in layers 15-21 and are outlined.}
\label{fig:binding}
\end{figure*}

%% file: secs/5_progress.tex
\input{figures/relay_progress}
\section{Progress: VLMs Keep a Cursor to Plan the Next Answer}
\label{sec:progress-select}

In Section~\ref{sec:binding}, we find that the binding heads resolve a pointer at the shape token, and the relay heads carry it to the colon token, where the model executes it to read the letter from the image. We now ask how the model solves the same task when several panels are correct and it has to answer all of them.  After each answer, the model has to keep track of how far it has come, which we call \emph{progress}, and pick the next panel, which we call \emph{select}. Interestingly, we find that progress is a pointer in the image, ``done up to panel 4'', which we call the \emph{cursor}. The same binding heads bind each written answer to the panel it came from, the comma token turns that panel into the cursor, and the model selects the nearest unwritten panel after it, or stops.

\subsection{The Comma Token Holds Progress at Layers 20-21}
\label{subsec:progress}

Since the language backbone is autoregressive, only the tokens after the first answer can carry progress: the answer and comma tokens. To localize progress, we patch the residual vector of the source's comma token into the target's colon token, one layer at a time. The source has written one letter and the target none, so the target has no answer token of its own to read, and anything it skips it learned from the patched state. The two runs ask for different shapes and share no letter. They share only the panel the source has just answered from, which is also the target's first panel:
\begin{center}
\begin{tabular}{@{}r@{\ }l@{\quad}l@{}}
Source: & \inp{diamond/M, square/P, star/X, circle/U, diamond/V, star/A}{star}  & prefilled ``\texttt{Answer: X,}'' \\
Target: & \inp{square/R, plus/S, heart/Y, heart/T, diamond/E, heart/W}{heart}   & prefilled ``\texttt{Answer:}'' \\
\end{tabular}
\end{center}
With no patch, the target answers \texttt{Y}. If the patch carries progress past the first answer, the target moves on to its second answer, \texttt{T}. If it carries the source's resolved pointer to its next answer, the target answers with its own letter at panel 6, \texttt{W}. If it is a shallow copy of the source's next answer, the target answers \texttt{A}, the source's letter at panel 6. Figure~\ref{fig:relay_progress}b shows the four outcomes at every layer. Up to layer 17 the target mostly keeps \texttt{Y}. At layer 20 it answers \texttt{T} in 91.0\% of cases, and in 70.0\% at layer 21. The target skips a panel it never answered from, because a source run with a different shape and different letters answered from it. From layer 22 the source's next pointer \texttt{W} takes over, and from layer 30 the patch is a shallow copy of \texttt{A}. This shows that the comma token at layers 20-21 holds progress, before it holds the next pointer, and that progress is about a panel, not about a shape or a letter. What form does this progress take?

\subsection{Progress Is an Ordinal Position: ``Done up to Here''}
\label{subsec:progress-cursor}

Progress could be a count (``two answers done''), a set of panels (``panels 2 and 4 done''), or an ordinal position (``done up to panel 4''). To tell them apart, we build a pair in which each form predicts a different answer. The source has written two letters, from panels 2 and 4, and its next answer is at panel 5. The target has hearts in panels 1, 2, 3, and 6, so its own first answer, panel 1, is not a panel the source has answered from:
\begin{center}
\begin{tabular}{@{}r@{\ }l@{\quad}l@{}}
Source: & \inp{diamond/M, star/X, circle/U, star/V, star/P, diamond/A}{star}    & prefilled ``\texttt{Answer: X, V,}'' \\
Target: & \inp{heart/Y, heart/T, heart/E, plus/R, square/S, heart/W}{heart}  & prefilled ``\texttt{Answer:}'' \\
\end{tabular}
\end{center}
We patch the source's comma token into the target's colon token as before. If progress is a count, ``two answers done'', the target skips two hearts and answers \texttt{E}. If it is a set of panels, ``panels 2 and 4 done'', the target answers with the first heart in neither panel, its own \texttt{Y}. If it is an ordinal position, ``done up to panel 4'', the target answers with the first heart after panel 4, \texttt{W}. Figure~\ref{fig:relay_progress}c shows the target's answer at every layer. Up to layer 7 the target answers \texttt{T} in 92.4 to 98.4\% of cases: a comma token by itself says ``one letter was written'', before it has read anything (App.~\ref{app:progress-state}). At layers 13-18 it answers \texttt{E} in 77.8 to 90.8\% of cases, so by layer 13 the comma token has counted the written letters. At layer 20 the target answers \texttt{W} in 98.4\% of cases, and in 97.6\% at layer 21. It keeps its own answer \texttt{Y}, which a set of panels would predict, in at most 1.2\% of cases at these layers. So at the layers where the comma token holds progress, progress is an ordinal position: the list is done up to panel 4. We call this position the \emph{cursor}. Where does the cursor come from?

\subsection{Binding Heads Set the Cursor at the Last Answer Token}
\label{subsec:progress-source}

The answer tokens are the first tokens that know what has been written. To test whether an answer token sets the cursor, we take a target that has written two letters, and a source image in which the last written letter sits one panel further to the right:
\begin{center}
\begin{tabular}{@{}r@{\ }l@{\quad}l@{}}
Source image: & \inp{heart/Y, square/M, heart/E, heart/T, plus/R, heart/W}{heart} & \\
Target:       & \inp{heart/Y, square/M, heart/T, heart/E, plus/R, heart/W}{heart} & prefilled ``\texttt{Answer: Y, T,}'' $\to$ \texttt{E}
\end{tabular}
\end{center}
The source image is the target's image with the letters of panels 3 and 4 swapped, so the shapes, the matching panels, and the written letters are unchanged, and only the panel holding \texttt{T} has moved. We let the binding heads at the last answer token, \texttt{T}, read the source's keys, while their queries and everything else stay the same. If the binding heads bind the written letter to its panel, the cursor moves from panel 3 to panel 4, and the target skips panel 4 and answers \texttt{W}. This happens in 84.2\% of cases with the 16 binding heads, and in 0.0\% with the same number of random heads from the same layers. The 32 heads that move the cursor the most at the answer token do so in 88.0\% of cases, and they sit in the layers of the binding heads (App.~\ref{app:many-mark}). The target repeats a written letter in at most 2.4\% of cases (App. Figure~\ref{fig:appendix_rebind}). The same patch at the first answer token, with \texttt{Y} moved to panel 4, leaves the answer at \texttt{E} in 96.4 to 98.8\% of cases. So the cursor is set at the last answer token.

The comma token needs this answer token. When we block the attention from the last comma token to the last answer token, the model writes the same letter again in 67.8\% of cases, while blocking its attention to the colon token leaves 99.8\% of answers unchanged (App.~\ref{app:progress-relay}). The cursor marks a panel, not a letter: when the next matching panel holds the letter just written, the model writes that letter again in 95.8\% of lists (App.~\ref{app:many-mark}). In other words, binding happens twice, largely through the same heads. At the shape token they bind the requested shape to its panels, and at an answer token they bind the letter just written to the panel it came from. By layer 21 the comma token has turned that panel into the cursor. The 8B and 32B hold this cursor cleanly (App.~\ref{app:model_size}).

\subsection{Select: The Nearest Unwritten Panel After the Cursor, or Stop}
\label{subsec:select}

With the matching panels and the cursor both at the comma token, we can read the choice from the model without any patch, by prefilling the letters out of order (Appendix Figure~\ref{fig:appendix_many_probe}). With only the second heart written, the model answers with the third in 92.8\% of lists and goes back to the first in 2.6\%. With only the third written, it answers with the fourth in 64.4\% and goes back to the second in 34.4\%. With the third and then the first written, it answers with the second in 95.6\%. So select starts at the cursor and takes the nearest unwritten panel, preferring the right, and goes back when nothing is left to the right. The written panels also count as done: with the second and then the first written, the model answers with the third in 87.2\% of lists, and when the last comma token cannot attend to the second letter, it writes that letter again in 52.8\% (App.~\ref{app:select-checks}). The gaze heads do not read a written panel: patching their queries to point at an already written panel moves the answer there in 0.6\% of cases, against 88.2\% for an unwritten one (App.~\ref{app:select-refuse}).

To localize where the choice is made, we hide the next panel from every text token from layer $l$ on, and sweep $l$ (Appendix Figure~\ref{fig:appendix_select_sweep}). Hidden from the first layer, the model moves on to the following panel in 84.8\% of cases, so the other matching panels are there to fall back on. Hidden from layer 20 on, it moves on in 72.2\% of cases, from layer 21 on in 37.4\%, and from layer 25 on in 2.8\%. Probes on the comma token's residual give the same window without any patch: the matching panels are readable with 95.9\% accuracy per panel at layer 17 and 99.3\% at layer 21, and the next panel with 98.0\% at layer 18 and 100.0\% at layer 20 (App.~\ref{app:select-readout}). The comma token also finds the matching panels again by itself: a source's set of matching panels patched into the shape token changes the next answer in only 21.0\% of cases, and in 64.5\% when the comma token keeps only the gaze heads' view of the image (App.~\ref{app:many-again}). So the choice is made at the comma token in layers 20-25.

The cursor acts on this choice through the comma token's residual. We move the cursor by letting the 32 answer-token heads read the source image's keys, and then hold the comma token's residual at the target's own values at one layer. Held at layer 23 or later, the target's own answer comes back in 99.4\% of cases, while held at layer 19 or earlier, the patch keeps its effect in 69.8\%. Holding the gaze heads' queries at the comma token instead brings the answer back in 99.8\% (App.~\ref{app:select-readout}). No small set of heads makes this choice: it is spread over layers 20-25, and the gaze heads read its result (App.~\ref{app:select-heads}). The same step decides when to stop. At the last letter the model chooses between a comma and the end of its turn. Patching the outputs of the heads found there makes a finished list go on in 100.0\% of cases and a continuing list stop in 96.8\%, against 0.2\% and 0.0\% for the same number of random heads (App.~\ref{app:select-checks}). On COCO images, when we block the comma token's attention to the object already described, the model names that object again in 79.6\% of cases (App.~\ref{app:coco-natural}).

We find one additional step in the plan when the prompt asks for a reading order. With \texttt{List them from right to left.} added to the question, 16 heads in layers 18-23 read this instruction. These heads act before the cursor forms, and the model uses what they read to set the direction of the cursor. Instead of taking the next panel to the right, the model takes the next panel to the left of the cursor. When we block these heads from the instruction, the list returns to left to right in 98.4\% of cases. 
Conversely, blocking every other head, the list stays right to left in 98.4\% (App.~\ref{app:right-to-left}).

%% file: figures/relay_progress.tex
\begin{figure*}[t]
\centering
\includegraphics[width=\textwidth]{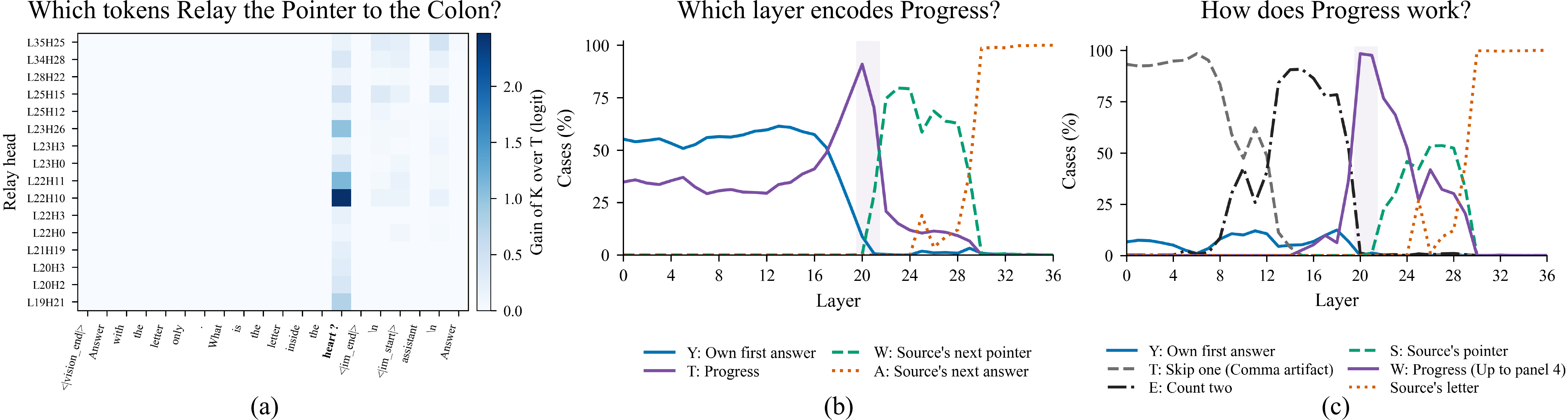}
\caption{(a)~\textbf{Relay heads transfer the pointer from the shape token to the colon.} When each relay head may read only one token at a time, the pointer is transferred mostly when the relay heads at the colon token read the shape token. Each cell is the mean gain of \texttt{K} over \texttt{T} on 500 held-out pairs. (b)~\textbf{A copied comma state skips a written panel at layers 20-21}. When patching the source's comma-token residual into the target's colon-token residual, one layer at a time: at layer 20 the target skips to its next candidate, \texttt{T}, in 91.0\pmv{2.4}\% of pairs, and in 70.0\pmv{3.9}\% at layer 21. From layers 22-28 it takes the source's next panel pointer, \texttt{W}, and from layer 30 the source's letter, \texttt{A}. (c)~\textbf{A copied comma state says where the list goes on}. With a more targeted setup, we find at layers 20 and 21 the model answers with the first candidate ``after panel 4'', \texttt{W}, in 98.4\pmv{1.1}\% and 97.6\pmv{1.3}\% of pairs. The shaded band marks layers 20-21.}
\label{fig:relay_progress}
\end{figure*}

%% file: secs/6_readers.tex
\input{figures/gaze_language}

\section{Executing the Pointer: Visual Retrieval and Translation}
\label{sec:reading}

Through binding, progress tracking, and selecting, VLMs resolve pointers to the image regions that contain the answer. In this section, we investigate how the model executes these pointers to retrieve the visual information and translate it into text.

\subsection{Gaze Heads Read the Panel Referenced by the Pointer}

The pointer is carried to the colon token, but the letter it points to is still in the image. Some attention heads at the colon token must build their queries using the pointer to be able to look back at the corresponding image region. We start by patching the queries of one attention head at a time at the colon token, from the source into the target, using the single-answer task setup from Section~\ref{sec:setup}. We rank the heads by how much each patch raises the logit of \texttt{K}, the target's letter at the source's panel. We find that the top 32 attention heads, in layers 19-34, execute the pointer: patching their queries moves the target to \texttt{K} in 74.0\% of cases, while the same number of random heads from the same layers does so in 0.0\% (Figure~\ref{fig:gaze_language}a). We call them \emph{gaze heads}.

We show that the gaze heads execute the pointer resolved by the binding heads, by testing whether the pointer acts through them and whether they are enough to read the image. First, we take the binding head patching experiment of Section~\ref{subsec:pointer}, which patches the source's resolved pointer into the target at the shape token, but keep the gaze heads' queries at the target's original values. The target keeps its own answer \texttt{T} in 98.4\% of cases, while keeping the same number of random heads lets the patched pointer through in 86.5\%. So when the gaze heads' queries are overwritten, the patched pointer has no effect, which shows that the pointer acts through them. Second, we block the attention to the image from every head at every token, except the shape token and the 32 gaze heads at the colon token. The target still answers \texttt{T} in 95.1\% of cases, against 4.8\% when random heads see the image in their place. So the image needs to be read at only few tokens: at the shape and answer tokens to resolve the pointer, and by the gaze heads at the colon token to execute it. The gaze heads' queries carry the pointer, the image keys match it by place, and the image values bring the letter. When the gaze heads read the source's image values instead, the target answers with the source's letter at its own panel, \texttt{C}, in 93.7\% of cases (App.~\ref{app:gaze-checks}). When we block the gaze heads' attention to the image, the target still answers \texttt{T} in 88.2\% of cases. This is not because other heads at the colon token read the pointer in their place, but because the tokens between the shape token and the colon token also read the letter from the image. When we block both paths, only 2.6\% of answers survive (App.~\ref{app:one-knockouts}).

Our gaze heads resemble the published Gaze Heads~\citep{gandikota2026gaze}. Unlike our gaze heads, which we find by patching, the published Gaze Heads were found by observing the attention patterns of a model describing comic book panels. Most of our gaze heads are among the published Gaze Heads (23 of 32 in the top 100), yet our smaller set is causally more effective. Patching the queries of the top 100 published heads moves the target to \texttt{K} in 64.2\% of cases, against 74.0\% for our 32 gaze heads, and letting them read the source's image values gives \texttt{C} in 50.0\%, against 93.7\% (Figure~\ref{fig:gaze_language}b). Our set is also smaller because our task is narrow: it retrieves text from images, while describing a comic panel draws on many types of visual content, such as color, objects, and mood. \citet{gandikota2026gaze} show that individual Gaze Heads specialize in fetching these types of content. When we repeat the search on tasks that retrieve a shape or a drawn object instead of text, we find partly different heads (App.~\ref{app:read-objects}).

\subsection{Gaze Heads Return a Concept, Not a Word}
\label{subsec:language}

The gaze heads read the region at the pointer, but it is unclear whether they return the meaning of the visual concept or the textual representation for it. To test this, we use colored shapes without letter labels and ask ``\texttt{What shape is the green marker?}'' in English, German, French, or Chinese. The model answers with the right shape in the question's language in 100.0\% of cases (App.~\ref{app:read-implementation}). This setup lets us study the model when several answers are correct depending on the language of the question (e.g., \texttt{circle} vs.\ \texttt{cercle}).

We take the collective LogitLens token distribution of the 32 gaze heads and find that it is nearly the same for one image under two languages, with a cosine of 0.95, and different for two images under one language, 0.50. The top logits of the gaze heads are always English or Chinese, irrespective of the question's language. When we patch the outputs of the gaze heads from a source that asks in another language about another image, the target names the source's shape in its own language in 82.0\% of cases and never changes the language, against 0.0\% for the same number of random heads. So the gaze heads return the concept, not the word (App.~\ref{app:read-content}). How does the concept become a word in the question's language?

The language is added by a different set of heads. We start by patching the output of one attention head at a time from a source that asks in another language. We rank the heads by how much each patch raises the logit of the target's answer in the source's language. We find that the top 16 attention heads, in layers 19-35, add the language: patching their outputs moves the language in 95.3\% of cases without ever copying the source's shape, while the same number of random heads does so in 0.0\% (Figure~\ref{fig:gaze_language}c). We call them \emph{language heads}. The late MLPs then use them to write the word (App.~\ref{app:read-language}). So an answer ends with a translation step: the gaze heads return a concept, the language heads add the language of the question, and the late MLPs join the two into a word. On COCO images with natural list questions (``\texttt{What animals are in this image?}''), the gaze heads at the comma token attend more to the object named next than to the object just named in 83.6\% of lists, and setting the language heads to their English averages turns 96.5\% of non-english answers into English (App.~\ref{app:coco-natural}). They also generalize to scattered shapes on a blank canvas (App.~\ref{app:no-panels}).

%% file: figures/gaze_language.tex
\begin{figure*}[t]
\centering
\includegraphics[width=\textwidth]{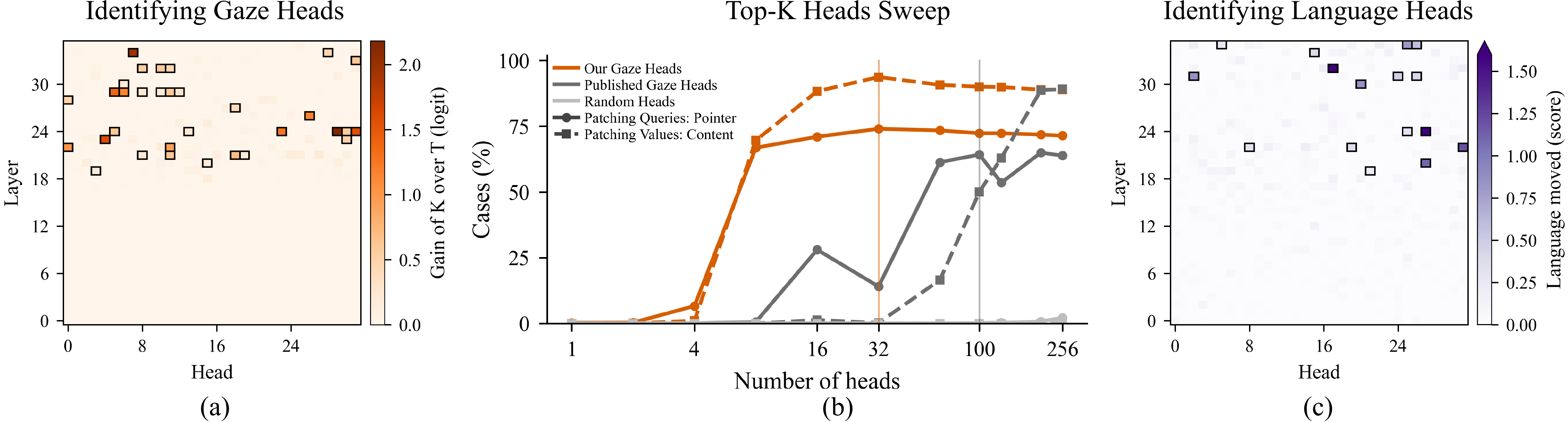}
\caption{(a)~\textbf{Gaze heads exist after layer 20.} Pointer logit gain when one head's query at the colon token is patched. The 32 gaze heads are outlined, all in layers 19-34. (b)~\textbf{Our gaze heads vs.\ the published Gaze Heads~\citep{gandikota2026gaze}.} Solid lines replace the queries and count the target's letter at the source's panel, \texttt{K}. Dashed lines let the heads read the source's image values and count the source's letter at the target's panel, \texttt{C}. The vertical lines mark our 32 heads and the top 100 published heads. Our 32 heads beat the top 100 published heads, and random heads stay at or below 2.1\pmv{1.0}\%. (c)~\textbf{Language heads set the language.} Scoring the language change of a word when patching attention head outputs from a source in a different language. The 16 language heads are outlined, in layers 19-35. The color scale is capped at 1.6; L24H27 scores 5.03.}
\label{fig:gaze_language}
\end{figure*}

%% file: secs/7_conclusion.tex
\section{Conclusion}
We find a structured plan inside VLMs for describing images. The model recomputes this plan at every answer to decide which region to describe next, then reads that region and narrates what it sees. At the shape token, binding heads match the user-requested text against the image to resolve a pointer to every matching image region. Relay heads carry this pointer to the current token, where gaze heads execute it and retrieve the visual content. Language heads along with the late MLPs translate the retrieved visual content into the language of the question. When several regions match, the model tracks its progress with the same components. Binding heads bind each written answer to the region it came from, the comma token turns that region into a cursor, the place in the image up to which the list has been written, and the model selects the nearest unwritten region after the cursor.

The plan is structured and modular: each step can be isolated and edited, with a clear causal effect on what the model outputs. This makes it a tool to explain, audit, and debug model outputs. At the same time, isolating this plan lets us steer the model's planning and its outputs. More broadly, our work offers a way to study how models plan over long, multi-step tasks and how they reason over what they see.

%% file: secs/misc.tex
\ificlrfinal
\section*{Acknowledgments}
RG, ASS and DB are supported by Open Philanthropy. RG and DB are supported by NSF grant \#2403304. AB and DB are supported by the NSF National Deep Inference Fabric Grant \#2408455. 

\section*{Code}
Source code, demo, and datasets for reproducing our results can be found at \href{https://plan.baulab.info/}{\textcolor[rgb]{0.21,0.49,0.74}{plan.baulab.info}} and at our GitHub repo \href{https://github.com/rohitgandikota/vlm-planning/}{\textcolor[rgb]{0.21,0.49,0.74}{github.com/rohitgandikota/vlm-planning}}.
\fi

\section*{AI use statement}

We used generative AI tools in four ways. First, to polish the writing and for spell check. Second, we used AI agents to scale up experiments that we designed and wrote ourselves, and to propose alternative hypotheses, which we then tested and ruled out in the paper. Third, we used them to search for related papers; we read the papers and wrote the related work ourselves. Fourth, we used them to clean up the structure of our codebase. We reviewed all AI-assisted work, and we take responsibility for the final content of this paper, including its text, claims, and code.

%% file: secs/appendix.tex
\appendix
\providecommand{\fittable}[1]{\resizebox{\ifdim\width>\linewidth\linewidth\else\width\fi}{!}{#1}}
\input{secs/appendix/A_implementation}
\input{secs/appendix/B_binding}
\input{secs/appendix/C_relay}
\input{secs/appendix/D_progress}
\input{secs/appendix/E_select}
\input{secs/appendix/F_reading_language}

\input{secs/appendix/G_generalization}

%% file: secs/appendix/A_implementation.tex
\section{Implementation Details}
\label{app:task}
\label{app:task-implementation}

This appendix gives everything the experiments of the main paper share: the model and its settings, the images and prompts of the two tasks, the pairs and splits, how a patch is applied and how attention is blocked, how heads are chosen, the language tasks, and how reliable the model is with no patch.

\subsection{Model and Settings}
Unless stated otherwise, all experiments use Qwen3-VL-8B-Instruct at revision \texttt{0c351dd0}. Appendices~\ref{app:model_size} and~\ref{app:model-families} give the settings of the other models. We load it in 32-bit floating point with the eager attention implementation, so that we can read and patch every attention weight and every intermediate state. TensorFloat-32 is turned off. Decoding is greedy. The language model has 36 layers, a hidden size of 4,096, and 32 query heads of width 128 per layer. It has 8 key-value groups, so four query heads share each group of keys and values. We never change a weight, and we apply every patch during a forward pass. The vision encoder runs once per image, and we cache its output and never patch it.

For each run we cache the keys and values of every token up to the last image token. We then recompute the text tokens after the image in one forward pass on top of this cache, and we apply every patch inside that pass. With no patch, this pass reproduces the letter scores of an ordinary forward pass to within $7\times10^{-5}$. We process several patched copies of one run as one batch.

\subsection{The Single-Answer Task}
\paragraph{Images.} Each image is a strip of 1,536 by 384 pixels holding four square panels of 384 pixels, each with a 3-pixel gray border on a white background. Each panel holds one marker, a shape outline with a 6-pixel black stroke and white fill. The shape is about 106 pixels across (140 for the star, whose inside is smaller), and one bold capital letter in 48-point DejaVu Sans Bold sits at its center. We draw the center of the marker uniformly inside the panel, at least 78 pixels from each edge. The source and the target of a pair share marker positions. The shapes are star, square, circle, triangle, diamond, plus, and heart, and each image uses four different shapes. Letters come from a pool of 21 capital letters (Appendix~\ref{app:task-native}). The image becomes 576 image tokens, a grid of 48 by 12, so each panel is 12 tokens wide.

\paragraph{Prompt.} The user turn holds the image followed by the text \texttt{Answer with the letter only. What is the letter inside the \{shape\}?} There is no system prompt. After the chat template's assistant tag we append \texttt{Answer:}, so the prompt ends with the colon token and the next token is the answer. Every shape name is one token, and every answer is the single token made of a space and the letter. The prompts of a source and a target have the same length and differ at exactly one position, the shape token. The image tokens sit at positions 4 to 579, the shape token at position 593, and the colon token at position 601. The tokens after the shape token are the question mark, the end-of-turn token, a newline, the assistant tag (two tokens), a newline, \texttt{Answer}, and the colon token.

\paragraph{Pairs and splits.} In each pair, the source and the target use the same four shapes in different orders, ask for different shapes, and use eight different letters. We keep a pair only if the five outcome letters of Section~\ref{sec:setup} (\texttt{T}, \texttt{D}, \texttt{K}, \texttt{P}, and \texttt{R}) are all different in both directions. The model must also answer both of its questions correctly with no patch. We use one set of source-target pairs to rank heads (the discovery pairs), a second set to fix the number of heads (the validation pairs), and a third, held-out set to report every result.

\paragraph{Scoring.} After a patch, we take the most likely next token over the full vocabulary and score it as one of the outcome letters or as a miss. Some tables also report \emph{Gaze}, the cases in which the published Gaze Heads \citep{gandikota2026gaze} look at the expected panel. For each of the top 100 heads of that ranking, we sum its attention from the colon token over the image tokens of each panel. We add these sums over the 100 heads and take the panel with the largest total. We never refit that ranking.

\subsection{The List Task}
\label{app:many-implementation}
\paragraph{Images.} Each image is one row of six panels of 384 by 384 pixels. Every panel holds one marker drawn as in the single-answer task. The markers are larger, so that the letter sits clear of the outline: the circle has a radius of 72 pixels and the star of 100. Marker centers are uniform in 106 to 278 pixels on both axes and are shared by all images of a pair.

\paragraph{Prompt.} The user turn holds the image followed by \texttt{Answer with the letters only, separated by a comma and a space. What is the letter inside each \{shape\}?} The assistant turn starts with \texttt{Answer:}. The model writes \texttt{ K, U, E}: a letter token, a comma token, a letter token. So the last comma token chooses the next letter, and the last letter chooses between a comma and the end of the turn. Letters that we write for the model are ordinary tokens of the assistant turn. Throughout the appendix we write $r_1<r_2<r_3<r_4$ for the target's four matching panels.

\paragraph{Which shapes.} We ask for stars, hearts, or triangles, which the model lists reliably. It does not list squares and pluses reliably. The other panels hold any of the remaining shapes, except the ones the model confuses with the asked shape in a preliminary check. So there is no triangle or plus when we ask for stars, and no diamond or heart when we ask for triangles. In the same check, when the asked shape sat in five of six panels, the model often listed the sixth panel too, so we use four.

\paragraph{Pairs.} Each pair has one target and several sources that share the asked shape, the marker centers, and the written letter. The target has the asked shape in four panels, which cycle over all 15 choices of four panels out of six. The main source has the asked shape at $r_2$ and $r_4$, and the target's first letter written at $r_2$. All its other letters are new, so a source letter in the answer is a copied read. For the first answer the same source carries a new letter in place of the shared one. For stopping, a source has the asked shape at $r_1$ only. The supporting experiments add sources with the asked shape at $r_1$ and $r_3$ (the moved set), at $r_3$ and $r_4$, and at $r_1$ and a panel that holds another shape in the target. The progress experiments of Section~\ref{sec:progress-select} draw their own pairs at random from the same shapes and letters, with the panels stated in each subsection. They keep a pair only if both unedited runs give the expected next letter.

\paragraph{Splits and counting.} The splits are those of the single-answer task: discovery, validation, and held-out pairs. We always test the head sets of Sections~\ref{sec:binding} and~\ref{sec:reading} first, unchanged, and only then search again. We count a case when the unedited runs it needs give the right next token. With no patch, the target's second letter is right in 97.4\pmv{1.0}\% of candidate pairs and the source's in 99.5\pmv{0.4}\%. The first letters are right in 95.9\pmv{1.2}\% and 99.2\pmv{0.5}\%. The source with one matching panel ends its turn in 96.1\pmv{1.2}\%.

\paragraph{Hiding a panel.} To hide a panel from a token we remove that token's attention to the image tokens of the panel, in all heads of the chosen layers. The instruction tokens before the shape token also read the image, so ``every text token'' means every token after the image.

\subsection{Patches, Blocked Attention, and How Heads Are Chosen}
\label{app:one-implementation}
Every patch acts while the target processes one token. We record source states from the source's own unedited run at the same token position. This is possible because the two prompts align token by token. Everything that is not named stays the target's own, and all later computation runs normally on the patched state.

\paragraph{Residual patch.} At layer $l$ we replace the residual vector of the token with the source's vector at the same layer. A patch at layer $l$ replaces the input to layer $l$, and layer 36 stands for the input to the final normalization.

\paragraph{Query patch.} For one head we replace the query vector of the token, after the query normalization and the rotary position encoding, with the source's. The two runs share positions, so the position encoding is identical.

\paragraph{Key patch.} For one head we recompute only its attention scores to the image tokens, using the target's query and the source's image keys from the key-value group that head reads. Its scores to the text tokens, its values, and all other heads are unchanged. Because four query heads share each group of keys, this patch changes what one head reads without changing the keys the other three heads see.

\paragraph{Value patch.} For a head processing the colon token or a comma token, let $\alpha_s$ be its attention weight on source token $s$ and $v_s$ the value it reads there. We replace the head's output $o$ with $o+\sum_{s\in S}\alpha_s(v_s^{\text{new}}-v_s)$, where $S$ is the set of source tokens named in the experiment. The new values $v^{\text{new}}$ come from the source, or from a run with the binding-head patch, which is the target with only the binding heads' outputs patched from the source at the shape token (Appendix~\ref{app:one-sources}). The attention weights are the target's own. For the gaze heads, we apply the same patch to the image tokens.

\paragraph{Output patch.} We replace the output of a head at the token, taken before the layer's shared output projection, with the source's. An MLP patch replaces the output of one MLP at the token.

\paragraph{Blocking attention.} To block the attention from a token to a set of source tokens, we set the chosen heads' attention scores from that token to those source tokens to the most negative value before the softmax. The softmax then renormalizes the head's remaining attention as usual. To hold a state at its own value we record it from the unedited target and write it back on top of the patch.

\paragraph{Choosing heads.} For each attention patch, we start by applying the patch to one head at a time, for all 1,152 heads, on the discovery pairs. We score each head by the change in the gap between the logit of the letter the patch should produce and the logit of the target's own letter, averaged over cases and both directions. We then apply the patch to the top $K$ heads together on the validation pairs, for $K$ in 1, 2, 4, \ldots, 128 (256 for the gaze heads). A rule fixed before these runs takes the smallest $K$ for which the patch moves the letter in at least 80\% of cases, or the best $K$ if none does. On the single-answer task, the Gaze count must reach 80\% as well. We then run the chosen heads once on the held-out pairs.

\paragraph{Controls.} We repeat every patch on a random set of heads of the same size. Each random head comes from the same layer as the head it replaces and lies outside the selected set. We repeat every blocking experiment on matched random heads or on a neighboring token.

\paragraph{Intervals.} A percentage written as $x\pm y$ carries the half-width $y$ of its 95\% interval. We resample the held-out pairs with replacement 2,000 times, keeping the two directions of a pair together, and take the 2.5th and 97.5th percentiles.

\subsection{Overview of the Main-Paper Experiments}
\label{app:overview}
Tables~\ref{tab:overview-binding} to~\ref{tab:overview-reading} list every experiment of the main paper in the order of the text. For each one, they give the token where we apply the patch, the activation taken from the source or the attention that we block, the answer that the tested hypothesis predicts, and the result.

\input{tables/appendix/experiment_overview}
\clearpage

\subsection{The Language Tasks}
\label{app:read-implementation}
Section~\ref{subsec:language} needs answers that differ between languages, which a letter does not. Images are strips of four panels as in the single-answer task. In the \emph{colored-shape} task each panel holds one filled shape, all four shapes differ, and exactly one is cyan. The other three take distinct colors from orange, green, pink, yellow, red, and purple. The question asks for the shape of the cyan marker: in English \texttt{What shape is the cyan marker? Reply with one shape word only.}, in German \texttt{Welche Form hat die cyanfarbene Markierung? Antworte nur mit einem Wort f\"ur die Form.}, in French \texttt{Quelle est la forme du marqueur cyan ? R\'eponds avec un seul mot de forme.}, and in Chinese the direct translation. In the \emph{object} task, each panel holds a shape outline with a small drawn object inside it, a color emoji from the Noto Color Emoji font at 96 pixels. The asked shape is always the circle, and the questions are \texttt{What is inside the circle? Answer with one word only.} and its translations. No prompt names a language, and the assistant turn is empty, so the last prompt token is the newline after the assistant tag. It is the same token in every language and every image, and all patches act there or at generated tokens. We generate answers greedily for up to eight tokens and match them against lists of accepted words per concept and language.

\paragraph{Objects.} We started from 24 objects whose names differ in all four languages and kept an object only if the model named it correctly at all four panel positions in all four languages. Twenty-one passed. The crescent moon (read as a croissant in French), the top hat, and the sneaker failed. We use 16 of the 21.

\paragraph{Images, splits, and cases.} Each task has a set of discovery images and a separate set of held-out images. We use the discovery images to rank heads and to compute the per-language averages and vectors. Concepts and panel positions are balanced. Each held-out target is paired with three source images of a different concept. Targets ask in German, French, or Chinese and sources in any other language, which gives nine ordered language pairs per patch. We fixed the number of language heads, the layers of the held MLPs, and the layer and scale of the language vector in an earlier study on comic images (Appendix~\ref{app:read-comic}), before any run on these tasks. We apply a patch at every generated token. When the patch uses a source, we advance the source run in step by feeding it the target's generated tokens. To \emph{set} a head to a language, we overwrite its output with the average of its output at the last prompt token under that language, over the discovery images. Nothing from the test image is involved.

\paragraph{The color question.} In the \emph{color question} each panel holds one shape in its own color, with no letter. We ask for the color of a named shape, so the shape name is again the shape token. The question, the shape name and the prefill are all written in one language, and the model answers with the color word in that language. For example, in French, \cinp{diamond/green!60!black, star/blue, heart/red, circle/black}{heart}[fr] prefilled ``\texttt{Réponse :}'' $\to$ \texttt{rouge}.

We document the prompts and color concepts used in each language in Table~\ref{tab:color-prompts} and Table~\ref{tab:color-words} respectively.

\input{tables/appendix/color_prompts}
\input{tables/appendix/color_words}

\begin{center}
\begin{tabular}{@{}r@{\ }l@{\quad}l@{\quad}l@{}}
Source:    & \cinp{diamond/green!60!black, star/blue, heart/red, circle/black}{heart}[fr] & prefilled ``\texttt{Réponse :}'' & $\to$ \texttt{rouge} \\
Target: & \cinp{star/yellow, diamond/purple, circle/gray, heart/pink}{diamond}[de]      & prefilled ``\texttt{Antwort:}''  & $\to$ \texttt{lila} \\
\end{tabular}
\end{center}

\paragraph{Reading the gaze heads' output.} We use the 32 gaze heads unchanged, at the last prompt token of the language tasks, which is the token that produces the first answer word. Their \emph{summed output} is the sum over these heads of the head's output after the output projection. We read it against the vocabulary by multiplying it with the output embedding, with no normalization. For the output patch of Section~\ref{subsec:language}, the target asks in language $B$ about image $Y$ and the source asks in language $A\neq B$ about one of three images $X$ with a different concept. We replace the outputs of a head set with the source's at every generated token, with the source run advanced on the target's tokens. The answer is scored as the source's concept in the target's language, the target's concept in the source's language, the source's word, or no change, on the cases whose target asks in German, French, or Chinese.

\subsection{How Reliable Is the Task?}
\label{app:task-native}
Before any patch, we ask about every shape in every image. We read the answer in two ways: the next token after \texttt{Answer:}, and free generation with no prefix. With the final pool of 21 letters, the next token after \texttt{Answer:} is correct in 98.9\pmv{0.2}\% of questions, and free generation with no prefix in 98.9\pmv{0.3}\%. The published Gaze Heads look at the asked panel in 97.9\pmv{0.4}\%.

We first ran this check with all 26 letters. The next token was correct in 98.0\% of questions when ignoring letter case. Of the errors, 61.0\% were letters that do not appear in the image at all, led by Q read as O, J as A, L as I, and P as D. In 97.6\% of these errors, the Gaze Heads were on the correct panel. The model was looking at the right place and misreading the letter. Only 0.8\% of the answers were the letter of another panel. We removed Q, O, I, L, and J from the letter pool for every experiment in this paper. On the language tasks, the model answers the unedited colored-shape questions in the question's language in 100.0\pmv{0.0}\% of runs and the object questions in 99.0\%.

%% file: tables/appendix/experiment_overview.tex
\begin{table}[ht]
\centering
\footnotesize
\caption{\textbf{The experiments of Section~\ref{sec:binding}, single-answer task.} \emph{Token} is the target token where the patch is applied. \emph{Patched or blocked} is the only activation taken from the source, or the attention that is blocked; everything else is the target's own (Appendix~\ref{app:one-implementation}). \emph{Target answers} is the answer that the tested hypothesis predicts, in the letters of Section~\ref{sec:setup}: \texttt{T} is the target's own letter, \texttt{D} the source's letter, \texttt{K} the resolved pointer, \texttt{P} the request, and \texttt{R} the layout. Percent of cases as reported in the main text, with random heads from the same layers in parentheses.}
\label{tab:overview-binding}
\begin{tabularx}{\linewidth}{@{}>{\raggedright\arraybackslash}p{2.4cm}>{\raggedright\arraybackslash}p{1.1cm}>{\raggedright\arraybackslash}X>{\raggedright\arraybackslash}p{2.3cm}>{\raggedright\arraybackslash}p{2.4cm}@{}}
\toprule
Experiment & Token & Patched or blocked & Target answers & Result \\
\midrule
Residual patch by layer & colon & the residual vector, one layer at a time & \texttt{K} & \texttt{K} at layers 23-29; \texttt{D} from layer 30 \\
Residual patch at every token & every text token & the residual vector, one token and one layer at a time & logit of \texttt{K} & shape token at layers 16-20; colon token at 23-29 \\
Binding heads & shape & outputs of the 16 heads, layers 15-21 & \texttt{K} & 85.9 (random $\le$0.1) \\
Query heads & shape & queries of the top 8 heads & \texttt{P} & 81.0 \\
Key heads & shape & the top 8 heads read the source's image keys & \texttt{R} & 83.1 \\
Query and key heads & shape & queries and image keys of both sets & \texttt{K} & 95.0 \\
Binding heads, image values & shape & the binding heads read the source's image values & \texttt{T} & 95.4 \\
Relay heads & colon & values that the 16 heads, mostly in layers 19-23, read from the previous tokens & \texttt{K} & 87.5 \\
Relay heads, one token at a time & colon & values that the relay heads read from one token only & \texttt{K} & shape token 47.3; question mark 9.9; any other $\le$0.2 \\
Relay heads blocked & colon & the relay heads' attention to the text tokens is blocked & \texttt{T} & 7.8 (random 99.8) \\
Relay heads read the binding patch & colon & the relay heads of an unedited target read the shape token's values from a run with the binding-head patch & \texttt{K} & 89.0 \\
\quad target's values kept & colon & binding-head patch kept; the relay heads read the target's original values & \texttt{T} & 99.8 \\
\bottomrule
\end{tabularx}
\end{table}

\begin{table}[ht]
\centering
\footnotesize
\caption{\textbf{The experiments of Section~\ref{sec:progress-select}, list task.} Columns as in Table~\ref{tab:overview-binding}. In the second and third rows the source's comma-token residual is patched into the target's colon token, where the target has written nothing. Rows marked \emph{no patch} change only the prefilled answer.}
\label{tab:overview-progress}
\begin{tabularx}{\linewidth}{@{}>{\raggedright\arraybackslash}p{2.4cm}>{\raggedright\arraybackslash}p{1.1cm}>{\raggedright\arraybackslash}X>{\raggedright\arraybackslash}p{2.3cm}>{\raggedright\arraybackslash}p{2.4cm}@{}}
\toprule
Experiment & Token & Patched or blocked & Target answers & Result \\
\midrule
Binding on a list & shape & no patch: image attention of the binding heads and the 2 query and key heads outside them, and a probe on their summed output & the matching panels & 79.6 of image attention; probe 97.7 \\
Progress at the comma token & colon & the source's comma-token residual, one layer at a time; the source wrote one letter, at the target's first matching panel & its second answer & 91.0 at layer 20; 70.0 at 21 \\
Count, set, or position & colon & the same patch; the source wrote two letters, from panels 2 and 4 & its third match (count), its first (set), or the first after panel 4 (position) & count 77.8-90.8 at layers 13-18; position 98.4 at 20, 97.6 at 21; set $\le$1.2 \\
Cursor set at the last answer token & last answer & the 16 binding heads read the keys of an image with the last written letter one panel to the right & the letter after the moved panel & 84.2 (random 0.0); the 32 answer-token heads 88.0 \\
\quad first answer token & first answer & the same patch, with the first written letter moved & its own answer & 96.4-98.8 \\
Comma token blocked from the answer token & last comma & attention to the last answer token is blocked & the written letter again & 67.8 (colon token instead: 99.8 unchanged) \\
Same letter in the next panel & last comma & no patch: the next matching panel holds the letter just written & that letter again & 95.8 \\
Letters prefilled out of order & last comma & no patch: the written letters are prefilled out of order & the nearest unwritten panel after the cursor & 92.8, 64.4, 95.6, 87.2 \\
Earlier letter blocked & last comma & attention to an earlier written letter is blocked & that letter again & 52.8 \\
Gaze heads sent to a written panel & last comma & gaze-head queries patched to point at a written panel & that panel's letter & 0.6 (unwritten panel 88.2) \\
Hiding the next panel & every text token & the next panel is hidden from layer $l$ on & the following panel & 84.8 (first layer), 72.2 (20), 37.4 (21), 2.8 (25) \\
Probes & last comma & no patch: linear probes on the residual & matching panels; next panel & 95.9 at layer 17, 99.3 at 21; 98.0 at 18, 100.0 at 20 \\
Matching panels patched & shape & a source's set of matching panels; also when the comma token keeps only the gaze heads' view of the image & a changed next answer & 21.0; 64.5 \\
Holding the comma token & last comma & cursor moved by the 32 answer-token heads; then the residual is held at the target's own values at one layer & its own answer & 99.4 held at layer 23 or later; gaze-head queries held, 99.8 \\
Stopping & last letter & outputs of the heads found there & goes on; stops & 100.0; 96.8 (random 0.2; 0.0) \\
COCO, object blocked & last comma & attention to the object already described is blocked & that object again & 79.6 \\
\bottomrule
\end{tabularx}
\end{table}

\begin{table}[ht]
\centering
\footnotesize
\caption{\textbf{The experiments of Section~\ref{sec:reading}.} Columns as in Table~\ref{tab:overview-binding}; \texttt{C} is the source's letter at the target's panel. The language rows use colored shapes without letters and the question ``\texttt{What shape is the green marker?}'' in English, German, French, or Chinese. They have no prefilled answer, so their patches are applied at the last token and at every generated token.}
\label{tab:overview-reading}
\begin{tabularx}{\linewidth}{@{}>{\raggedright\arraybackslash}p{2.4cm}>{\raggedright\arraybackslash}p{1.1cm}>{\raggedright\arraybackslash}X>{\raggedright\arraybackslash}p{2.3cm}>{\raggedright\arraybackslash}p{2.4cm}@{}}
\toprule
Experiment & Token & Patched or blocked & Target answers & Result \\
\midrule
Gaze heads & colon & queries of the 32 heads, layers 19-34 & \texttt{K} & 74.0 (random 0.0) \\
Pointer through the gaze heads & shape, colon & binding-head patch at the shape token; gaze-head queries kept at the target's original values & \texttt{T} & 98.4; with random heads kept instead, the pointer gets through in 86.5 \\
Only the gaze heads see the image & every token & attention to the image is blocked, except at the shape token and for the 32 gaze heads at the colon token & \texttt{T} & 95.1 (random 4.8) \\
Gaze heads, image values & colon & the gaze heads read the source's image values & \texttt{C} & 93.7 \\
Gaze heads blocked & colon & the gaze heads' attention to the image is blocked & \texttt{T} & 88.2; 2.6 when the tokens between the shape token and the colon token are blocked as well \\
Published Gaze Heads & colon & queries of the top 100; image values of the top 100 & \texttt{K}; \texttt{C} & 64.2; 50.0 \\
Shape question in four languages & last token & no patch & the right shape in the question's language & 100.0 \\
Gaze heads in LogitLens & last token & no patch: cosine of their LogitLens token distributions & one image, two languages; two images, one language & 0.95; 0.50 \\
Gaze heads, another language & last token & outputs of the gaze heads from a source that asks in another language about another image & the source's shape in the target's language & 82.0 (random 0.0) \\
Language heads & last token & outputs of the 16 heads, layers 19-35, from a source that asks in another language & the target's shape in the source's language & 95.3 (random 0.0) \\
COCO, gaze heads & last comma & no patch: attention of the gaze heads & more on the object named next than on the one just named & 83.6 \\
COCO, language heads & last token & the language heads are set to their English averages & a non-English answer turns English & 96.5 \\
\bottomrule
\end{tabularx}
\end{table}

%% file: tables/appendix/color_prompts.tex
\begin{table}[ht]
\centering
\small
\caption{\textbf{The task instruction in each language.} The user turn is the question followed by the single-word instruction, and the assistant turn is prefilled with the answer prefix.}
\label{tab:color-prompts}
\begin{CJK}{UTF8}{gbsn}
{\fontencoding{T1}\selectfont
\fittable{%
\begin{tabular}{@{}llll@{}}
\toprule
 & Question & Instruction & Prefill \\
\midrule
English & What is the color of the \{shape\}? & Answer with a single word. & \texttt{Answer:} \\
French & Quelle est la couleur de la forme «~\{shape\}~» ? & Réponds par un seul mot. & \texttt{Réponse :} \\
Chinese & \{shape\}是什么颜色？ & 请用一个词回答。 & 答案： \\
German & Welche Farbe hat die Form „\{shape\}“? & Antworte mit einem einzigen Wort. & \texttt{Antwort:} \\
\bottomrule
\end{tabular}}}
\end{CJK}
\end{table}

%% file: tables/appendix/color_words.tex
\begin{table}[ht]
\centering
\small
\caption{\textbf{The eight color concepts in each language.} A readout counts the first token of the word. Forms in parentheses are also accepted in that language.}
\label{tab:color-words}
\begin{CJK}{UTF8}{gbsn}
{\fontencoding{T1}\selectfont
\fittable{%
\begin{tabular}{@{}lllll@{}}
\toprule
Concept & English & French & Chinese & German \\
\midrule
red & red & rouge & 红色 & rot \\
blue & blue & bleu (bleue) & 蓝色 & blau \\
green & green & vert (verte) & 绿色 & grün \\
yellow & yellow & jaune & 黄色 & gelb \\
purple & purple & violet (violette) & 紫色 & lila (violett) \\
black & black & noir (noire) & 黑色 & schwarz \\
pink & pink & rose & 粉色 (粉红色, 粉) & rosa (pink) \\
gray & gray (grey) & gris (grise) & 灰色 (灰) & grau \\
\bottomrule
\end{tabular}}}
\end{CJK}
\end{table}

%% file: secs/appendix/B_binding.tex
\section{Binding: Details and Supporting Experiments}
\label{app:binding}

This appendix supports the binding heads of Section~\ref{sec:binding}, which resolve the pointer. It covers the residual patch at every token, the head searches behind the binding heads, the MLP that writes the request, what the binding heads add to the residual stream, which requests they bind, and how they behave on a list.

\subsection{The Residual Patch at Every Token}
\label{app:one-token-map}
\input{figures/appendix/marker_sweep}
Section~\ref{subsec:pointer} tracks the pointer by patching the residual vector at every token and every layer. Figure~\ref{fig:token-map} shows this map for all five outcome letters of Section~\ref{sec:setup}. Patching any token before the shape token never changes the answer. After the shape token, no single token changes the answer until the colon token. So the state that travels between the two is spread over several tokens. This is why Section~\ref{sec:binding} patches the values of all of them together when it looks for the relay heads.

The shape token carries two things in turn (Figure~\ref{fig:marker-sweep}). Up to layer 15, the patch moves the request: the target answers the source's question on its own image, \texttt{P}. Around layer 20, the patch moves the pointer, \texttt{K}, in 36.1\pmv{3.2}\% of the held-out cases. The effect then fades, because the relay heads have already read the shape token by then. At the colon token, the pointer holds from layers 23-29 and peaks at 78.6\pmv{2.5}\% at layer 24. From layer 30 on, the patch copies the source's letter.

\input{figures/appendix/token_map}

\subsection{Head Searches and the Number of Heads}
\label{app:one-screens}
To choose the binding heads, we run three single-head searches at the shape token, one each for the query, key, and output patches. Figure~\ref{fig:appendix_binding_qk} shows the score of every head over all 36 layers and 32 heads. The three searches pick out heads in the same band of middle layers, and many heads rank high in all three. Figure~\ref{fig:appendix_k_sweep} reports every $K$ on the validation pairs. The query and key patches pass the selection rule of Appendix~\ref{app:one-implementation} at 8 heads, and the output and relay patches pass it at 16. Beyond these sizes, the key, output, and relay patches stay flat, while adding more heads makes the query patch worse. The binding heads are the top 16 heads of the output search of Section~\ref{subsec:pointer}. The query and key searches select their own 8 query heads and 8 key heads, and we report each patch with its own heads. Together, the three sets form a union of 18 heads in layers 15-21, and the experiments that use this union say so. Table~\ref{tab:one-answer} gives every patch on the held-out cases, and Table~\ref{tab:selected-heads} lists the selected heads.

\input{figures/appendix/binding_qk}
\input{tables/appendix/one_answer_edits}
\input{tables/appendix/selected_heads}
\input{figures/appendix/k_sweep}

\input{figures/appendix/mlp_sweep}
\paragraph{Queries that carry the pointer.} The query patch is the least clean of the three. With the 8 selected heads, the target answers the source's question on its own image in 81.0\pmv{2.5}\% of cases. But in 8.1\pmv{1.8}\% of cases, it answers at the source's \emph{location}. A query that moves the pointer must already say where to look, not only what to look for. The pointer forms inside the shape token while these heads run (Figure~\ref{fig:marker-sweep}), so the queries of some of them carry part of it. When we patch one head at a time, two heads in layer 19, L19H24 and L19H27, move the request almost alone. Heads in layers 15, 17, 18, and 21 (L15H9, L15H10, L17H4, L18H15, L21H11) move the pointer about as much as the request. We then rank the heads by how much a query moves the request \emph{minus} how much it moves the pointer. With this ranking, the patch needs 128 heads to pass the rule, and it then moves the request in 94.5\% of cases and the pointer in 0.0\%. So once the heads that mix in the pointer are ranked low, the queries carry the request alone.

\subsection{The First MLP Writes the Request}
\label{app:one-mlp}
The queries of the query heads carry the request, so something must write the request into the shape token before the binding layers. To find it, we patch the output of one MLP at a time at the shape token, for all 36 layers. Only the first MLP moves the request. Patching its output makes the target answer the source's question in 98.3\pmv{0.8}\% of cases, and no other MLP does so in more than 6.1\pmv{1.4}\% (Figure~\ref{fig:mlp-sweep}). The shape name is one token, so the first MLP is where this token's identity becomes a state that the binding heads can use.

\subsection{What the Binding Heads Write}
\label{app:binding-checks}
The binding heads could add a code for the panel, or the content of that panel. To tell these apart, we patch the queries, keys, and values of the heads one part at a time. When the 10 query and key heads, 8 of which are binding heads, keep their own queries and keys but read the source's image values, the answer is unchanged in 96.8\pmv{1.0}\% of the held-out cases. The same patch on the 16 binding heads leaves it unchanged in 95.4\pmv{1.2}\%. When we patch the queries, image keys, and values of these 10 heads together, the pointer moves in 82.6\pmv{2.4}\%. Random heads change the answer in at most 0.1\% of cases under any of these patches. So the binding heads add to the residual stream a code for where the answer is, not for what is there. The gaze heads do the opposite: their image values carry the letter (Appendix~\ref{app:gaze-checks}).

\subsection{Binding Heads Bind Any Request Word}
\label{app:other-bindings}
Every test of Section~\ref{sec:binding} names a shape. To test whether the binding heads bind shape words or any request, we change what the question names and keep the head sets of the main text unchanged. These are the same binding heads, relay heads, and gaze heads, with no new search. We use four variants of the four-panel task. In the color variant, every panel holds a circle filled with a distinct color, and the question is ``\texttt{What is the letter inside the red circle?}'', so the shape token is the color word. In the category variant, each panel holds a small drawn object with a letter next to it, and the question is ``\texttt{What is the letter next to the cat?}''. In the two-attribute variant, the panels hold colored shapes, with a red circle and a blue star as distractors for ``\texttt{the red star}'', so the request spans two words. In the word-label variant, the panels hold the paper's shapes with a three-letter word in place of the letter. We score it on the word's initial under the paper's question, and on the word under a question that asks for the word. With no patch, the model answers all four variants in 100.0\% of cases, and in 95.4\pmv{1.3}\% for word labels. Table~\ref{tab:other-bindings} repeats the patches of Sections~\ref{sec:binding} and~\ref{sec:reading} on each variant.

\input{tables/appendix/other_bindings}

The chain is the same in every variant. The binding heads' queries move the request, and their image keys move the layout. Their queries and keys together, and their outputs, move the pointer. The relay values carry the pointer, and restoring them restores the answer. These patches work in 75 to 100\% of cases, against at most 0.7\% for random heads. With two attributes, the two words divide the request. The queries at the color word alone move the color part: the target names the item with the source's color and its own shape in 78.6\pmv{2.5}\% of cases. The queries at the shape word alone move the shape part, in 85.9\pmv{2.1}\%. The outputs move the pointer mainly at the second word: in 72.4\pmv{2.9}\% of cases at the shape word, 23.8\pmv{3.3}\% at the color word, and 93.8\pmv{1.5}\% at both. With word labels, the binding and relay chain is unchanged, but the 32 gaze heads bring back only a letter's worth of content. When the model must write the word, the gaze heads' image values give the source's word in 1.4\pmv{0.8}\% of cases, and its initial in 15.8\pmv{2.2}\% more. When only the shape token and the gaze heads see the image, the model writes its own word in 16.1\pmv{2.3}\% of cases and its initial in 79.9\pmv{2.4}\%. So the binding heads bind whatever the question names to the image, while the gaze heads read one letter.

\subsection{Binding on a List}
\label{app:binding-tokens}
\paragraph{The shape-token heads find every matching panel, and write the set flat.} With four panels holding the asked shape, the 16 binding heads and the 2 query and key heads outside them at the shape token spread their image attention over all of them. Of this attention, 79.6\pmv{0.2}\% falls on the four matching panels, and 27.4\pmv{0.8}\% of that falls on the panel the model answers from first. To read what these heads write, we take their summed contribution to the residual stream at the shape token on the held-out pairs of the list task. We fit a ridge probe on half of the pairs and read it on the other half. A probe for ``panel $p$ has the asked shape'' reads each panel with 97.7\pmv{0.9}\% accuracy. A probe for ``which panel is first'' reads it in 100.0\pmv{0.0}\% of pairs. But a probe that sees only the six matching-panel scores already reads it in 92.8\pmv{3.2}\%, so the first panel is the leftmost matching panel and not a separate signal. In the matching-panel directions, centered, the first matching panel scores $+0.31$, the other matching panels $+0.28$, and the non-matching panels $-0.58$. So the set is written flat, and Appendix~\ref{app:many-again} follows it to the answer token.

%% file: figures/appendix/marker_sweep.tex
\begin{wrapfigure}{R}{0.4\textwidth}
\vspace{-10pt}
\centering
\includegraphics[width=\linewidth]{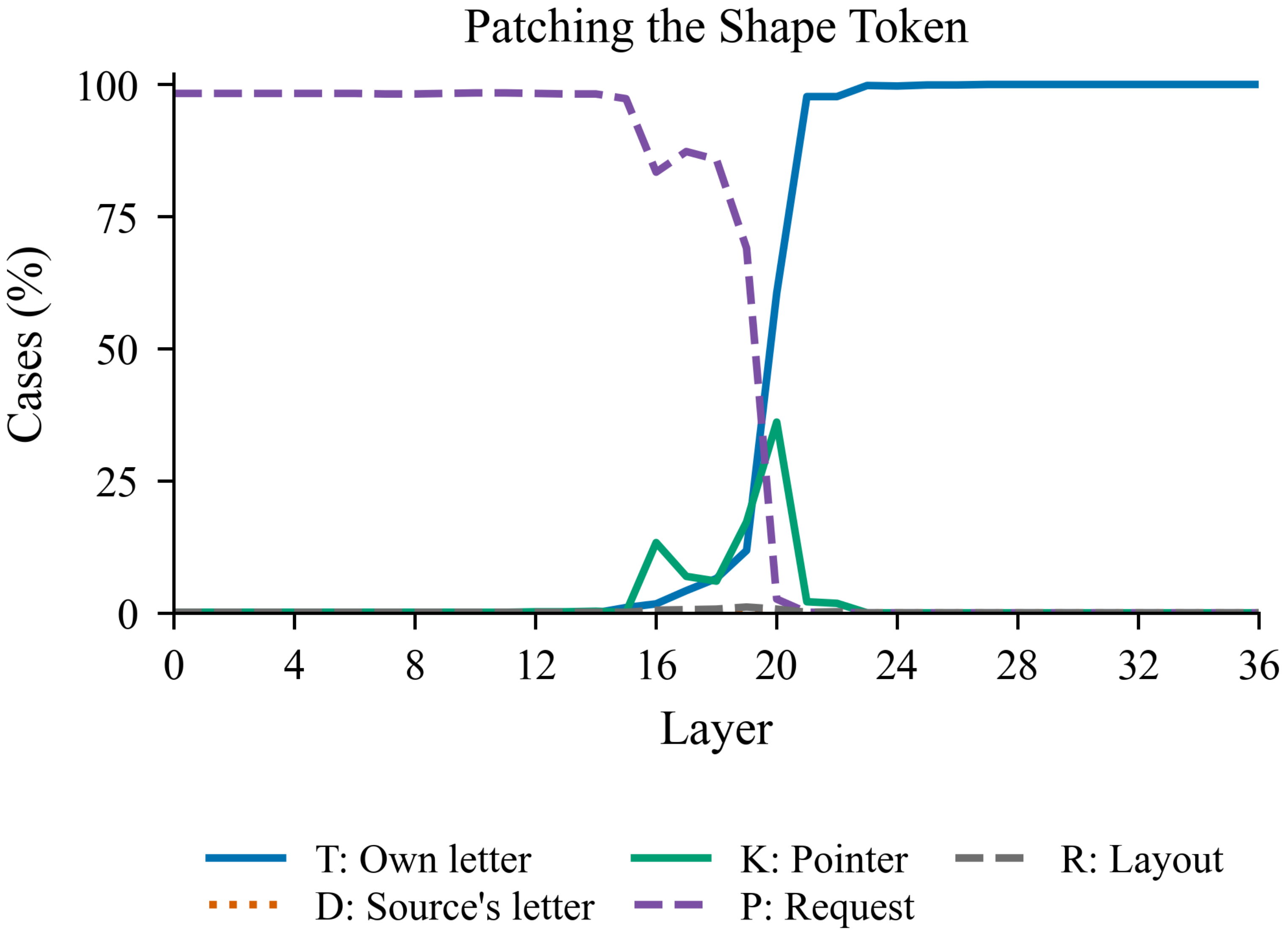}
\caption{\textbf{At the shape token, the request comes before the pointer.} We patch the shape token's residual vector one layer at a time. The patch moves the request, \texttt{P}, up to layer 15, and the pointer, \texttt{K}, at layer 20, in 36.1\pmv{3.2}\% of the held-out cases.}
\label{fig:marker-sweep}
\vspace{-10pt}
\end{wrapfigure}

%% file: figures/appendix/token_map.tex
\begin{figure*}[t]
\centering
\includegraphics[width=\textwidth]{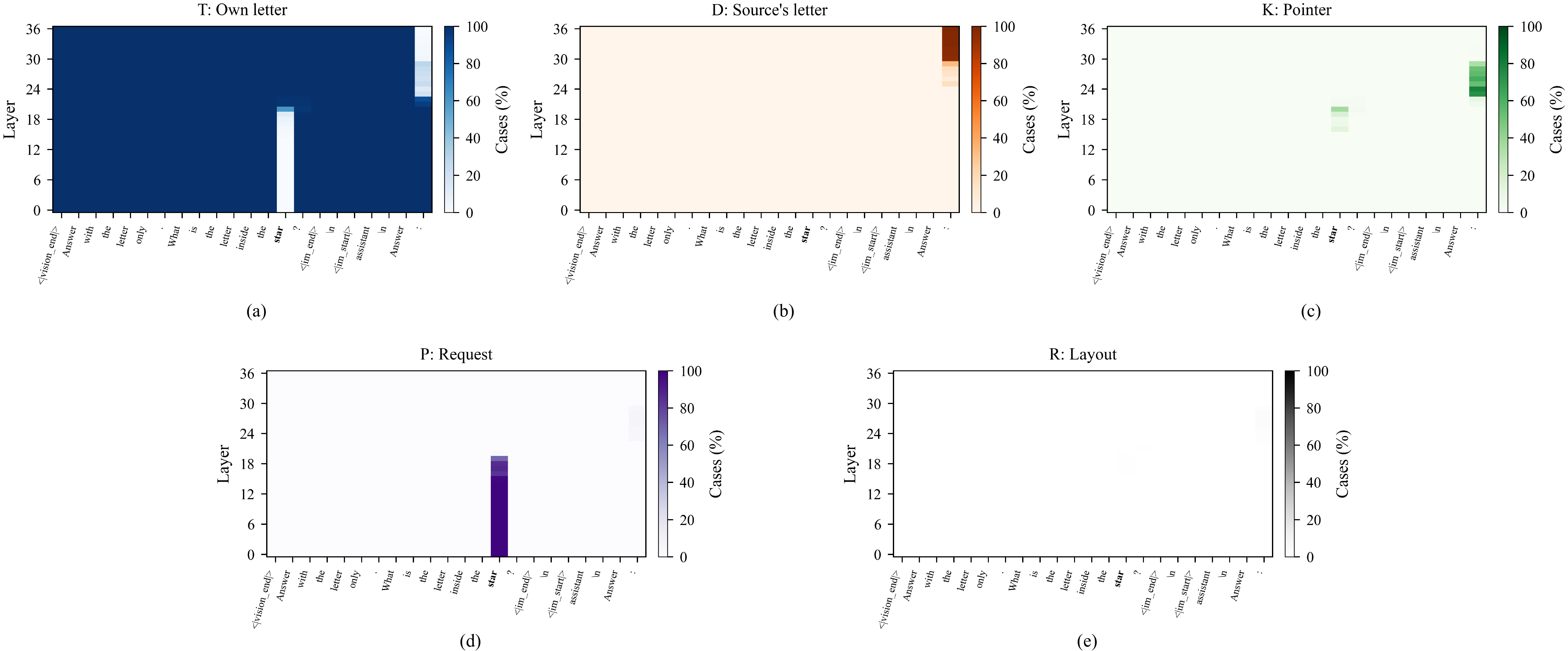}
\caption{\textbf{Only the shape token and the colon token change the answer.} Each cell patches the source's residual vector into the target at one text token and one layer, and counts the cases that answer with one outcome letter of Section~\ref{sec:setup}. (a)~\texttt{T}, the target's own letter. (b)~\texttt{D}, the source's letter. (c)~\texttt{K}, the target's letter at the source's panel. (d)~\texttt{P}, the target's letter inside the source's shape. (e)~\texttt{R}, the target's letter where the source's layout puts the asked shape. The shape token is shown in bold. The request, \texttt{P}, sits at the shape token up to layer 19. The pointer, \texttt{K}, sits at the shape token around layer 20 and at the colon token from layers 23-29. The source's letter, \texttt{D}, comes through the colon token from layer 30. All panels use the 500 held-out pairs.}
\label{fig:token-map}
\end{figure*}

%% file: figures/appendix/binding_qk.tex
\begin{figure*}[t]
\centering
\includegraphics[width=\textwidth]{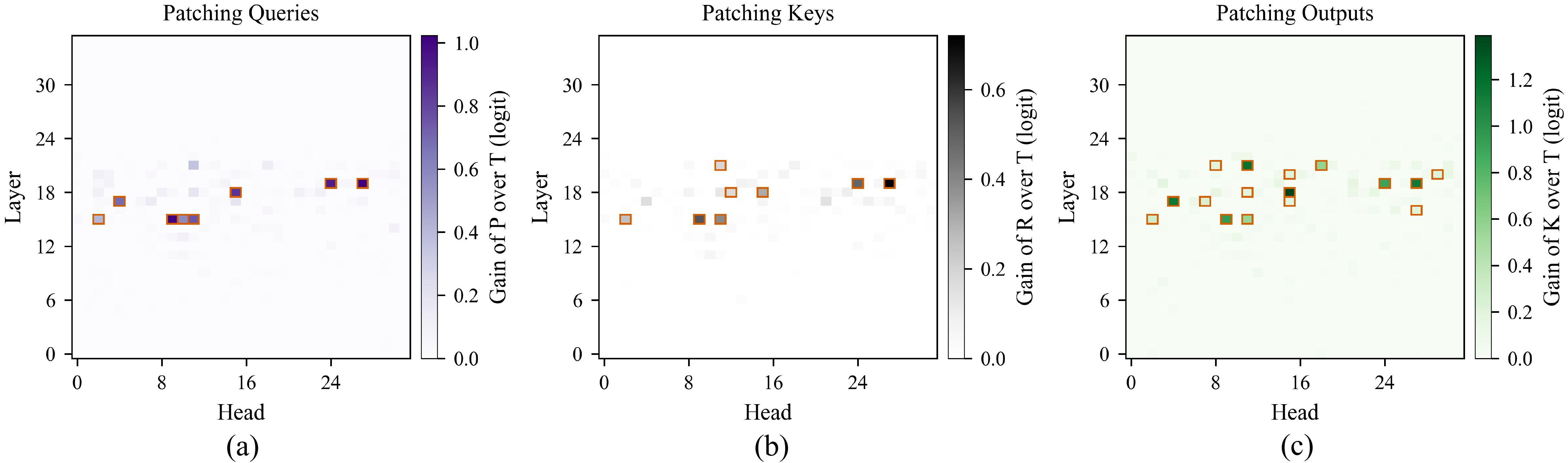}
\caption{\textbf{Queries carry the request, keys carry the layout, and outputs carry the pointer.} Each cell patches one attention head at the shape token from the source on its own, averaged over the 200 discovery pairs. (a)~We patch the head's query and score the gain of \texttt{P}, the target's letter inside the source's shape, over \texttt{T}. (b)~We let the head read the source's image keys and score the gain of \texttt{R}, the target's letter where the source's layout puts the asked shape, over \texttt{T}. (c)~We patch the head's output and score the gain of \texttt{K}, the pointer, over \texttt{T}. The heads that each search selects are outlined: 8 query heads, 8 key heads, and 16 output heads, all in layers 15-21. The 16 output heads are the binding heads of the main text.}
\label{fig:appendix_binding_qk}
\end{figure*}

%% file: tables/appendix/one_answer_edits.tex
\begin{table}[ht]
\centering
\small
\caption{\textbf{Each patch moves one variable.} Each row patches one kind of state from the source into the target, at the shape token or, for the relay values, at the colon token. \emph{Letter} is the percent of cases whose next token is the letter that the moved variable predicts. \emph{Gaze} is the percent of cases in which the published Gaze Heads, which we do not patch, move to the matching panel. All rows use the 500 held-out pairs.}
\label{tab:one-answer}
\fittable{%
\begin{tabular}{@{}lccc@{}}
\toprule
Patch (token) & Moved & Letter & Gaze \\
\midrule
Queries of 8 heads (shape token) & request & 81.0\pmv{2.5}\% & 73.3\pmv{2.8}\% \\
Image keys of 8 heads (shape token) & layout & 83.1\pmv{2.3}\% & 82.3\pmv{2.3}\% \\
Queries and image keys, 10 heads (shape token) & pointer & 95.0\pmv{1.5}\% & 93.0\pmv{1.6}\% \\
Outputs of 16 heads (shape token) & pointer & 85.9\pmv{2.4}\% & 81.2\pmv{2.5}\% \\
Values read by 16 heads (colon token), from the binding-head patch & pointer & 89.0\pmv{2.1}\% & 85.2\pmv{2.2}\% \\
Same, source's own values & pointer & 87.5\pmv{2.0}\% & 87.8\pmv{1.9}\% \\
Same, target's values restored & none & 99.8\pmv{0.2}\% & 99.8\pmv{0.2}\% \\
Output of the first MLP (shape token) & request & 98.3\pmv{0.8}\% & 98.4\pmv{0.8}\% \\
\midrule
Strongest random-head control & & 0.1\pmv{0.1}\% & \\
\bottomrule
\end{tabular}}
\end{table}

%% file: tables/appendix/selected_heads.tex
\begin{table}[ht]
\centering
\small
\caption{\textbf{The selected heads.} L and H give the layer and the head. The binding heads are the 16 heads that the output search selects. Eight of the 10 query and image-key heads are binding heads, so together they form a union of 18 heads, and the experiments that use this union name it.}
\label{tab:selected-heads}
\fittable{%
\begin{tabular}{@{}lp{10.5cm}@{}}
\toprule
Patch & Selected heads \\
\midrule
Queries & L19H27, L15H9, L19H24, L18H15, L15H11, L17H4, L15H10, L15H2 \\
Image keys & L19H27, L15H9, L19H24, L15H11, L18H15, L15H2, L21H11, L18H12 \\
Head outputs & L18H15, L21H11, L17H4, L19H27, L15H9, L19H24, L21H18, L15H11, L15H2, L17H7, L20H29, L21H8, L16H27, L20H15, L17H15, L18H11 \\
Values at the colon token (relay heads) & L22H10, L22H11, L35H25, L23H26, L25H15, L19H21, L34H28, L23H0, L20H2, L20H3, L23H3, L25H12, L28H22, L21H19, L22H3, L22H0 \\
Queries at the colon token (gaze heads) & L24H29, L34H7, L23H4, L24H31, L29H5, L24H23, L26H26, L29H6, L22H0, L22H11, L21H18, L24H5, L21H11, L33H31, L32H10, L29H11, L24H30, L32H8, L23H30, L28H0, L34H28, L32H11, L27H18, L29H12, L21H8, L24H13, L29H8, L30H6, L20H15, L29H10, L21H19, L19H3 \\
\midrule
Queries, comic images & L15H2, L15H9, L15H11, L17H4, L18H15, L19H24, L19H27, L21H11 \\
Image keys, comic images & L15H2, L15H9, L15H11, L17H4, L17H21, L18H15, L19H24, L19H27 \\
\bottomrule
\end{tabular}}
\end{table}

%% file: figures/appendix/k_sweep.tex
\begin{figure*}[t]
\centering
\includegraphics[width=\textwidth]{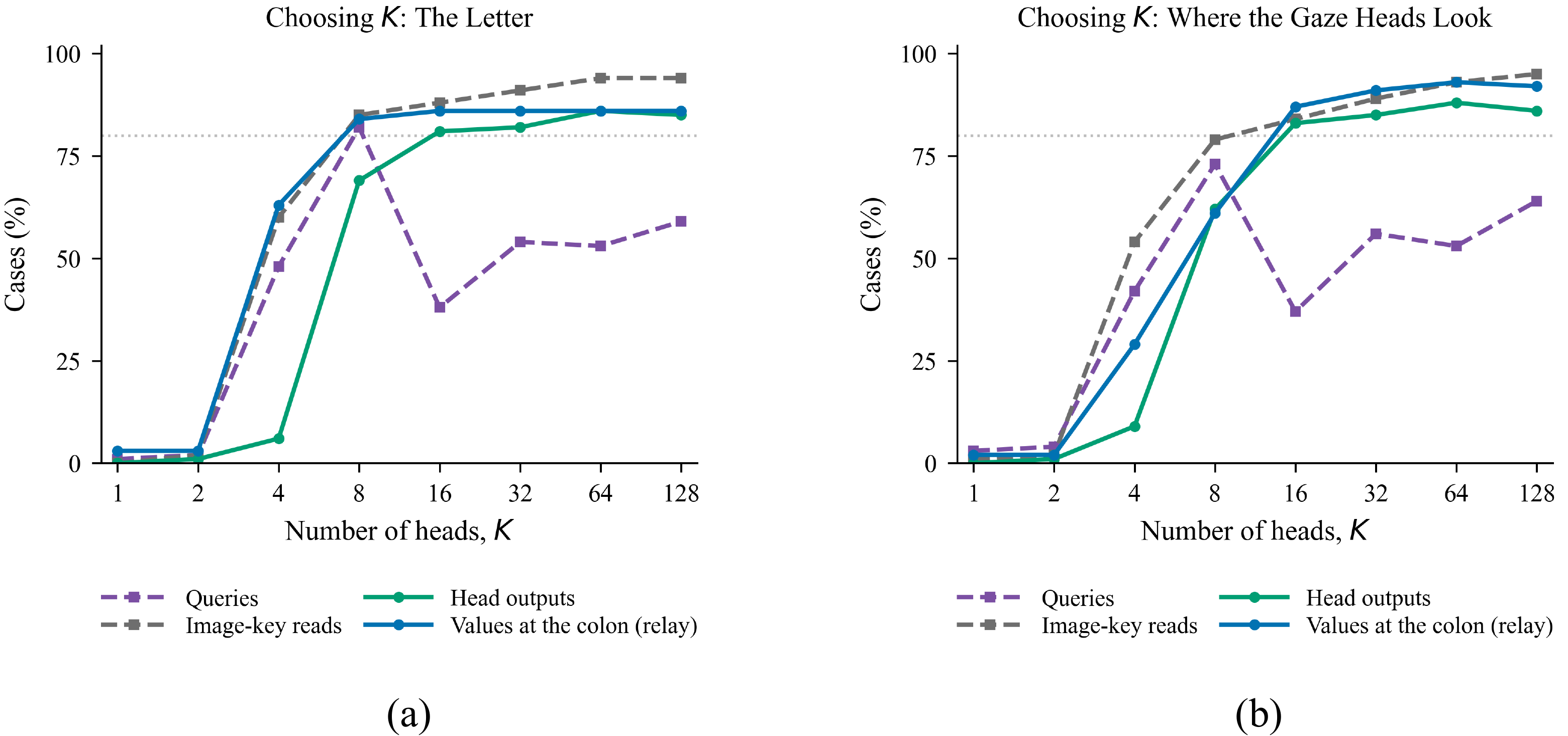}
\caption{\textbf{The selection rule fixes the number of heads.} We patch the top $K$ heads of each single-head search together on the 100 validation pairs. These are the queries, image keys, and outputs at the shape token, and the values that the relay heads read at the colon token. (a)~Cases whose next token is the letter the patch predicts. (b)~Cases in which the published Gaze Heads look at the matching panel. The dotted line marks the 80\% rule. The queries and the image keys first pass it on both readouts at 8 heads, with 83.5\% and 80.5\% for the queries and 85.0\% and 84.5\% for the image keys. The outputs and the relay values first pass it at 16 heads.}
\label{fig:appendix_k_sweep}
\end{figure*}

%% file: figures/appendix/mlp_sweep.tex
\begin{wrapfigure}{R}{0.4\textwidth}
\vspace{-10pt}
\centering
\includegraphics[width=\linewidth]{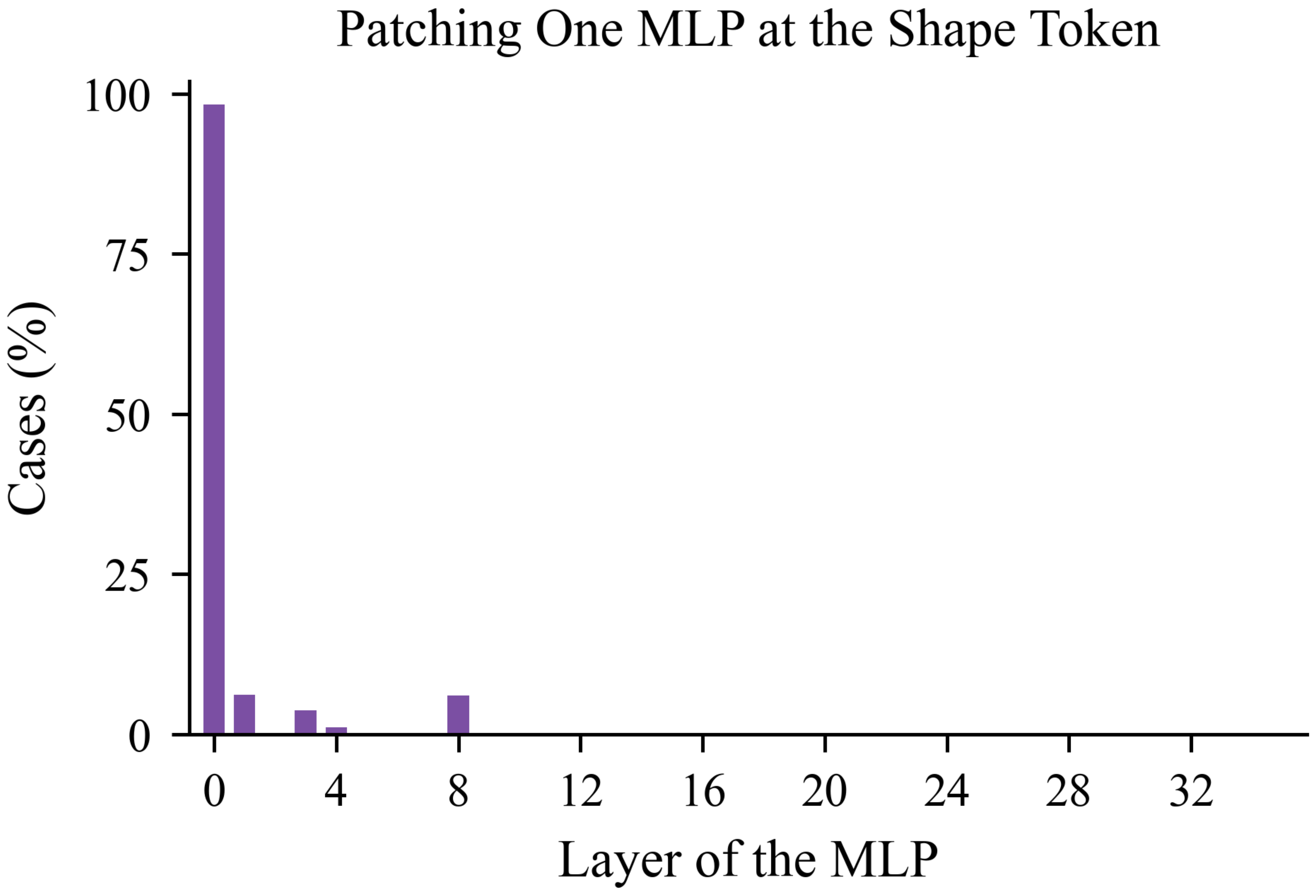}
\caption{\textbf{The first MLP writes the request.} We patch the output of one MLP at a time at the shape token. Patching the first MLP moves the request, \texttt{P}, in 98.3\pmv{0.8}\% of the held-out cases, and patching any other MLP moves it in at most 6.1\pmv{1.4}\%.}
\label{fig:mlp-sweep}
\vspace{-10pt}
\end{wrapfigure}

%% file: tables/appendix/other_bindings.tex
\begin{table}[ht]
\centering
\small
\caption{\textbf{The paper's heads work on other requests.} We use the head sets exactly as found on the shape task, with no new search. Each cell is the percent of held-out cases whose next token is the expected letter, and the first column repeats the paper's row images. We score word labels on the initial under the paper's question (``words'') and on the word under a question that asks for the word (``word asked''). Random heads from the same layers reach at most 4.0\% in any cell.}
\label{tab:other-bindings}
\fittable{%
\begin{tabular}{@{}lcccccc@{}}
\toprule
Patch (expected) & Shapes (paper) & Color & Category & Two attributes & Words & Word asked \\
\midrule
Binding queries (request) & 81.0\pmv{2.5} & 98.9\pmv{0.6} & 82.1\pmv{2.3} & 84.4\pmv{2.1} & 85.0\pmv{2.2} & 86.2\pmv{2.1} \\
Binding image keys (layout) & 83.1\pmv{2.3} & 99.8\pmv{0.2} & 97.5\pmv{1.0} & 81.5\pmv{2.4} & 80.9\pmv{2.4} & 86.1\pmv{2.1} \\
Queries and keys (pointer) & 95.0\pmv{1.5} & 100.0\pmv{0.0} & 99.9\pmv{0.1} & 93.9\pmv{1.5} & 87.8\pmv{2.1} & 97.0\pmv{1.0} \\
Binding outputs (pointer) & 85.9\pmv{2.4} & 96.9\pmv{1.1} & 96.9\pmv{1.1} & 93.8\pmv{1.5} & 75.4\pmv{2.8} & 85.8\pmv{2.2} \\
Relay values, from the binding-head patch (pointer) & 89.0\pmv{2.1} & 97.9\pmv{0.9} & 95.3\pmv{1.3} & 94.3\pmv{1.5} & 79.9\pmv{2.4} & 86.3\pmv{2.1} \\
\quad target's values restored (own) & 99.8\pmv{0.2} & 100.0\pmv{0.0} & 100.0\pmv{0.0} & 100.0\pmv{0.0} & 99.6\pmv{0.4} & 100.0\pmv{0.0} \\
Gaze queries patched (pointer) & 74.0\pmv{2.9} & 92.1\pmv{1.6} & 92.0\pmv{1.8} & 88.5\pmv{1.9} & 59.7\pmv{3.0} & 74.8\pmv{2.6} \\
Gaze image values (source's letter) & 93.7\pmv{1.6} & 99.6\pmv{0.4} & 99.8\pmv{0.2} & 99.1\pmv{0.6} & 87.9\pmv{2.1} & 1.4\pmv{0.8} (initial 15.8\pmv{2.2}) \\
Only the shape token and gaze heads see the image (own) & 95.1\pmv{1.3} & 100.0\pmv{0.0} & 99.9\pmv{0.1} & 99.8\pmv{0.2} & 99.6\pmv{0.4} & 16.1\pmv{2.3} (initial 79.9\pmv{2.4}) \\
Residual patch at the colon token, pointer at its peak (layer 24) & 78.6\pmv{2.5} & 95.3\pmv{1.4} & 91.6\pmv{1.8} & 93.4\pmv{1.4} & 46.1\pmv{3.1} & 74.2\pmv{2.7} \\
\bottomrule
\end{tabular}}
\end{table}

%% file: secs/appendix/C_relay.tex
\section{Relay: Details and Supporting Experiments}
\label{app:relay}

This appendix supports the relay heads of Section~\ref{sec:binding}, which carry the resolved pointer. We show which tokens they read, how their queries, keys, and values divide the work, what they carry on a list, and every blocking experiment on the way from the shape token to the colon token.

\subsection{Which Tokens Carry the Pointer to the Colon Token}
\label{app:one-sources}
\paragraph{The relay heads carry the state made at the shape token.} We first check that the relay heads move the pointer resolved at the shape token, and not something else in the source's prompt. To do this, we build a run with the \emph{binding-head patch}: the target with only the binding heads' outputs patched from the source at the shape token, and nothing else. Its prompt, its image, and every other state are the target's own. We then let the relay heads of an unedited target read the values of this run. The pointer moves in 89.0\pmv{2.1}\% of cases, as often as when they read the source's own values (87.5\pmv{2.0}\%). Conversely, if we keep the binding patch but let the relay heads read the target's original values, the target's answer comes back in 99.8\pmv{0.2}\%. So the relay heads carry the pointer resolved at the shape token. Figure~\ref{fig:appendix_relay}a shows the single-head search that found the 16 relay heads.

\input{figures/appendix/relay}

\paragraph{Each relay head, one source token at a time.} Figure~\ref{fig:relay_progress}a and Figure~\ref{fig:appendix_relay}b use the source's own values. To find which token the relay heads read the pointer from, we let one relay head at a time, at the colon token, read the source's value at a single text token after the image. Its query, keys, and attention weights stay the target's own. We repeat this for every text token before the colon token. Each cell is the mean gain of the logit of \texttt{K} over \texttt{T}, relative to the unedited target, on the held-out pairs. We draw the shape token and the question mark after it as one column, the sum of their two gains. When all 16 relay heads read the source's value at the shape token alone, the target answers \texttt{K} in 47.3\pmv{3.1}\% of cases. At the question mark alone it does so in 9.9\pmv{1.8}\%, and at every other token in at most 0.2\pmv{0.2}\%. So the relay heads read the pointer mostly from the shape token.

\subsection{Relay Heads: Further Checks}
\label{app:relay-checks}
\paragraph{Queries and keys route, values carry.} To see which part of the relay heads carries the pointer, we patch their queries, keys, and values at the colon token one at a time. Patching their values moves the pointer in 87.5\pmv{2.0}\% of cases. Patching their queries with the source's leaves the answer unchanged in 86.9\pmv{2.1}\% of cases, and patching the keys of the text they read leaves it unchanged in 99.2\pmv{0.6}\%. Patching the queries and keys together leaves it unchanged in 98.4\pmv{0.9}\%. So the queries and keys only route the read, and the values carry the pointer. The queries are the least clean of the three, because the late relay heads build their queries from a residual stream that already holds the pointer.

\paragraph{One panel, not the set.} On the list task, the relay heads could carry the source's first panel or the whole set of the source's panels. When the relay heads at the colon token read the source's values, the first answer moves to the source's first panel in 78.5\% of cases, against 0.0\% for random heads. To test whether they carry one panel or the set, we hide the source's first panel from every text token in every layer. If the set had traveled, the target would go on to the source's other panel. It does so in only 8.4\% of cases. It falls back to its own first panel in 31.9\%, and it goes to the panel after the hidden one in 14.7\%. It still reads the hidden panel in 30.9\%, since hiding a panel from the text is never complete. When every head reads the source's values instead, the counts are 48.7\%, 22.5\%, and 4.7\%. So the relay heads carry one panel, the first.

\subsection{Blocked Attention: All Conditions}
\label{app:one-knockouts}
Table~\ref{tab:knockouts-full} reports every blocking experiment in the target's own run, each with its control.

The pointer reaches the colon token by two routes. When we block only the colon token's attention to the shape token, the answer is kept in 99.7\pmv{0.4}\% of cases. When we block only the attention of the tokens in between to the shape token, it is kept in 90.6\pmv{1.8}\%. When we block both, it is kept in only 25.3\pmv{2.5}\%. So the pointer reaches the colon token both directly and through the tokens in between.

Binding is spread over more heads than the smallest set that can move the pointer. When we block the 16 binding heads and the 2 query and key heads outside them at the shape token from the image, the answer is kept in 84.9\pmv{2.1}\% of cases. When we block every head there from the image, it is kept in 40.4\pmv{3.1}\%. When we block the shape token and every later token from the image, 32.6\pmv{2.8}\% of answers survive, because the instruction tokens before the shape token also read the image.

When blocking breaks the pointer, the wrong answers lean toward the leftmost panel. In the two experiments that block the pointer, 39.8\pmv{2.3}\% of the wrong answers are the letter of the leftmost panel, against one in four by chance.

\input{tables/appendix/knockouts_full}

The tokens between the shape token and the colon token also carry the letter itself. This shows when we block the gaze heads of Section~\ref{sec:reading} from the image.

\paragraph{Who reads when the gaze heads are blocked from the image.} When the gaze heads cannot see the image, another set of heads might read the letter in their place. To look for them, we block the 32 gaze heads from the image at the colon token and run the single-head query-patch search again over all 1,152 heads. We then patch the top $K$ heads together, for $K$ from 1 to 256. No set moves the answer to the source's panel. They reach 0.0\% on the validation pairs at every $K$, and 3.1\% (top 32) and 5.6\% (top 128) on held-out cases, against 74.0\pmv{2.9}\% for the 32 gaze heads with nothing blocked. Blocking these heads from the image, on top of the 32 gaze heads, lowers the answer only from 88.2\pmv{2.0}\% to 82.1\% to 86.2\%. Blocking random heads instead gives the same (87.8\% to 92.9\%). So the backup is not another set of heads that read the image at the colon token.

The backup is an earlier read. The tokens between the shape token and the colon token read the letter from the image, and the colon token takes it from them. When we block every head at the colon token from the image, the answer survives in 98.5\% of cases. When we block the tokens between the shape token and the colon token from the image, it survives in 96.4\%. When we block both the 32 gaze heads at the colon token and these tokens from the image, it survives in only 2.6\%. Either route alone is enough. This is why the gaze heads are sufficient but not necessary.

%% file: figures/appendix/relay.tex
\begin{figure*}[t]
\centering
\includegraphics[width=\textwidth]{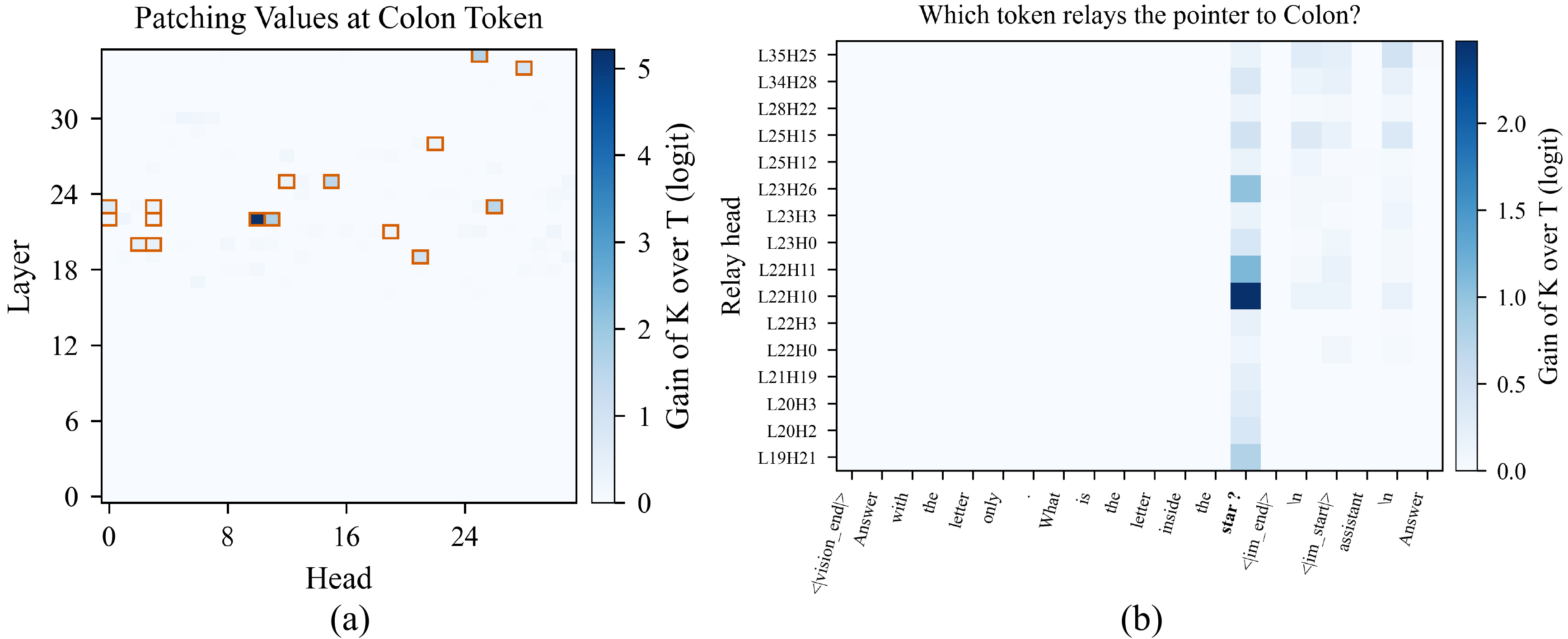}
\caption{\textbf{Relay heads carry the pointer from the shape token to the colon token.} (a)~At the colon token, we let one attention head at a time read the source's values from the shape token and the tokens after it, while its own query and keys stay unchanged. Each cell is the mean gain of \texttt{K} over \texttt{T} on the 200 discovery pairs. The 16 relay heads that the search selects are outlined, in layers 19-35. (b)~Each relay head alone reads the source's value at one text token only, while its query, keys, and attention stay unchanged. Each cell is the mean gain of \texttt{K} over \texttt{T} on the held-out pairs. When all 16 relay heads read only the shape token, the target answers \texttt{K} in 47.3\pmv{3.1}\% of cases, and when they read only the ``\texttt{?}'', it does so in 9.9\pmv{1.8}\%.}
\label{fig:appendix_relay}
\end{figure*}

%% file: tables/appendix/knockouts_full.tex
\begin{table}[ht]
\centering
\small
\caption{\textbf{Every blocking experiment on the single-answer task.} \emph{Letter} is the percent of cases that keep the target's correct answer. \emph{Gaze} is the percent of cases in which the published Gaze Heads stay on the correct panel. All values are on the 500 held-out pairs.}
\label{tab:knockouts-full}
\fittable{%
\begin{tabular}{@{}p{9.2cm}cc@{}}
\toprule
Attention blocked & Letter & Gaze \\
\midrule
Nothing blocked & 100.0\pmv{0.0}\% & 97.4\pmv{0.9}\% \\
The 10 query and key heads blocked from the image at the shape token & 85.3\pmv{2.2}\% & 81.6\pmv{2.5}\% \\
\quad same number of random heads & 100.0\pmv{0.0}\% & 97.7\pmv{0.9}\% \\
The 16 binding heads and the 2 query and key heads outside them, blocked from the image at the shape token & 84.9\pmv{2.1}\% & 74.7\pmv{2.6}\% \\
Every head blocked from the image at the shape token & 40.4\pmv{3.1}\% & 35.9\pmv{3.0}\% \\
Every head blocked from the image, from the shape token to the token before the colon token & 32.6\pmv{2.8}\% & 30.0\pmv{2.7}\% \\
\quad control: the text before the shape token blocked from the image & 99.3\pmv{0.5}\% & 99.6\pmv{0.4}\% \\
Only the shape token sees the image & 5.3\pmv{1.4}\% & 24.6\pmv{2.2}\% \\
Later tokens cannot attend to the shape token & 25.3\pmv{2.5}\% & 26.5\pmv{2.6}\% \\
\quad control: they cannot attend to the token before it & 99.9\pmv{0.1}\% & 97.5\pmv{1.0}\% \\
\quad only the colon token cannot attend to it & 99.7\pmv{0.4}\% & 97.2\pmv{1.0}\% \\
\quad only the tokens in between cannot attend to it & 90.6\pmv{1.8}\% & 92.4\pmv{1.7}\% \\
Relay heads cannot attend to the text at the colon token & 7.8\pmv{1.6}\% & 19.1\pmv{2.3}\% \\
\quad same number of random heads & 99.8\pmv{0.2}\% & 97.9\pmv{0.9}\% \\
\bottomrule
\end{tabular}}
\end{table}

%% file: secs/appendix/D_progress.tex
\section{Progress: Details and Supporting Experiments}
\label{app:progress}

This appendix supports Section~\ref{sec:progress-select}. We describe the pairs of the progress experiments, the state the comma token holds before the cursor forms, and the heads that set the cursor at the answer token. We then show the two routes from the answer token to the comma token, and we write the cursor with no source as a small subspace. Appendix~\ref{app:select} gives the details of select. Figure~\ref{fig:appendix_mechanism_multi} draws the whole mechanism on a list with two answers.

\input{figures/appendix/appendix_mechanism_multi}

\subsection{Implementation Details}
\label{app:progress-implementation}
The three experiments of Sections~\ref{subsec:progress} to~\ref{subsec:progress-source} use their own pairs. We draw them at random with the images and prompt of Appendix~\ref{app:many-implementation}. The source and the target ask for different shapes, drawn from star, heart, and triangle. The non-matching panels hold shapes the model does not confuse with the asked one, and the twelve letters of a pair are all different. The design fixes the matching panels of each experiment. In the skip experiment, the source's written panel is the target's first matching panel, drawn from panels 1, 2, or 3. In the cursor experiment, the source has its shape at panels 2, 4, and 5 and has written panels 2 and 4, and the target has its shape at panels 1, 2, 3, and 6. In the rebind experiment, the target has its shape at panels 1, 3, 4, and 6 and has written panels 1 and 3. We keep a pair only if both unedited runs give the expected next letter, and we run each experiment on held-out pairs. The residual patch copies the source's comma-token residual into the target's colon token at one layer, for every layer from 0 to 36, and reads the target's next token. The rebind patch is the key patch of Appendix~\ref{app:one-implementation}, applied at an answer token with the source image's keys.

\subsection{The State Before the Cursor}
\label{app:progress-state}
Section~\ref{subsec:progress-cursor} finds that a comma token's residual copied into a colon token changes the target in three stages. In the early layers the target skips one matching panel. In the middle layers it counts the written letters. From layer 20 it starts after the last written panel. We use three runs to ask what the early state holds.

\paragraph{The early state says ``one letter was written'', whichever letter it was.} We take the held-out targets of the list task and run each image twice more, once with $r_1$ written and once with $r_2$ written. We then copy the residual of that comma token, at one layer, into the colon token of the run with nothing written (Figure~\ref{fig:progress-state}a). With the source that wrote $r_1$, the target answers $r_2$ in 65.8\% of cases already at layer 14, and in 96.4\% at layer 21. With the source that wrote $r_2$, the target also answers $r_2$ in 75.1\% of cases at layer 19, and $r_3$ in only 2.1\%. From layer 23 it answers $r_3$ in 82.4\% of cases, rising to 92.7\% at 32. So up to layer 19 the copied state says only that one letter has been written, and the target skips its first matching panel whichever letter that was. From layer 20 the state says which panel was written, and the target starts after it.

\paragraph{The early skip is tied to panel 1.} In the skip experiment of Section~\ref{subsec:progress}, the source's written panel is the target's first matching panel, and it is panel 1, 2, or 3. We split the pairs by that panel (Figure~\ref{fig:progress-state}b). In the early layers, the target skips its first matching panel in 80.6\pmv{4.5}\% of pairs when the shared panel is panel 1, in 8.6\pmv{3.3}\% when it is panel 2, and in 4.7\pmv{2.4}\% when it is panel 3. At layer 20 the three groups skip in 99.5\pmv{0.8}\%, 97.5\pmv{2.2}\%, and 74.2\pmv{7.1}\% of pairs. At layer 21 they skip in 53.2\pmv{6.7}\%, 72.0\pmv{7.4}\%, and 88.4\pmv{5.1}\%. So the early lean falls on panel 1 and not on a count of matching panels. The cursor forms at layer 20 for panels 1 and 2, and one layer later for panel 3.

\paragraph{Count, then cursor, then the source's next pointer.} We use the standard source of Appendix~\ref{app:many-implementation}, which has the asked shape at $r_2$ and $r_4$ and has written the target's first letter at $r_2$. Here a count predicts $r_2$, the written panel predicts $r_3$, and the source's own next panel predicts $r_4$ (Figure~\ref{fig:progress-state}c). The target answers $r_2$ in 86.5\% of cases at layer 19. It answers $r_3$ in 68.4\% at layer 20 and 53.9\% at 21, and $r_4$ in 77.7\% at layer 22 and 87.6\% at 24. At layer 30 it answers with the source's letter in 99.5\%. This is the same order as in the cursor experiment, on a pair whose source has written one letter.

\input{figures/appendix/progress_state}

\paragraph{The copy itself does nothing, and the state needs a real letter.} We run two controls on the skip design. First, we copy the colon token of a source that has written nothing into the target's colon token. The target keeps its own answer in 98.6 to 100.0\% of pairs at every layer through 29, so patching a comma-token residual into a colon token is not by itself what moves the answer. Second, we prefill the source with a letter that is drawn in neither image. This source makes the target skip the shared panel in 53.8\pmv{4.4}\% of pairs at layer 20, against 91.0\pmv{2.4}\% for the real source on the same pairs. The skip happens mostly when the shared panel is panel 1 (96.8\pmv{2.4}\%), and less for panel 2 (27.6\pmv{7.6}\%) and panel 3 (29.0\pmv{6.8}\%). So this source carries only the early lean, not a panel. The cursor needs a letter that is in the image.

\subsection{Other Searches and Patches at the Answer Token}
\label{app:many-mark}
\paragraph{The answer-token searches.} We search over every head at the answer token, as we do at the shape token (Figure~\ref{fig:label-screens}). We score each of the 1,152 heads on the discovery pairs of the list task in three ways. The first score is the gain in progress when the head reads the source's image keys. The second is the gain when the head's output is the source's. The third, in the target's own run, is the gain toward repeating the written letter when the head's output is set to zero. The heads that move progress sit in the layers of the binding heads. Zeroing single heads at the answer token pushes the model toward writing the letter again, and the top heads of that search sit in layers 15-34. The same zeroing search at the comma token finds only late heads, in layers 19-35, and none of them is a binding head. So the comma token reads the cursor and does not write it.

\input{figures/appendix/label_screens}

\paragraph{The key patch with one letter written.} We let the 32 heads that the key-patch search selects at the answer token read the source's image keys. The next answer then moves to the letter after the source's panel in 73.8\pmv{3.8}\% of held-out cases, against 0.0\pmv{0.0}\% for random heads. Every new search on the list task finds mostly heads of the matching set of Sections~\ref{sec:binding} and~\ref{sec:reading} (Table~\ref{tab:many-heads}). Of the 32 heads found at the shape token, 15 are binding heads. Of the 16 found at the colon token, 10 are relay heads, and of the 16 found at the comma token, 15 are gaze heads. The 64 heads found at the last letter mix 1 binding, 10 relay, 6 gaze, and 0 language heads.

\input{tables/appendix/many_heads}

\paragraph{All conditions of the rebind.} Figure~\ref{fig:appendix_rebind} gives every condition of the rebind in Section~\ref{subsec:progress-source}. Rebinding the last answer token to panel 4 moves the next answer past panel 4 with the 32 answer-token heads (88.0\pmv{2.8}\%), the 16 binding heads (84.2\pmv{3.2}\%), and all 1,152 heads (62.2\pmv{4.4}\%). It never does so with 16 random heads from the binding heads' layers. In the rest of the pairs the target still answers from panel 4 (9.8\pmv{2.6}\%, 13.4\pmv{3.0}\%, and 35.6\pmv{4.2}\%). In any condition, the target repeats a written letter in at most 2.4\pmv{1.3}\% of pairs. Conversely, rebinding the first answer token moves the answer in at most 3.6\pmv{1.6}\% of pairs, and leaves it at panel 4 in 96.4\pmv{1.6}\% to 98.8\pmv{0.9}\%.

\input{figures/appendix/rebind}
\paragraph{The cursor is a panel, not a letter.} If the cursor were the written letter rather than its panel, the model would also skip a second copy of that letter in the image. To test this, we put the written letter into $r_3$ as well, so that after ``$W$, $r_2$,'' the next matching panel holds $W$ again. The model writes $W$ again in 95.8\pmv{1.7}\% of lists, against 96.6\pmv{1.6}\% when $r_3$ holds a new letter. The letter adds only a weak pull. What the model crosses off is the panel.

\paragraph{The answer token needs nothing from the question.} When we block the answer token's attention to the shape token, the answer token still sets the cursor: the next answer is unchanged in 98.0\pmv{1.2}\% of cases. When we block the comma token's attention to the shape token instead, 73.2\pmv{3.8}\% of answers survive. The comma token also uses the shape token to find the matching panels (Appendix~\ref{app:many-again}).

\subsection{Two Routes From the Answer Token to the Comma Token}
\label{app:progress-relay}
The comma token needs the cursor. When we block the last comma token's attention to the answer token, the model writes the letter again in 67.8\pmv{4.0}\% of cases. Blocking its attention to the colon token instead leaves 99.8\pmv{0.3}\% of answers unchanged (Table~\ref{tab:many-knockouts}). The comma token can learn where the last letter came from in two ways. It can read the answer token, which has bound the letter to its panel, or it can find the letter in the image itself. Either way is enough on its own. With the answer token blocked from the image, the next answer is kept in 91.4\pmv{2.4}\% of cases. With the comma token blocked from the image, it is kept in 98.6\pmv{1.0}\%. With both blocked from the image, it is kept in only 18.2\pmv{3.4}\%.

The first route carries the cursor, and it is spread over many heads. We patch the source's values of the answer token, as every head at the comma token reads them. While the comma token can still look at the image, this moves the next letter in only 17.0\% of cases. Once we block the comma token from the image, except through the gaze heads, the same patch moves the answer off the target's own panel in 99.0\% of cases. It goes to the source's cursor in 40.7\% and to the source's own next panel in 58.2\%. Giving the comma token the source's image keys instead moves it in 10.8\%. The comma token reads the answer token in two ranges of layers, and either range is enough. With this attention blocked in layers 0-24 the answer survives in 96.9\% of cases, and with it blocked in layers 20-35 it also survives in 96.9\%. Blocking the attention of the 64 highest-ranked heads to the answer token gives 16.5\% repeats, against 0.0\% for 64 random heads. We also run a single-head search for this value patch, scored with the comma token blocked from the image in the same way. It selects 8 heads that do not reach the 80\% rule. On the held-out cases, these heads move the cursor in 63.9\% with the comma token blocked (random heads 11.9\%), and in 5.7\% without. Two of the eight are relay heads, and none is a gaze head. So this value patch finds no small set of heads.

\input{tables/appendix/many_knockouts}

\paragraph{What looks like a count is the record of earlier panels.} A target that has written only $r_3$ answers $r_2$ in 34.4\pmv{4.2}\% of cases and $r_4$ in most of the rest. We take a run on the same image that wrote $r_1$, $r_2$, $r_3$. Copying the residual of its comma token at layer 22 makes the target answer $r_4$ in 98.5\% of cases. Letting the relay heads read the values of its third letter does the same in 89.3\% (all heads 81.2\%). So the last written token carries the earlier panels, because it read the earlier answer tokens.

\subsection{The Cursor Is a Small, Writable Subspace}
\label{app:cursor-subspace}
Section~\ref{subsec:progress-cursor} shows the cursor by copying a source's comma token. Here we write the cursor with no source, and ask how many dimensions it takes.

\paragraph{Directions.} On development images we prefill the model's own letters up to a matching panel $p$, for every $p$, and record the residual of the last comma token at layers 19 to 22. The direction for $p$ is the mean of these residuals minus the mean over all $p$. The six directions have rank 5 at every layer. Neighboring $p$ are similar and distant $p$ are opposed (cosine 0.67 between panels 1 and 2, $-0.73$ between 1 and 5). We also record the mean comma-token residual and the mean colon-token residual with nothing written. Their difference, the ``comma offset'', has norm 53 at layer 20, larger than any direction.

\paragraph{Writing the cursor.} On fresh images with nothing written, we add the direction for $p$ to the colon-token residual at layer 21 and read the next token. An image counts for $p$ when a matching panel lies after $p$. The direction alone does nothing: 0.0 to 2.2\% of images start after $p$, the same as with a random direction. The colon token is missing the comma state. Adding the comma offset alone moves every target to a fixed position near panel 4 (mean chosen panel 4.2 to 4.6 for every $p$). Adding the offset plus the direction for $p$ tracks $p$. The target then starts at the first matching panel after $p$ in 97.7\pmv{1.6}\%, 96.3\pmv{1.7}\%, 96.8\pmv{1.5}\%, 96.1\pmv{1.8}\%, and 96.5\pmv{1.9}\% of images for $p=1$ to 5, with the mean chosen panel at 2.49, 3.42, 4.38, 5.26, and 5.95. The offset plus a random direction of the same norm gives 8.8\%, 38.0\%, 63.3\%, 48.3\%, and 15.7\% (mean of five draws). Replacing the colon-token residual by the mean comma state for $p$ gives the same numbers. A direction taken from layer 10 does nothing, and one from layer 30 overshoots.

\paragraph{Five dimensions.} Let $S$ be the span of the six directions. We use the cursor design of Section~\ref{subsec:progress-cursor} and a second design in which the source has written panel 2 only, so that the cursor predicts panel 3 while the mean comma state predicts panel 6. On both, we patch only the component of the source's comma token inside $S$, only its complement, or both. The $S$ component alone, written into a colon token, changes nothing (the own answer in 100.0\pmv{0.0}\%), because the comma offset lies mostly outside $S$. Once the rest of the residual is comma-like, $S$ decides the target. The source's $S$ part with the mean's complement gives the source's cursor in 68.8\pmv{4.0}\% of cases at layer 21. Conversely, the source's complement with the mean's $S$ part gives the mean's target in 96.0\pmv{1.7}\%. The full patch gives the source's cursor in 87.4\pmv{2.8} to 96.6\pmv{1.5}\%. A random five-dimensional subspace moves nothing (0.0\%). So the cursor is a five-dimensional code inside the comma-token residual, and the rest of that residual says only that the token is a comma. Appendix~\ref{app:cursor-layouts} tests these directions on other layouts.

%% file: figures/appendix/appendix_mechanism_multi.tex
\begin{figure*}[ht]
\centering
\includegraphics[width=0.92\textwidth]{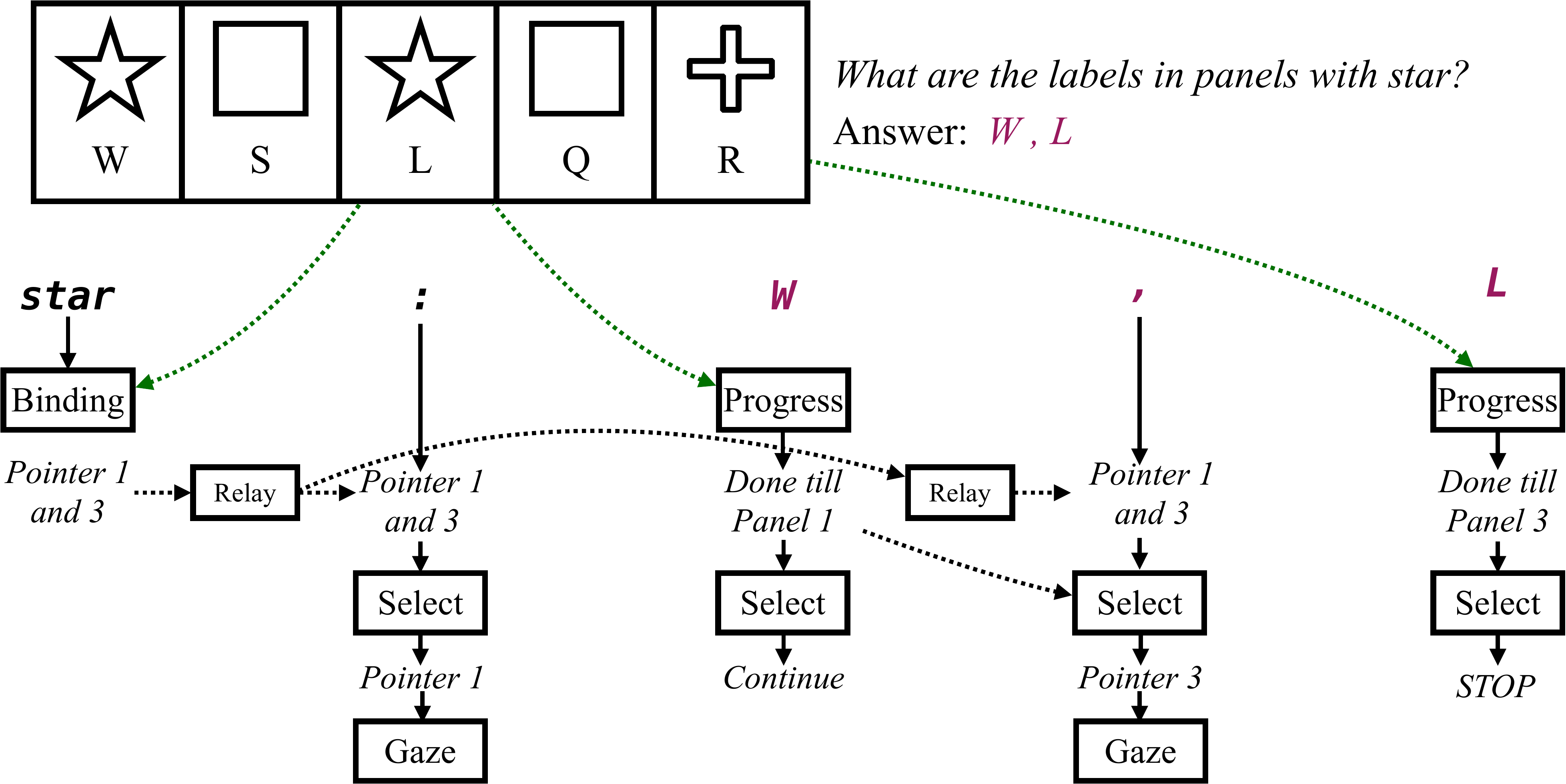}
\caption{\textbf{The mechanism on a list with two answers.} The image has stars in panels 1 and 3, and the model answers \texttt{W, L}. At the shape token \texttt{star}, binding heads write pointers to panels 1 and 3, and relay heads carry them to the colon token. There, select picks panel 1, and gaze heads read its letter, \texttt{W}. At the answer token \texttt{W}, progress records that the list is done up to panel 1. A matching panel is left, so the list continues. At the comma token, relay heads bring the pointers to panels 1 and 3 again, select uses the progress to pick panel 3, and gaze heads read \texttt{L}. At the answer token \texttt{L}, progress is done up to panel 3. No matching panel is left, so the model stops.}
\label{fig:appendix_mechanism_multi}
\end{figure*}

%% file: figures/appendix/progress_state.tex
\begin{figure*}[t]
\centering
\includegraphics[width=\textwidth]{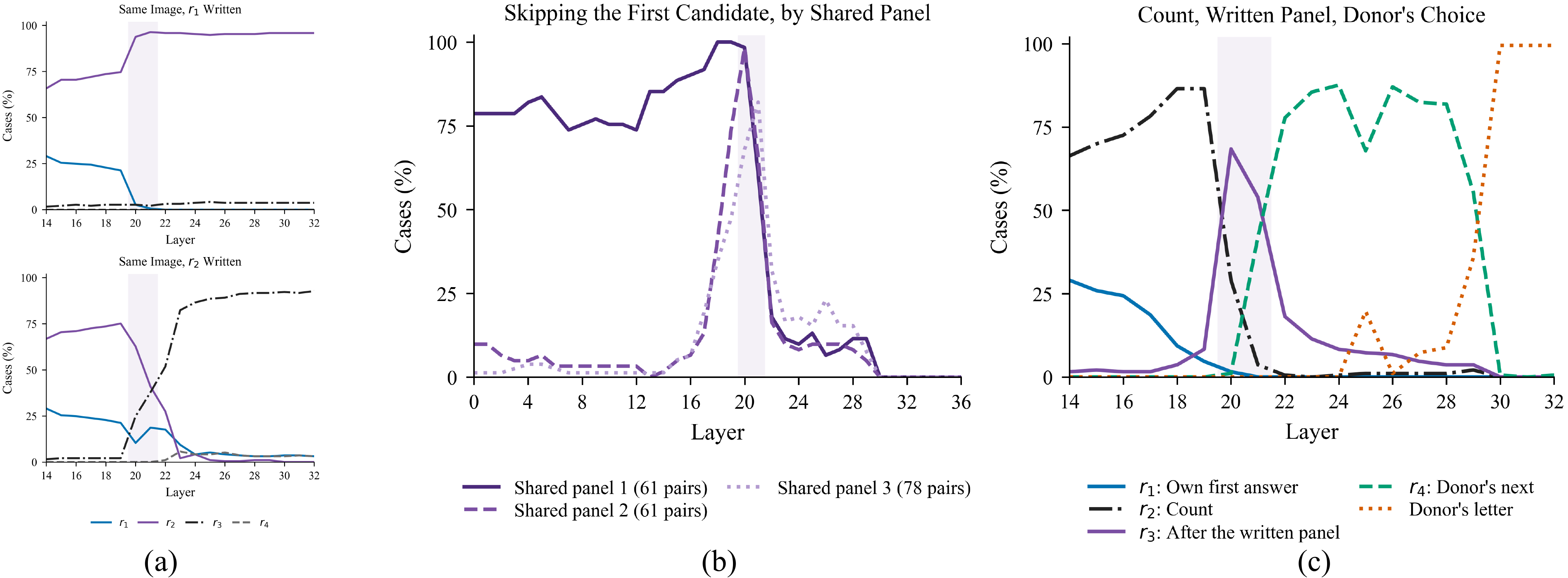}
\caption{\textbf{The state the comma token holds before the cursor forms.} We copy the residual of a comma token, at one layer, into a colon token that has written nothing, and read the target's next letter. (a)~The same image, with $r_1$ written (top) and with $r_2$ written (bottom), on the held-out targets. With $r_1$ written the target answers $r_2$ in 65.8\% of cases at layer 14 and 96.4\% at layer 21. With $r_2$ written it also answers $r_2$, in 75.1\% at layer 19, and answers $r_3$ from layer 23 (82.4\%). Up to layer 19 the copied state says only that one letter was written. From layer 20 the target starts after the written panel. (b)~The skip experiment of Section~\ref{subsec:progress}, split by the shared first matching panel. Averaged over layers 0-17, the target skips it in 80.6\pmv{4.5}\% of pairs when the shared panel is panel 1, 8.6\pmv{3.3}\% for panel 2, and 4.7\pmv{2.4}\% for panel 3. At layer 20 it skips in 99.5\pmv{0.8}\%, 97.5\pmv{2.2}\%, and 74.2\pmv{7.1}\%, and at layer 21 in 53.2\pmv{6.7}\%, 72.0\pmv{7.4}\%, and 88.4\pmv{5.1}\%. (c)~The standard source, which has written the target's first letter at $r_2$ and has its own next matching panel at $r_4$. The target answers with a count, $r_2$, at layer 19 (86.5\%), with the panel after the written one, $r_3$, at layer 20 (68.4\%), with the source's own choice, $r_4$, at layer 24 (87.6\%), and with the source's letter from layer 30. The shaded band marks layers 20-21.}
\label{fig:progress-state}
\end{figure*}

%% file: figures/appendix/label_screens.tex
\begin{figure*}[t]
\centering
\includegraphics[width=\textwidth]{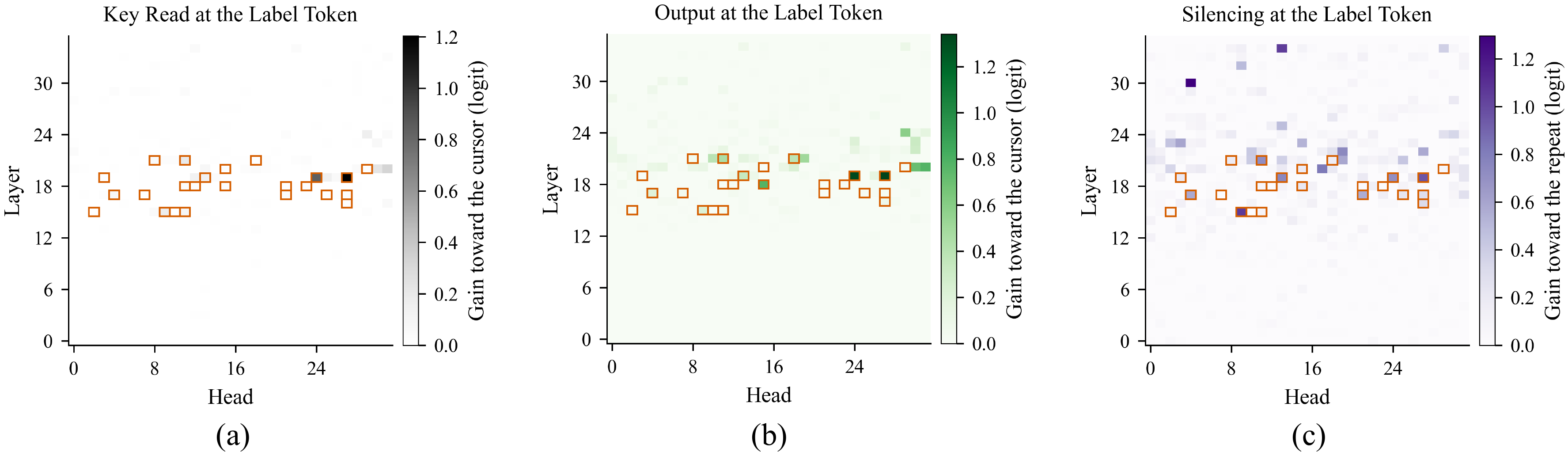}
\caption{\textbf{The heads that set the cursor at the answer token sit in the layers of the binding heads.} We patch one head at a time at the answer token of the list task. In (a), the head reads the source's image keys, and in (b), we replace the head's output with the source's. Each cell is the mean gain of the cursor's letter on the discovery pairs. In (c), we set the head's output to zero in the target's own run. Each cell is the mean gain toward repeating the written letter on the discovery pairs. The 16 binding heads are outlined.}
\label{fig:label-screens}
\end{figure*}

%% file: tables/appendix/many_heads.tex
\begin{table}[ht]
\centering
\small
\caption{\textbf{Each patch on the list task, with the heads of Sections~\ref{sec:binding} and~\ref{sec:reading} and with the heads a new search finds.} Each row is one patch on the list task. Each cell is the percent of held-out list cases in which the patch moves the expected variable.}
\label{tab:many-heads}
\fittable{%
\begin{tabular}{@{}lccc@{}}
\toprule
Patch & Heads of Sections~\ref{sec:binding} and~\ref{sec:reading}, unchanged & Random heads & Heads found here ($K$) \\
\midrule
First answer: values read at the colon token & 78.5\% & 0.0\% & 83.0\pmv{3.2}\% (16) \\
First answer: shape token outputs & 53.4\% & 0.5\% & 74.8\pmv{4.0}\% (32) \\
First answer: shape token image key reads & 9.9\% & 0.0\% & 10.8\pmv{2.8}\% (128) \\
First answer: colon token queries & 46.1\% & 0.0\% & 57.8\pmv{4.2}\% (128) \\
Last comma token: queries (next panel moves) & 88.2\pmv{2.8}\% & 0.0\% & 86.0\pmv{2.9}\% (16) \\
\bottomrule
\end{tabular}}
\end{table}

%% file: figures/appendix/rebind.tex
\begin{wrapfigure}{R}{0.4\textwidth}
\vspace{-10pt}
\centering
\includegraphics[width=\linewidth]{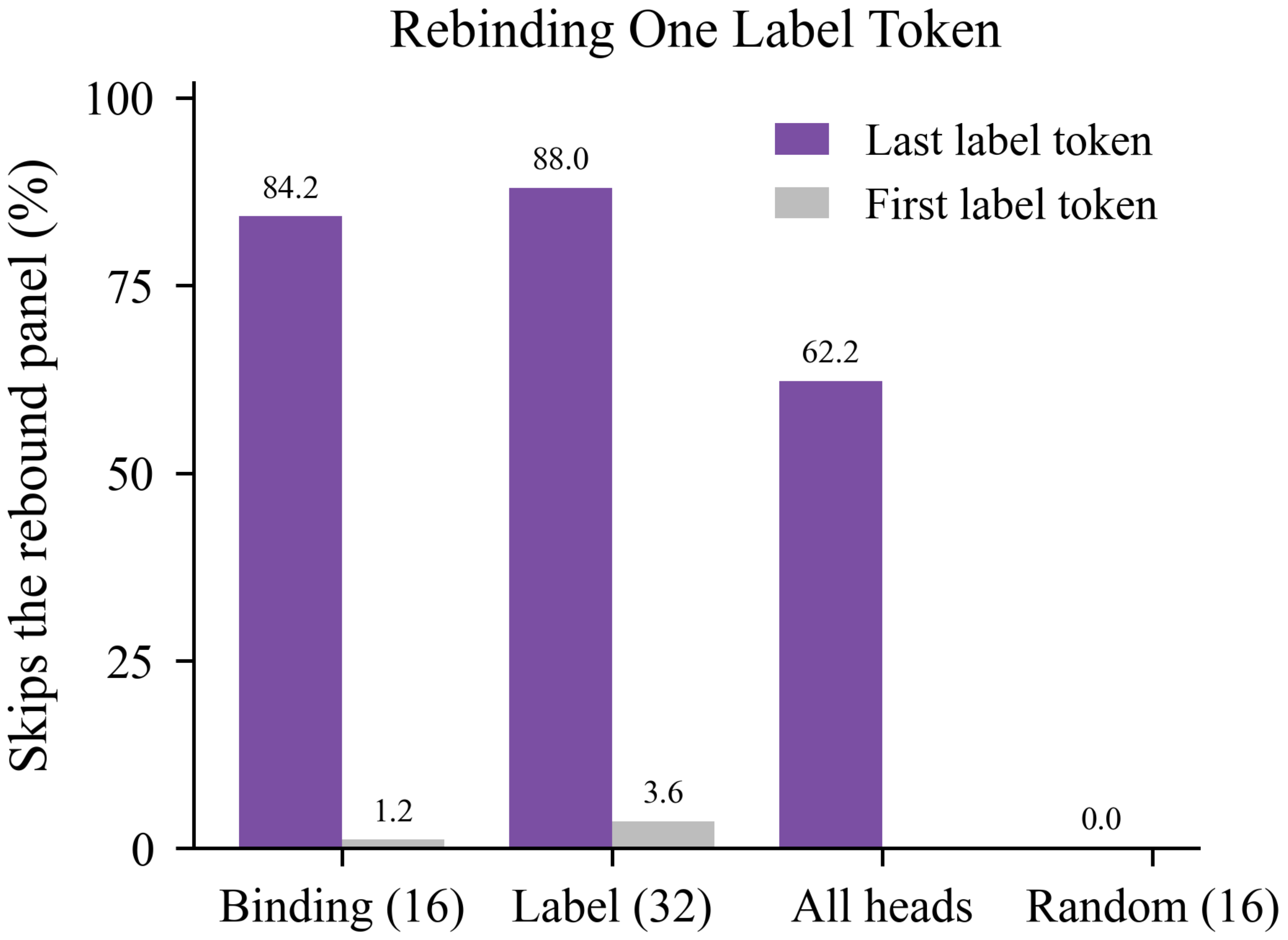}
\caption{\textbf{The cursor is set by the last answer token.} Pairs that skip the rebound panel, for each head set. We test the random heads at the last answer token only.}
\label{fig:appendix_rebind}
\vspace{-10pt}
\end{wrapfigure}

%% file: tables/appendix/many_knockouts.tex
\begin{table}[ht]
\centering
\small
\caption{\textbf{Blocking attention in the target's own run, at the second answer.} We block the attention in each row and read the next letter. Percent of held-out cases with one letter written.}
\label{tab:many-knockouts}
\fittable{%
\begin{tabular}{@{}p{9.4cm}cc@{}}
\toprule
Blocked & Own answer kept & Repeats the written letter \\
\midrule
Nothing blocked & 100.0\pmv{0.0}\% & 0.0\pmv{0.0}\% \\
Shape token blocked from the image & 68.8\pmv{3.8}\% & 0.0\pmv{0.0}\% \\
Later tokens cannot attend to the shape token & 66.4\pmv{4.0}\% & 0.0\pmv{0.0}\% \\
\quad control: later tokens cannot attend to the token before the shape token & 100.0\pmv{0.0}\% & 0.0\pmv{0.0}\% \\
Relay heads cannot attend to the text at the comma token & 65.6\pmv{4.0}\% & 0.0\pmv{0.0}\% \\
\quad random heads cannot attend to the text at the comma token & 100.0\pmv{0.0}\% & 0.0\pmv{0.0}\% \\
Answer token blocked from the image & 91.4\pmv{2.4}\% & 5.2\pmv{1.9}\% \\
Comma token blocked from the image & 98.6\pmv{1.0}\% & 0.0\pmv{0.0}\% \\
Answer token and comma token blocked from the image & 18.2\pmv{3.4}\% & 0.0\pmv{0.0}\% \\
Comma token cannot attend to the answer token & 27.4\pmv{3.9}\% & 67.8\pmv{4.0}\% \\
\quad control: comma token cannot attend to the colon token & 99.8\pmv{0.3}\% & 0.0\pmv{0.0}\% \\
Only the shape token, the answer token, and the gaze heads at the comma token see the image & 86.8\pmv{2.8}\% & 0.0\pmv{0.0}\% \\
Only the shape token and the gaze heads at the comma token see the image & 43.8\pmv{4.3}\% & 0.0\pmv{0.0}\% \\
\bottomrule
\end{tabular}}
\end{table}

%% file: secs/appendix/E_select.tex
\section{Select: Details and Supporting Experiments}
\label{app:select}

This appendix supports Section~\ref{subsec:select}. We give the rule that select follows and show how the model stops. We then show how the colon and comma tokens find the matching panels again, and what we can read from the comma token's residual with no patch. Finally, we hide a panel from a layer on, test the read that refuses a written panel, and describe the searches for selection heads.

\subsection{The Rule Select Follows, and Stopping}
\label{app:select-checks}
\input{figures/appendix/many_probe}
\paragraph{Out of order.} To see which panel select picks, we write letters out of order, with no patch, and read the next letter of the held-out targets (Figure~\ref{fig:appendix_many_probe}). With only the second heart written, the model answers with the third in 92.8\% of lists, and goes back to the first in 2.6\%. With the third written, it answers with the fourth in 64.4\% and goes back to the second in 34.4\%. With the third and then the first written, it answers with the second in 95.6\%. With the second and then the first written, it answers with the third in 87.2\%. Next, we block the comma token's attention to one of the written letters, on the lists whose next letter is right with nothing blocked (Table~\ref{tab:mechanism-order}). When the comma token cannot attend to an earlier written letter, that panel is written again in 52.8\pmv{4.6}\% of cases after ``$r_2$, $r_1$'', and in 57.0\pmv{4.5}\% after ``$r_3$, $r_2$, $r_1$''. The rest still skip it, so the later answer tokens carry part of the record forward. Select starts at the cursor and takes the nearest unwritten panel, preferring the right. Every answer token contributes to what counts as done.

\input{tables/appendix/mechanism_order}

\paragraph{Stopping.} At the last letter, the model chooses between a comma and the end of its turn. A source that has written its only matching panel stops, and a target with matching panels left goes on. We patch the outputs of the 64 heads found at the last letter. This makes a finished list go on in 100.0\pmv{0.0}\% of cases, and a continuing list stop in 96.8\pmv{1.5}\%, against 0.2\pmv{0.3}\% and 0.0\pmv{0.0}\% for the same number of random heads. The values these heads read make a continuing list stop in 88.2\pmv{2.8}\%. The binding heads' outputs from a run that still has matching panels make a finished list go on in 95.9\% of cases, and the relay heads' values do so in 84.5\%. No small set of the earlier heads makes a continuing list stop. In the target's own run, a finished list goes on when the later tokens cannot attend to the shape token (99.8\pmv{0.3}\%). It also goes on when the relay heads cannot attend to the text at the last letter (69.2\pmv{4.1}\%, random heads 0.0\pmv{0.0}\%). A continuing list never stops, whichever attention we block. After it goes on, a finished list adds only letters of its own image. The 64 heads found at the comma token for the next panel (Appendix~\ref{app:select-heads}) make a finished list go on in 99.5\% of cases, but random heads do so in 29.5\%. These heads make a continuing list stop in 33.7\%, against 0.0\% for random heads. On the single-answer task, with the letter written after \texttt{Answer:}, the model ends its turn in 100.0\pmv{0.0}\% of runs, whichever attention we block. There, the instruction decides the stop. Only the list task makes the model decide from the image and its own words.

\subsection{When the Model Skips a Panel, the Last Pointer Is Still Right}
\label{app:select-errors}
The model makes few errors on six panels. So we let it write whole lists on longer strips, with ten panels and seven matching panels, and read its states without any patch. It writes 75.8\% of these lists without an error. The most common error is a skip: the model passes over the next matching panel and answers from a later one. 3.2\% of answers are skips, and 86.5\% of the skips pass over a single matching panel. The other errors are rare. The model answers from a non-matching panel in 1.3\% of answers, goes back to an earlier unwritten panel in 0.5\%, repeats a written letter in 0.1\%, and never stops early.

We compare the skips that follow at least one written letter with correct answers at the same position in the list. At a skip, the earlier steps of the mechanism are intact. The binding heads at the last answer token put their strongest attention on the panel of the written letter in 100.0\% of skips, as in correct answers. A probe on the residual at layer 21 and the comma token reads the last written panel in 92.1\% of skips, against 97.5\% of correct answers. The error comes at select. At a skip, the gaze heads at the comma token put their strongest attention on the correct next panel in only 25.4\% of cases, against 95.6\% for correct answers. In 68.3\% of skips, they look at the panel the model then writes. A non-matching panel lies between the last written panel and the correct next one in 71.4\% of skips, against 28.2\% of correct answers. So at a skip, the pointer is right and the choice is wrong. Skips happen most often when the next matching panel is not the neighbor of the last one. These are the model’s own errors on our synthetic list task. We do not test repetition or early stopping on natural images and leave it for future work.

\subsection{The Answer Token Finds the Candidates Again}
\label{app:many-again}
The colon and comma tokens do not only take the matching panels from the shape token. Each of them also binds the question to the image again by itself.

\paragraph{The first answer.} On the list task, the relay heads are not the only path. We patch the target's residual vector at one token and one layer, for every token after the image. This finds the same two tokens as on the single-answer task, the shape token and the colon token. When the relay heads cannot attend to the text at the colon token, the first answer survives in 66.8\pmv{4.1}\% of cases, against 7.8\pmv{1.6}\% on the single-answer task. The colon token of the list prompt binds the question to the image itself. With the shape token blocked from the image, the first answer survives in 67.2\pmv{4.1}\% of cases. With the colon token blocked from the image, it survives in 99.4\pmv{0.7}\%. So either token can bind the question on its own, and the shape token's route is the stronger one.

\paragraph{The next answer.} We build a source with the asked shape at $r_1$ and $r_3$ and the same written letter at $r_1$. It differs from the target only in the set of matching panels, and this moved set makes the target skip $r_2$. Patching the source's residual vector into the shape token, at the best layer, does this in 14.8\pmv{3.0}\% of cases. Patching it into the comma token does so in 78.4\%. Replacing the outputs of the 32 heads found at the shape token does so in 21.0\pmv{3.4}\% of cases, against 0.0\pmv{0.0}\% for random heads. The same patch does so in 64.5\pmv{4.1}\% when the comma token keeps only the gaze heads' view of the image. It does so in 96.8\pmv{1.5}\% when the answer token and the comma token also read the source's image keys. So the comma token's fresh look usually wins. In the target's own run, blocking the comma token from the image keeps 91.8\pmv{2.4}\% of answers right. Blocking every later token's attention to the shape token keeps 66.5\pmv{4.0}\%.

The fresh look already knows what is done. We use the panel directions learned at the shape token (Appendix~\ref{app:binding-tokens}). In these directions, the binding heads' contribution at the answer token scores the written panel at $+0.50$ and the next panel at $-0.09$. The next panel scores below the written panel in 85.6\pmv{4.4}\% of held-out pairs, so a written panel has the sign of a matching panel when it is recorded. The same directions score the comma token's contribution to the written panel at $-0.33$ and to the next panel at $+0.20$. The written panel scores below the non-matching panels in 66.0\pmv{5.6}\% of pairs, and below the other matching panels in 89.2\pmv{4.0}\%. So the answer token records the panel, and the comma token subtracts it.

\subsection{Reading the Comma Token's State}
\label{app:select-readout}
\paragraph{Probes by layer.} We ask what the comma token's residual holds with no patch. On the held-out pairs of the list task, we record the residual of the last comma token at every layer, fit a ridge probe on half of them, and read it on the other half (Figure~\ref{fig:select-readout}a). A probe for ``panel $p$ has the asked shape'' reads each panel with 95.9\pmv{1.0}\% accuracy at layer 17 and 99.3\pmv{0.5}\% at 21, from a chance level of two thirds. A probe for ``which panel was written'' names the written panel in 94.0\pmv{2.8}\% of pairs at layer 12 and 100.0\pmv{0.0}\% at 18. A probe for ``which panel is next'' names the next panel in 98.0\pmv{1.8}\% at layer 18 and 100.0\pmv{0.0}\% at 20. In its centered scores, the written panel sits below the other matching panels at every layer. The flat starts of the curves are the majority baselines, since the first panels of a list are usually the leftmost ones. The colon token behaves the same way with nothing written, and its matching-panel probe peaks at layers 20-23.

\input{figures/appendix/restore}
\paragraph{Attention by layer.} We read the recorded attention of the same runs. At layer 21, the binding heads at the comma token put 51.6\pmv{1.9}\% of their image attention on the next panel and 3.9\pmv{0.2}\% on the written panel. At layers 15-19, their shares on the written panel, the next panel, and the other matching panels are all small and alike. At layer 20, the gaze heads give 22.3\pmv{0.6}\% to the next panel and 23.0\pmv{0.7}\% to each other matching panel. By layer 23, these shares are 61.9\pmv{1.4}\% and 7.4\pmv{0.4}\% (Figure~\ref{fig:select-readout}b). At the shape token, the binding heads put 79.6\pmv{0.2}\% of their image attention on the matching panels, and 27.4\pmv{0.8}\% of that on the one chosen first. At the colon and comma tokens, the gaze heads put 43.8\pmv{0.7}\% of their attention to matching panels on the chosen one for the first answer, and 65.9\pmv{0.6}\% for the second.

\input{figures/appendix/select_readout}

\paragraph{Holding the comma token's residual after the cursor is moved.} We take the held-out pairs with one letter written and let the 32 answer-token heads read the source's image keys at the answer token. This moves the cursor: the target answers $r_3$ in 73.8\pmv{3.8}\% of cases (the 16 binding heads and the 2 query and key heads outside them: 48.2\pmv{4.4}\%). We then hold the comma token's residual at one layer at the target's own values, on top of the patch (Figure~\ref{fig:restore}). Held at layer 19 or earlier, the patch keeps its effect in 69.8\pmv{3.9}\% of cases. Held at layer 23 or later, the target answers its own $r_2$ again in 99.4\pmv{0.7}\% (binding heads 99.0\pmv{0.9}\%). The hold alone changes nothing, with the target's own answer in 100.0\pmv{0.0}\% of cases at every layer tried. Holding the gaze heads' queries at the comma token instead of the residual brings the answer back in 99.8\pmv{0.3}\%. So the cursor set at the answer token acts through the comma token's residual up to layer 23, and through what the gaze heads read from it.

\subsection{Hiding a Panel From a Layer On}
\label{app:select-sweep}
\input{figures/appendix/select_sweep}
To find the layers where the choice happens, we hide the next matching panel from every text token from layer $l$ on (Figure~\ref{fig:appendix_select_sweep}). Hidden from the first layer, the model moves on to the following matching panel in 84.8\% of cases, so the other matching panels are there to fall back on. Hidden from layer 20 on, it moves on in 72.2\%. Hidden from layer 21 on, it moves on in 37.4\%, and from layer 25 on, in 2.8\%. So the choice happens at the comma token in layers 20-25. Hiding a non-matching panel, from every text token except the shape token, moves the answer to the following matching panel in at most 0.2\pmv{0.3}\% of cases at any layer. Two more results support this reading. Hiding the first matching panel the same way makes the model start from $r_2$ in 43.6\pmv{4.3}\% of cases when it is hidden from the first layer, and in 18.4\pmv{3.4}\% when it is hidden from layer 22 on, with the fall in layers 16-22. The first panel is harder to hide than the second, because it is the leftmost and its content also reaches the text through the image tokens of later panels. A copied pointer holds one panel. We copy the colon state of a source whose matching panels are $r_2$ and $r_4$ into the target at layer 24, and hide $r_2$ from the colon token from that layer on. The target goes back to its own first matching panel in 76.6\pmv{3.7}\% of cases, and to the source's other matching panel in only 3.2\pmv{1.5}\%.

\subsection{The Read Refuses a Written Panel and a Wrong Shape}
\label{app:select-refuse}
We point the gaze heads, with the source's queries, at a panel the target should not read. When the panel holds the asked shape and is not yet written, the answer goes there in 88.2\pmv{2.8}\% of cases. When the panel holds another shape, the answer goes there in 22.4\pmv{3.8}\% of cases. It goes there in 85.1\pmv{3.1}\% when the gaze heads also read the source's image keys for that panel, so the shape check runs through the image keys the gaze heads see. When the panel is already written, the answer goes there in 0.6\pmv{1.0}\% of cases, and in 22.0\pmv{6.2}\% when the gaze heads also read the source's image keys for that panel. The refusal of a written panel goes through the answer token. When the comma token cannot attend to that answer token, the gaze heads are followed in 97.6\pmv{2.2}\% of these cases, and in 100.0\pmv{0.0}\% with the source's image keys as well. When we block the answer token itself from the image, the refusal stands (6.1\pmv{3.7}\%), because the comma token can bind the letter to the image by itself (Appendix~\ref{app:progress-relay}). So the cursor also stops the gaze heads from reading a written panel.

\subsection{The Searches for Selection Heads}
\label{app:select-heads}
We searched three times for heads that make the choice. In the first search, we start by replacing the output of one head at a time at the last comma token with the source's, for every head in layers 20-25. We rank the heads on the discovery pairs, scored for the next panel and for the cursor, and sweep the top sets on the validation pairs. No size reaches the 80\% rule for either outcome. The best size for the next panel, 64 heads, reaches 44\% there. On the held-out pairs, those heads move the next panel in 61.9\% of cases (random heads 0.0\%). Their queries move it in 40.7\%, and the cursor in 21.6\%. Their image keys move it in 0.0\%, and their text values in 5.7\%. Fourteen of the 64 are gaze heads. With the gaze heads' queries held at the target's values, the output patch moves nothing (0.0\%), so these heads act by writing what the gaze heads read. Replacing the outputs of the gaze heads themselves copies the source's letter in 100.0\% of cases, which is the read and not the choice.

The two other searches need no source. We rank the heads of layers 16-25 at the comma token in two ways. The first is the drop in the margin between the next letter and the best other matching panel's letter when the head's output is zeroed. The second is the shift of the head's image attention from the other matching panels onto the next panel. The top heads of the two rankings overlap (L21H11, L20H29, L20H30, L21H18, L23H26 in both). On the held-out pairs, zeroing the top 4, 8, 16, 32, or 64 heads of either ranking leaves the next letter unchanged in 87.1\% to 99.0\% of cases. Copying their outputs from the source moves the next panel in at most 33.5\% (the top 64 by margin drop), against at most 0.5\% for the same number of random heads. With the gaze heads' queries held, every such patch leaves the answer unchanged in 97.9\% or more. So select is spread over many heads of layers 20-25, whose joint result the gaze heads read.

%% file: figures/appendix/many_probe.tex
\begin{wrapfigure}{R}{0.4\textwidth}
\vspace{-10pt}
\centering
\includegraphics[width=\linewidth]{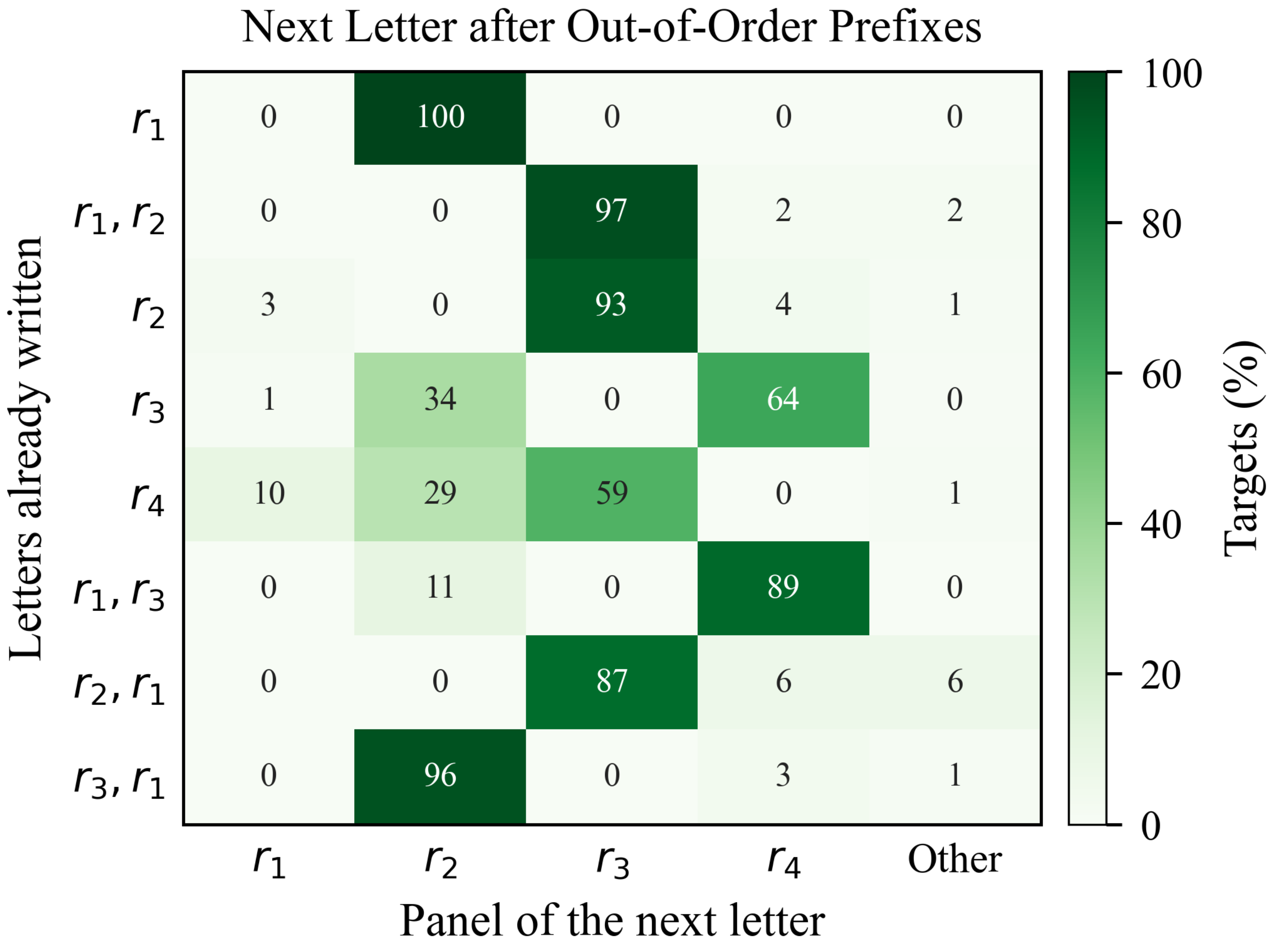}
\caption{\textbf{Select starts at the last written panel.} We write letters out of order, with no patch. Each row is a prefix of written letters. Each cell is the percent of held-out targets whose next letter comes from that matching panel.}
\label{fig:appendix_many_probe}
\vspace{-10pt}
\end{wrapfigure}

%% file: tables/appendix/mechanism_order.tex
\begin{table}[ht]
\centering
\small
\caption{\textbf{Letters written out of order, with attention blocked.} We block the comma token's attention to one of the written letters, and read which panel the next letter comes from. Each cell is a percent of targets. The rows use the lists whose next letter is right with nothing blocked: 87.2\% of the lists after ``$r_2$, $r_1$'' and 92.6\% after ``$r_3$, $r_2$, $r_1$''.}
\label{tab:mechanism-order}
\fittable{%
\begin{tabular}{@{}llrr@{}}
\toprule
Written so far & The comma token cannot attend to & Next unwritten panel & A panel already written \\
\midrule
$r_2, r_1$ & $r_2$ & 45.4\pmv{4.7}\% & 52.8\pmv{4.6}\% \\
$r_2, r_1$ & $r_1$ & 80.5\pmv{3.8}\% & 6.2\pmv{2.3}\% \\
$r_3, r_2, r_1$ & $r_3$ & 42.5\pmv{4.6}\% & 57.0\pmv{4.5}\% \\
$r_3, r_2, r_1$ & $r_2$ & 76.2\pmv{3.9}\% & 23.5\pmv{3.7}\% \\
$r_3, r_2, r_1$ & any letter but the last & 38.2\pmv{4.4}\% & 60.5\pmv{4.3}\% \\
\bottomrule
\end{tabular}}
\end{table}

%% file: figures/appendix/restore.tex
\begin{figure}[t]
\centering
\includegraphics[width=0.4\linewidth]{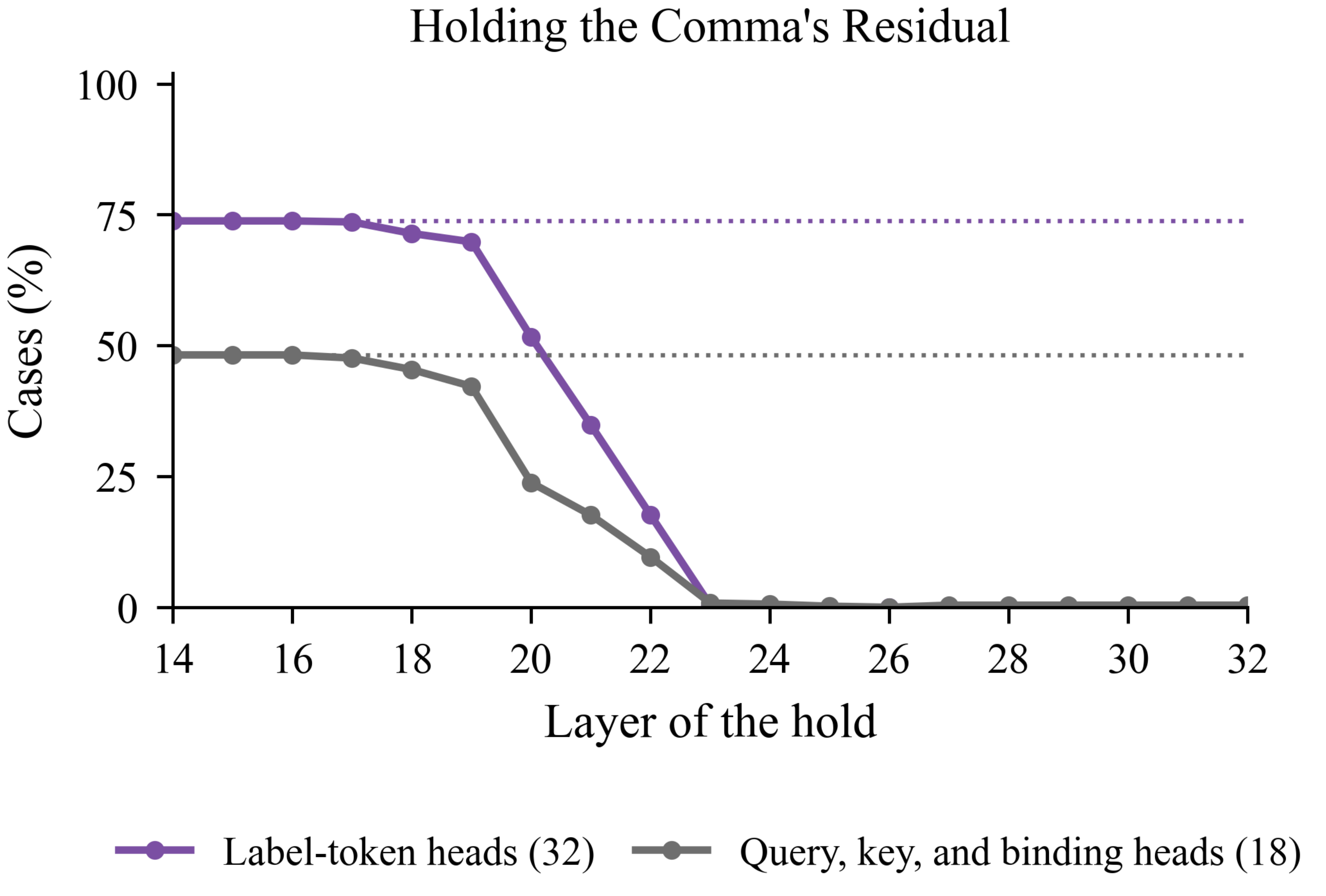}
\caption{\textbf{The cursor acts through the comma token's residual up to layer 23.} Pairs that still answer from the moved cursor when we hold the comma token's residual at one layer. Dotted lines show the patch alone.}
\label{fig:restore}
\end{figure}

%% file: figures/appendix/select_readout.tex
\begin{figure*}[t]
\centering
\includegraphics[width=\textwidth]{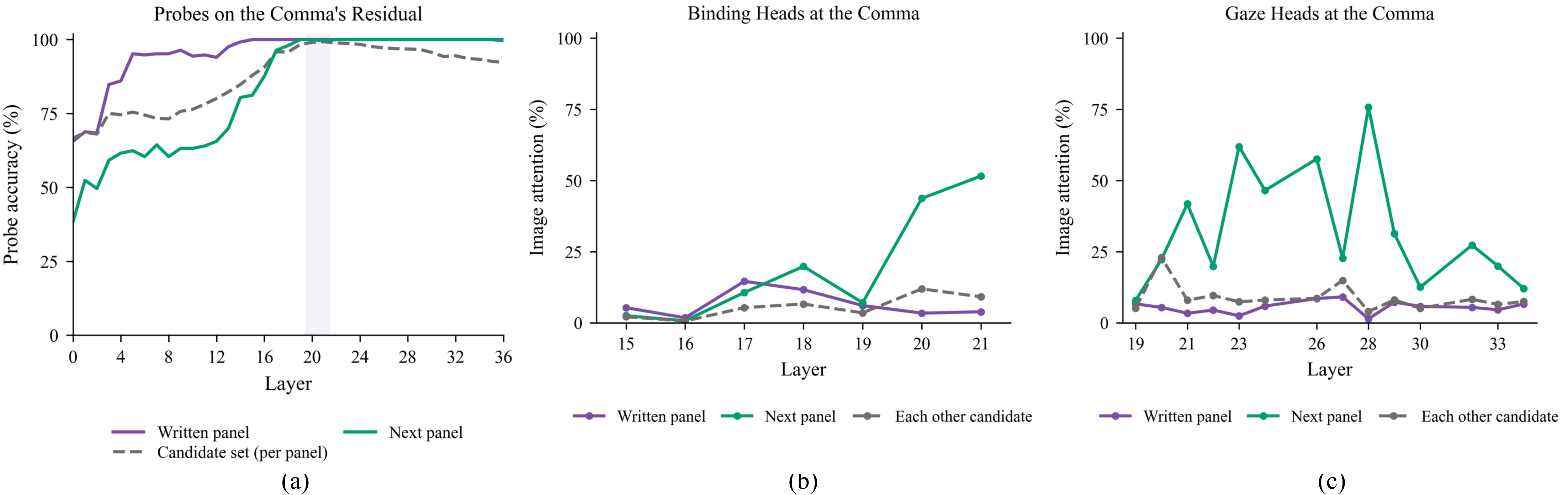}
\caption{\textbf{What the comma token's residual holds, with no patch.} (a)~We train ridge probes on the residual of the last comma token on half of the held-out pairs of the list task, and read them on the other half. The written panel is named in 94.0\pmv{2.8}\% of pairs at layer 12 and 100.0\pmv{0.0}\% at 18. The probes read whether each panel is a matching panel with 95.9\pmv{1.0}\% accuracy at layer 17 and 99.3\pmv{0.5}\% at 21, from a chance level of two thirds. The next panel is named in 98.0\pmv{1.8}\% at layer 18 and 100.0\pmv{0.0}\% at 20. (b)~Image attention of the binding heads at the comma token, by layer. At layer 21 they put 51.6\pmv{1.9}\% of it on the next panel and 3.9\pmv{0.2}\% on the written panel. (c)~Image attention of the gaze heads at the comma token, by layer. At layer 20 they give 22.3\pmv{0.6}\% to the next panel and 23.0\pmv{0.7}\% to each other matching panel, and by layer 23 they give 61.9\pmv{1.4}\% and 7.4\pmv{0.4}\%.}
\label{fig:select-readout}
\end{figure*}

%% file: figures/appendix/select_sweep.tex
\begin{wrapfigure}{R}{0.4\textwidth}
\vspace{-10pt}
\centering
\includegraphics[width=\linewidth]{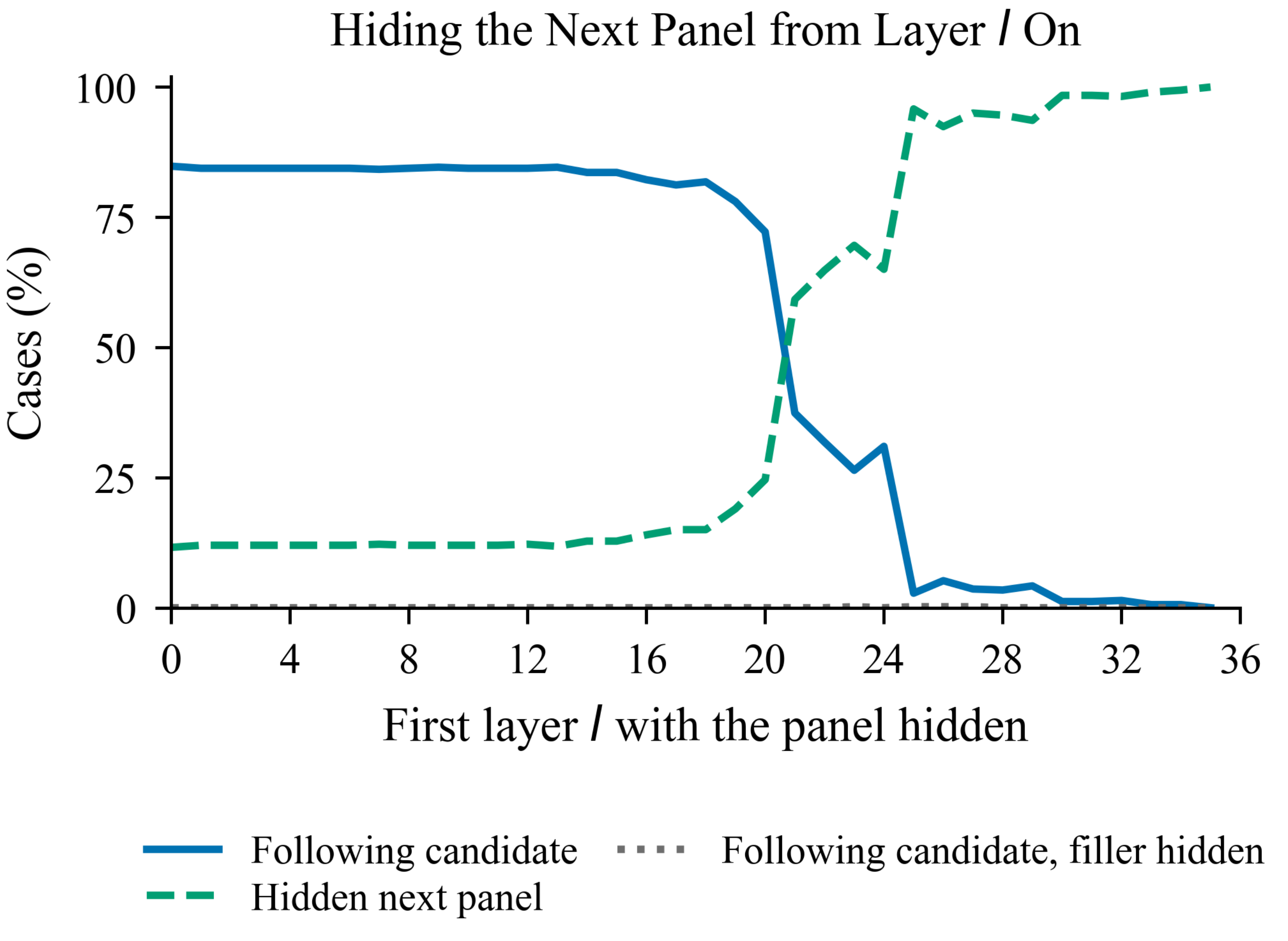}
\caption{\textbf{The next panel is chosen in layers 20-25.} Lists that move on to the following matching panel when we hide the next panel from layer $l$ on. Hiding a non-matching panel moves at most 0.2\pmv{0.3}\% of lists.}
\label{fig:appendix_select_sweep}
\vspace{-10pt}
\end{wrapfigure}

%% file: secs/appendix/F_reading_language.tex
\section{Reading and Language: Details and Supporting Experiments}
\label{app:reading}

This appendix supports Section~\ref{sec:reading}. It covers the two searches for the heads that execute the pointer, the comparison with the published Gaze Heads, the checks on the gaze heads' keys and values, the read on a list, what the gaze heads bring back, the object task, and every language experiment.

\subsection{The Two Searches, and Which Heads They Select}
\label{app:read-screens}
A head that reads the pointer must build its query from it, so the main text ranks every head by a query patch at the colon token. To check whether other heads also read the image, we run a second search. We block one head at a time from every image token, and score the drop in the margin of the target's own letter over the best of the other letters in its image. Figure~\ref{fig:appendix_gaze_rankings}b reports both searches for every $K$ on the validation pairs. The query-patch search rises steeply up to eight heads and stays flat from 16 to 256. So the part of the pointer's effect that it does not move is not a matter of missing heads. The blocking search finds no set of heads whose blocking breaks the answer. Even 256 blocked heads leave the answer intact in almost every case, and the heads this search ranks highest have no effect in any held-out test (Table~\ref{tab:readers}).

The read at the colon token has many possible routes, and removing a large share of them is not enough to stop it. When we block the 32 gaze heads from the image, the target still gives the right answer in 88.2\pmv{2.0}\% of the held-out cases. When we block the top 200 published Gaze Heads from the image, it does so in 98.9\pmv{0.8}\%. With the gaze heads blocked, the letter is read earlier, by the tokens between the shape token and the colon token (Appendix~\ref{app:one-knockouts}). So the gaze heads are sufficient, and the pointer acts only through their queries, but they are not the only heads that can read.

\input{figures/appendix/gaze_rankings}
\input{tables/appendix/readers}

\subsection{Our Gaze Heads and the Published Gaze Heads}
\label{app:read-overlap}
Table~\ref{tab:readers} already places the published top 100 and top 200 next to our gaze heads. Here we compare the rankings themselves, the effect against the number of heads, and the same ranking measured on our own task.

Figure~\ref{fig:appendix_gaze_rankings}a compares three rankings of all 1,152 heads. The first is our query-patch search. The second is the Gaze score of \citet{gandikota2026gaze} measured on our task: the attention a head places, from the colon token, on the image tokens of the asked panel, averaged over every question of Appendix~\ref{app:task-native}. The third is the original Gaze ranking from comic strips. Our gaze heads agree best with the Gaze score measured on the same task. The original and re-measured rankings share 57 of their top 100 heads, so the ranking itself shifts somewhat with the task. Of the 32 gaze heads, 23 are in the original top 100.

\paragraph{Effect against the number of heads.} To compare the rankings by their causal effect, we patch the top $K$ heads of each ranking together at the colon token, for $K$ in 1, 2, 4, 8, 16, 32, 64, 100, 128, 200, and 256, on the held-out pairs (Figure~\ref{fig:gaze_language}b). We use three rankings of all 1,152 heads: our query-patch search, the published Gaze Heads ranking of \citet{gandikota2026gaze} as published, and a random order of all heads, drawn anew for every pair and direction. Each $K$ gets two patches in separate runs. In the query patch, we replace the heads' queries with the source's, and a case counts when the target answers with its own letter at the source's panel. In the value patch, the heads keep their own queries and keys but read the source's image values, and a case counts when the target answers with the source's letter at the target's asked panel. Our 32 gaze heads reach 74.0\pmv{2.9}\% with the query patch and 93.7\pmv{1.6}\% with the value patch. The top 100 published heads reach 64.2\pmv{2.9}\% and 50.0\pmv{3.3}\%. The top 256 published heads reach 63.8\pmv{2.9}\% and 89.1\pmv{2.0}\%. Random heads stay at or below 2.1\pmv{1.0}\% at every $K$.

\paragraph{The same ranking on the same task.} The comparison with the published Gaze Heads in Section~\ref{sec:reading} changes two things at once: how the heads were chosen, by patching or by attention, and where, on our task or on comic strips. Table~\ref{tab:fair-baseline} removes the second difference. We rank all 1,152 heads by the Gaze score measured on our own task, and patch its top 32 and top 100 on held-out cases, next to our 32 gaze heads and the published top 100. On the query patch, the heads ranked by where they look on this task read the pointer less well than the 32 heads chosen by patching: 63.3\% and 65.3\%, against 74.0\pmv{2.9}\%. On the value patch, the re-measured top 100 reaches 93.4\%, close to our 93.7\pmv{1.6}\%, but with three times as many heads. Its top 32 gives 64.8\%.

\input{tables/appendix/fair_baseline}

\paragraph{Every head set against the published Gaze Heads.} The published list was made by observing where heads look while the model describes comic strips, with no patch and no task of ours. Table~\ref{tab:gaze-overlap-sets} counts how many heads of each set in this paper are in it. The published list is close to a superset of our gaze heads: 23 of the 32 are in its top 100, and 29 in its top 200. It holds almost none of our other heads. Of the 16 binding heads and the 2 query and key heads outside them, 3 are in the top 100, which is what chance gives. Of the 16 relay heads, 4 are, and of the 16 language heads, 0. The searches on the list task show the same pattern. The heads found at the last comma token, where the next panel is read, are published Gaze Heads, with 11 of 16 in the top 100. The heads found at the shape token and at the answer token mostly are not. When we measure the Gaze score again on our task, it ranks more of the binding, query, and key heads highly, 6 of 18 in its top 100, because on these images a binding head also looks at the shape that will be read. So the published Gaze Heads are the heads of one step, the read. Other heads do the binding, the relay, and the choice of language, and attention to the described region does not reveal them.

\input{tables/appendix/gaze_overlap_sets}

\subsection{Gaze Heads: Keys, Values, and the Read on a List}
\label{app:gaze-checks}
\paragraph{Keys match, values carry.} To see which part of the gaze heads does what, we patch their image keys and image values separately. When the gaze heads read the source's image keys with their own queries and values, the target keeps its answer in 93.2\pmv{1.6}\% of the held-out cases. When they read the source's image values, the target answers with the source's letter at its own panel in 93.7\pmv{1.6}\%. When they read both, it does so in 87.9\pmv{1.9}\%. So the queries carry the pointer, the keys on the image tokens match it by place, and the values bring the letter.

\paragraph{The same gaze heads read every answer of a list.} In the list task, the pointer to the next panel forms at the last comma token, and the same 32 gaze heads read it. When we patch their queries from the source, the second answer moves to the source's next panel in 88.2\pmv{2.8}\% of cases, against 0.0\pmv{0.0}\% for random heads. Conversely, when we hold their queries at the target's values, a moved cursor changes the answer in only 2.1\%. At the colon token of the list task, patching the queries of the 32 gaze heads moves the first answer in 46.1\% of cases. A new search there selects 128 heads, which move it in 57.8\pmv{4.2}\% (Table~\ref{tab:many-heads}). The colon token of the list prompt has more than one panel to choose from, so fewer of its answers follow the patched queries.

When we point the gaze heads at a panel already written, or at a panel of another shape, the model does not follow them. Appendix~\ref{app:select-refuse} gives this test together with select.

\subsection{What the Gaze Heads Bring Back}
\label{app:read-content}
Appendix~\ref{app:read-implementation} defines the readout of the gaze heads' summed output and the cross-language patch.

\paragraph{Are the gaze heads the same heads on this task?} We found the gaze heads on the letter task, with another prompt and another answer token. To check whether the same heads read the colored-shape task, we repeat a single-head search on its discovery images. We patch the output of one head at a time from the source, and score the gain of the source's concept word over the target's, both in the target's language. This search ranks 14 of the 32 gaze heads in its top 32, and 25 in its top 128.

\paragraph{Results.} The summed output of the gaze heads is nearly the same for one image under two languages, with a cosine of 0.95, and differs between two images under one language, 0.50. Read against the output vocabulary, its top word is the English name of the shape in 100.0\pmv{0.0}\% of images when the question is German, and in 84.2\pmv{3.2}\% when it is French. Otherwise it is the Chinese name. It is never the German or French name.

When we patch the outputs of the 32 gaze heads from the source, the target names the source's shape in its own language in 82.0\pmv{1.3}\% of cases. It changes the language in 0.0\pmv{0.0}\%, and keeps its own answer in 13.8\pmv{1.4}\%. Patching the 16 language heads moves the language in 95.3\pmv{0.5}\% and the shape in 0.0\pmv{0.0}\%. Patching 32 random heads moves the shape in 0.0\pmv{0.0}\% and the language in 0.3\pmv{0.2}\%. So the gaze heads move the concept and never the language, and the language heads of Section~\ref{subsec:language} do the opposite. The top 32 heads of the search on this task move the concept in every case. So the gaze heads found on the letter task are most, but not all, of what carries the concept here. With a source in the target's own language, the gaze heads move the concept in 86.8\pmv{1.2}\% of cases. Patching them at the first generated token only moves it in 67.7\%.

\subsection{The Object Task}
\label{app:read-objects}
With four shapes, a readout that names the shape could be a coincidence of a small vocabulary. To rule this out, we use the object task, which has sixteen concepts. With no patch, the model answers its questions in the right language with the right object in 99.0\% of runs. The summed output of the same 32 gaze heads has a cosine of 0.96 for one image under two languages, and 0.36 for two images under one language. Across readouts, its top word is the English name of the object in the circle in 48.8\% of them, and the start of that English name in 33.0\%. It is the Chinese name in 2.0\%, and the German or French name in 0.0\%. Patching the 16 language heads moves the language in 87.6\% of cases and the object in 0.0\%. So the separation of concept and language holds on both tasks.

What changes is which heads carry the concept. When we patch the outputs of the 32 gaze heads from a source that asks in another language about another object, the target names the source's object in only 9.8\% of cases. In 15.5\%, the target names the object that its own image holds at the source's asked panel. So part of what these heads pass on is still the pointer, which later heads use to read the image again. The heads that the single-head search of Appendix~\ref{app:read-content} finds on this task do carry the object. Patching its top 32 heads moves the object in 86.1\% of cases and the language in 0.0\%, against 0.0\% for random heads. This search ranks 14 of the 32 gaze heads in its top 32, and 28 in its top 128. Of its top 32 heads, 21 are also in the top 32 of the search on the colored-shape task. So the heads that read the pointer also read a letter or a plain shape, but a drawn object needs further heads after them.

\subsection{Language Heads: All Conditions, Both Tasks}
\label{app:read-language}
\paragraph{Finding the language heads.} To find the component that adds the language, we take a source that asks in language $A$ about image $X$ and a target that asks in language $B$ about image $Y$. We track the token of the target's answer in the source's language $A$, a word that neither run produces. We start by patching the output of one attention head at a time from the source. We rank the heads by this score on the discovery images. We keep the top 16 heads of this search, in layers 19-35 (Figure~\ref{fig:gaze_language}c).

Table~\ref{tab:language-runs} reports every language experiment on both tasks. The two tasks select largely the same heads: 10 of the 16 are shared, and the two strongest heads are the same in both. The object task is the harder one for almost every patch, because its answers are often several tokens long. Patching the heads at the first generated token only gives 78.4\% on this task, against 87.6\% at every token. To find what writes the word, we hold parts of the model at the target's values while patching the heads. Holding the last three MLPs, layers 33-35, keeps the language in place on both tasks, with 0.9\pmv{0.2}\% and 5.0\% moved. Holding MLPs 28-32 leaves 10.7\% and 4.4\% moved. Holding the other attention heads of layers 33-35 barely matters, with 91.3\% and 79.7\% moved. So the late MLPs, and not other late heads, write the word.

\paragraph{A single vector sets the language.} We take the mean difference between the summed outputs of the 16 heads under a language and under English, and add it to the residual at layer 24. This single vector does the same job as the per-head averages. On held-out images of the colored-shape task, it switches English runs to French in 100.0\pmv{0.0}\% of cases, to Chinese in 100.0\pmv{0.0}\%, and to German in 75.0\pmv{3.8}\%. Subtracted from German and French runs, it returns English in 100.0\pmv{0.0}\%. It also works on text-only prompts with no image, whose nouns never appear in any image. There it switches runs to French in 65.0\%, to Chinese in 100.0\%, and to German in 60.0\%.

\paragraph{The late MLPs accept concepts they have not been shown.} To test whether the translation step depends on the visual read, we add the vocabulary direction of a new English noun to the residual at layer 26 of a German, French, or Chinese run. The model then produces that noun in the question's language in 74.6\pmv{0.7}\% of cases on the colored-shape task, and in 67.0\% on the object task. It gives the English word in 4.6\pmv{0.2}\%. When we add the same direction late, after the language heads have acted, the model gives the English word instead, in 69.2\pmv{0.2}\%, and the translated word in 3.8\pmv{0.1}\%. So the translation step at the end of an answer takes any concept that reaches it.

\input{tables/appendix/language_runs}

\subsection{The Color Question: Language Heads and MLPs}
\label{app:color-task}
Here we study further how the late layers turn the visual concept from the gaze heads into a word, on the color question of Appendix~\ref{app:read-implementation}. This extends Section~\ref{subsec:language}.

\input{figures/appendix/colors_mlps_ablation}
\paragraph{Language heads.} We patch one component at a time from a source into a target that ask the question in two different languages about two different images. The two images show the same four shapes in different panels, colored with eight different colors. The source and the target ask about different shapes, so the target's concept, the source's concept, and the target's concept at the source's panel are three different words. The pairs cover all 12 ordered pairs of English, French, Chinese, and German, balanced over markers, concepts, and panels. We apply every patch at the colon token of the prefill, and compute scores from the logits of the patched target.

Figure~\ref{fig:color-heads} shows the head search on this question.

\paragraph{MLPs.} We repeat the MLP search on the color question. We start by patching the output of one MLP at a time at the colon token from the source (Figure~\ref{fig:color-mlps}a). We then patch the top $k$ MLPs together (Figure~\ref{fig:color-mlps}b). Ranked over all languages and both questions, the top 4 MLPs are layers 32-35, and the top 8 are layers 28-35. These late MLPs are also needed without any source. When we set their outputs to their mean over prompts in all languages, German and French answers fall back to English (Figure~\ref{fig:color-mlp-ablation}).

\input{figures/appendix/colors_heads}
\input{figures/appendix/colors_mlps}

\subsection{Zeroing the Heads, and Setting Them}
\label{app:read-zeroing}
Figure~\ref{fig:appendix_zeroing} compares zeroing the language heads with setting them to their English averages. Setting 16 heads returns English in every run of both tasks. Zeroing them works differently for each language of the question. On the colored-shape task, zeroing 16 heads returns English in 99.6\pmv{0.5}\% of German runs, 88.4\pmv{2.9}\% of French runs, and 0.0\pmv{0.0}\% of Chinese runs. Zeroing 32 or 64 heads brings French to 100\% and leaves Chinese at 0\%. Zeroing more heads breaks the output. Zeroing random heads never returns English. So for German and French the model needs the heads to keep the language, and without them it falls back to English. Chinese runs stay Chinese with the heads zeroed. Yet they switch to English, German, or French when we set the same heads to those averages (Figure~\ref{fig:appendix_switch}). So the heads can overrule the language, and for Chinese some other part of the model also carries it. Leaving English is the hardest direction, and it needs 32 heads instead of 16.

\input{figures/appendix/zeroing}
\input{figures/appendix/switch}

\paragraph{What keeps Chinese runs Chinese.} To find this other part, we zero the 16 heads in a Chinese run on the discovery images. We then patch one more component from a run on the same image asked in English, so that the two runs differ only in language. We score how far the first answer token moves toward the English word. No single head moves it far. Patching the whole residual of the last prompt token closes most of the gap from layer 4 on. So the token already knows the language of the conversation a few layers in, before any of the 16 heads act. MLPs 22-35 each move the answer when patched. On held-out images (Table~\ref{tab:chinese}), zeroing the 16 heads together with the 64 heads that rank highest in this search returns English in 90\% of Chinese runs. Zeroing the same 64 heads alone returns English in 0\%, and so does adding 64 random heads to the 16. Setting the early residual to its English average has the same effect: with the 16 heads zeroed it returns English, and alone it does not. So Chinese is carried twice, once by the 16 heads and once by an early trace in the residual stream that a wider set of heads and the late MLPs read. French has a weaker second copy, and German has none that we can detect.

\input{tables/appendix/chinese}

%% file: figures/appendix/gaze_rankings.tex
\begin{figure*}[t]
\centering
\includegraphics[width=\textwidth]{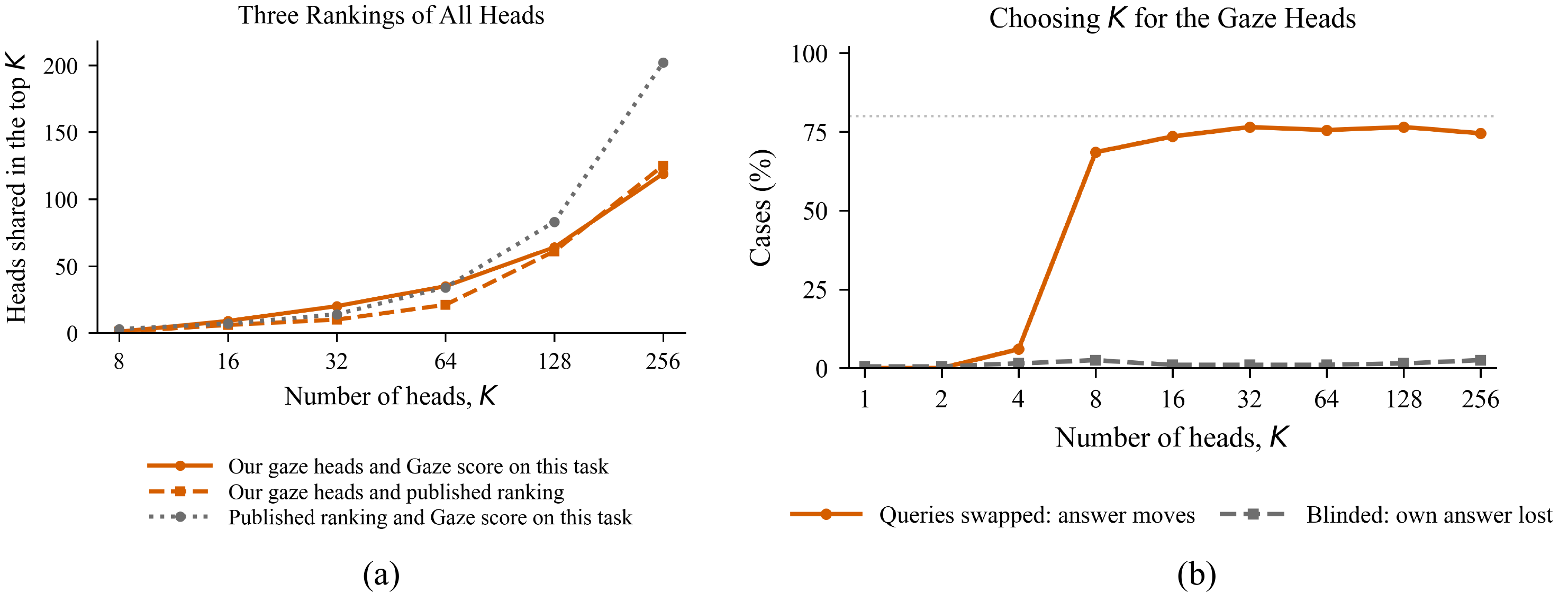}
\caption{\textbf{Finding the gaze heads, and how they compare with the published ranking.} (a)~The number of heads shared in the top $K$ of three rankings of all 1,152 heads: our query-patch search, the Gaze score measured on our task, and the published Gaze Heads ranking. From $K=16$ to 128, our ranking shares more heads with the Gaze score on this task than with the published ranking, 20 against 10 in the top 32. (b)~The top $K$ heads of the two searches, patched together at the colon token on the 100 validation pairs. Patching their queries moves the answer to the source's panel in 68.5\% of cases at $K=8$, and in 73.5\% to 76.5\% from 16 to 256 heads. This stays below the 80\% rule, so the rule takes its best size, 32 heads. Blocking up to 256 heads from the image loses the target's answer in at most 2.5\%.}
\label{fig:appendix_gaze_rankings}
\end{figure*}

%% file: tables/appendix/readers.tex
\begin{table}[ht]
\centering
\small
\caption{\textbf{The heads that read the pointer, found by patching, against the published Gaze Heads and the blocking search.} Every patch acts at the colon token only. Percent of the 500 held-out pairs. The published ranking plays no part in our search.}
\label{tab:readers}
\fittable{%
\begin{tabular}{@{}lccccc@{}}
\toprule
Head set & Queries patched: source's panel & Pointer moved, queries held: own kept & Source's image values: source's letter & Blocked from the image: own lost & Only these heads see: own kept \\
\midrule
32 gaze heads found by patching & 74.0\pmv{2.9}\% & 98.4\pmv{0.8}\% & 93.7\pmv{1.6}\% & 11.8\pmv{2.0}\% & 95.1\pmv{1.3}\% \\
same number of random heads & 0.0\pmv{0.0}\% & 5.9\pmv{1.6}\% & 0.0\pmv{0.0}\% & 0.2\pmv{0.2}\% & 4.8\pmv{1.2}\% \\
Published Gaze ranking, top 100 & 64.2\pmv{2.9}\% & 95.2\pmv{1.3}\% & 50.0\pmv{3.3}\% & 1.4\pmv{0.8}\% & 97.4\pmv{1.0}\% \\
Published Gaze ranking, top 200 & 64.9\pmv{3.1}\% & 96.2\pmv{1.1}\% & 88.6\pmv{2.1}\% & 1.1\pmv{0.8}\% & 98.6\pmv{0.8}\% \\
Heads from the blocking search & 0.0\pmv{0.0}\% & 8.3\pmv{2.0}\% & 0.0\pmv{0.0}\% & 2.1\pmv{0.9}\% & -- \\
\bottomrule
\end{tabular}}
\end{table}

%% file: tables/appendix/fair_baseline.tex
\begin{table}[ht]
\centering
\small
\caption{\textbf{Attention-ranked heads on the same task.} We rank all 1,152 heads by the Gaze score of \citet{gandikota2026gaze} measured on our task, the attention a head places from the colon token on the asked panel. Percent of held-out cases. Query patch: the target answers from the source's panel. Image values: the target gives the source's letter at its own panel.}
\label{tab:fair-baseline}
\fittable{%
\begin{tabular}{@{}lcc@{}}
\toprule
Head set & Queries patched: source's panel & Source's image values: source's letter \\
\midrule
32 gaze heads (chosen by patching) & 74.0\pmv{2.9}\% & 93.7\pmv{1.6}\% \\
Gaze score on our task, top 32 & 63.3\% & 64.8\% \\
Gaze score on our task, top 100 & 65.3\% & 93.4\% \\
Published Gaze Heads, top 100 & 64.2\pmv{2.9}\% & 50.0\pmv{3.3}\% \\
32 random heads & 0.0\% & 0.0\% \\
\bottomrule
\end{tabular}}
\end{table}

%% file: tables/appendix/gaze_overlap_sets.tex
\begin{table}[ht]
\centering
\small
\caption{\textbf{The published Gaze Heads are the heads of the read step.} The number of heads of each set that are in the published ranking of Gaze Heads, and in the Gaze score measured again on our task. The last column gives the number expected if the set were drawn at random from the 1,152 heads.}
\label{tab:gaze-overlap-sets}
\fittable{%
\begin{tabular}{@{}lcccccc@{}}
\toprule
Head set & Heads & Layers & In published top 100 & In published top 200 & In re-measured top 100 & Expected by chance in a top 100 \\
\midrule
Binding, query, and key heads & 18 & 15-21 & 3 & 6 & 6 & 1.6 \\
Relay heads & 16 & 19-35 & 4 & 5 & 3 & 1.4 \\
Gaze heads & 32 & 19-34 & 23 & 29 & 29 & 2.8 \\
Language heads & 16 & 19-35 & 0 & 0 & 0 & 1.4 \\
List task: outputs at the shape token & 32 & 14-21 & 5 & 11 & 13 & 2.8 \\
List task: image keys at the answer token & 32 & 15-34 & 8 & 20 & 19 & 2.8 \\
List task: queries at the last comma token & 16 & 23-35 & 11 & 15 & 14 & 1.4 \\
List task: outputs at the last letter (stop) & 64 & 17-35 & 11 & 18 & 12 & 5.6 \\
\bottomrule
\end{tabular}}
\end{table}

%% file: tables/appendix/language_runs.tex
\begin{table}[ht]
\centering
\small
\caption{\textbf{Every language experiment on both tasks.} Held-out images, with cross-image cases. A case succeeds when the target names its own concept in the source's or the target's language.}
\label{tab:language-runs}
\fittable{%
\begin{tabular}{@{}p{8.6cm}cc@{}}
\toprule
Experiment & Colored shapes & Objects in shapes \\
\midrule
Answers in the question's language, no patch & 100.0\pmv{0.0}\% & 99.0\% \\
Language heads patched from the source, every generated token & 95.3\pmv{0.5}\% & 87.6\% \\
\quad first generated token only & 90.7\% & 78.4\% \\
\quad copied source words & 0.0\pmv{0.0}\% & 1.3\% \\
Random heads & 0.0\pmv{0.0}\% & 0.5\% \\
Lowest-ranked heads & 0.0\% & 0.3\% \\
Source in the same language: word unchanged & 100.0\% & 91.1\% \\
Heads patched, MLPs 33-35 held & 0.9\pmv{0.2}\% & 5.0\% \\
Heads patched, MLPs 28-32 held & 10.7\% & 4.4\% \\
Heads patched, other attention heads of layers 33-35 held & 91.3\% & 79.7\% \\
Heads set to their English mean (no source): English & 100.0\% & 100.0\% \\
Heads set to another language's mean: that language & 77.3\% & 85.2\% \\
\quad random heads given the same means & 0.0\% & 0.0\% \\
Heads zeroed: English & 62.7\pmv{2.4}\% & 23.7\% \\
Language vector, held-out images: French / Chinese / German & 100.0\pmv{0.0} / 100.0\pmv{0.0} / 75.0\pmv{3.8}\% & 15.0 / 100.0 / 36.0\% \\
Language vector, text-only prompts: French / Chinese / German & 65.0 / 100.0 / 60.0\% & 0.0 / 100.0 / 7.1\% \\
Vector subtracted from German and French runs: English & 100.0\pmv{0.0}\% & 100.0\% \\
New noun added at layer 26: named in the question's language & 74.6\pmv{0.7}\% & 67.0\% \\
\bottomrule
\end{tabular}}
\end{table}

%% file: figures/appendix/colors_mlps_ablation.tex
\begin{wrapfigure}{R}{0.5\textwidth}
\vspace{-10pt}
\centering
\includegraphics[width=\linewidth]{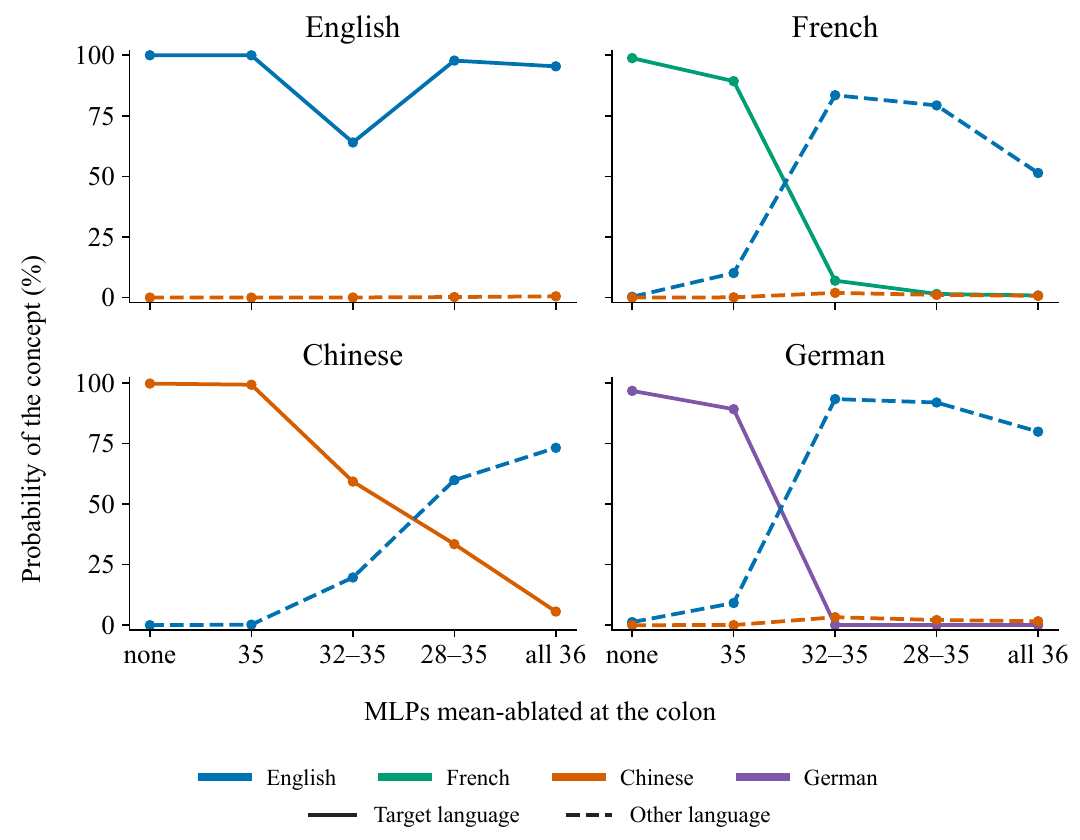}
\caption{\textbf{Without the late MLPs, German and French answers fall back to English.} We set the outputs of the top MLPs at the colon token to their mean over prompts in all languages, with no source. We then read the probability of the correct color word in the target language, which is the language of the question (solid), and in English and Chinese (dashed).}
\label{fig:color-mlp-ablation}
\vspace{-10pt}
\end{wrapfigure}

%% file: figures/appendix/colors_heads.tex
\begin{figure*}[t]
\centering
\begin{minipage}[t]{0.32\textwidth}\centering\includegraphics[width=\linewidth]{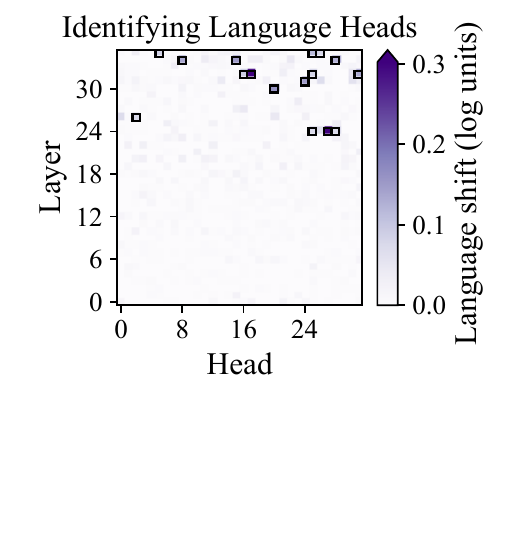}\\(a)\end{minipage}\hfill
\begin{minipage}[t]{0.32\textwidth}\centering\includegraphics[width=\linewidth]{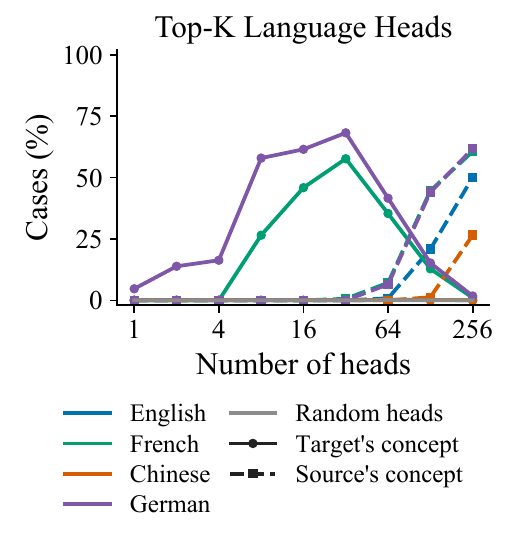}\\(b)\end{minipage}\hfill
\begin{minipage}[t]{0.32\textwidth}\centering\includegraphics[width=\linewidth]{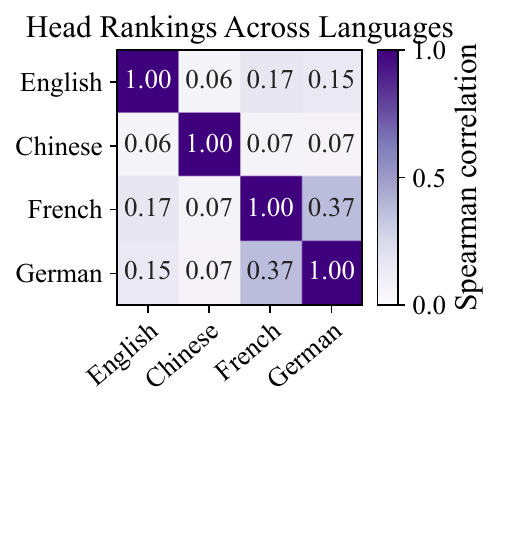}\\(c)\end{minipage}
\caption{\textbf{Language heads on the color question.} (a)~The language shift of each head, averaged over the 12 ordered pairs of English, French, Chinese, and German. We patch the output of one head at the colon token from the source, and measure how far the target's color words move from its own language toward the source's (log units). The top 16 heads are outlined. (b)~The top~$k$ heads patched together, by target language. We add heads in the order of a ranking over all languages. (c)~The Spearman correlation between the head rankings of each target language, over all 1{,}152 heads.}
\label{fig:color-heads}
\end{figure*}

%% file: figures/appendix/colors_mlps.tex
\begin{figure*}[t]
\centering
\begin{minipage}[t]{0.4\textwidth}\centering\includegraphics[width=\linewidth]{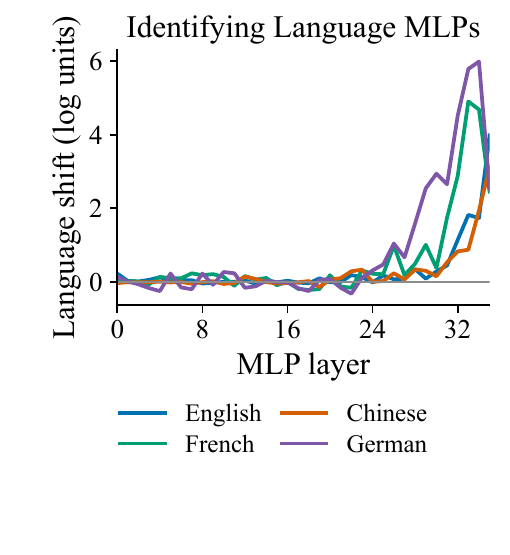}\\(a)\end{minipage}\hspace{0.05\textwidth}
\begin{minipage}[t]{0.4\textwidth}\centering\includegraphics[width=\linewidth]{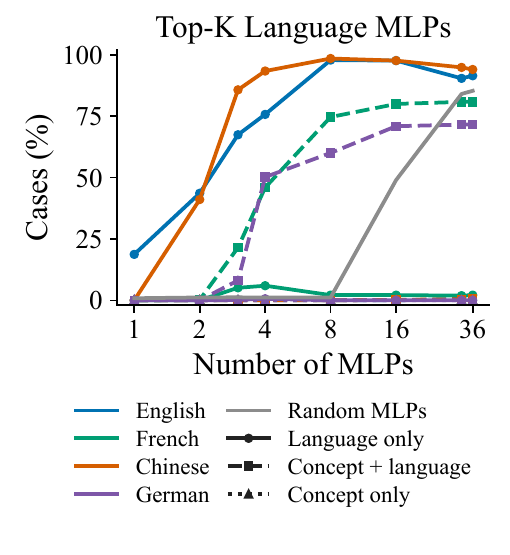}\\(b)\end{minipage}
\caption{\textbf{MLPs performing translation.} (a)~The language shift when we patch the output of one MLP at the colon token from the source, by target language, averaged over the three source languages. (b)~The top~$k$ MLPs patched together, ranked over all languages and both questions, grouped by source language and averaged over its three targets. The top 4 MLPs are layers 32-35, and the top 8 are layers 28-35.}
\label{fig:color-mlps}
\end{figure*}

%% file: figures/appendix/zeroing.tex
\begin{figure*}[t]
\centering
\includegraphics[width=\textwidth]{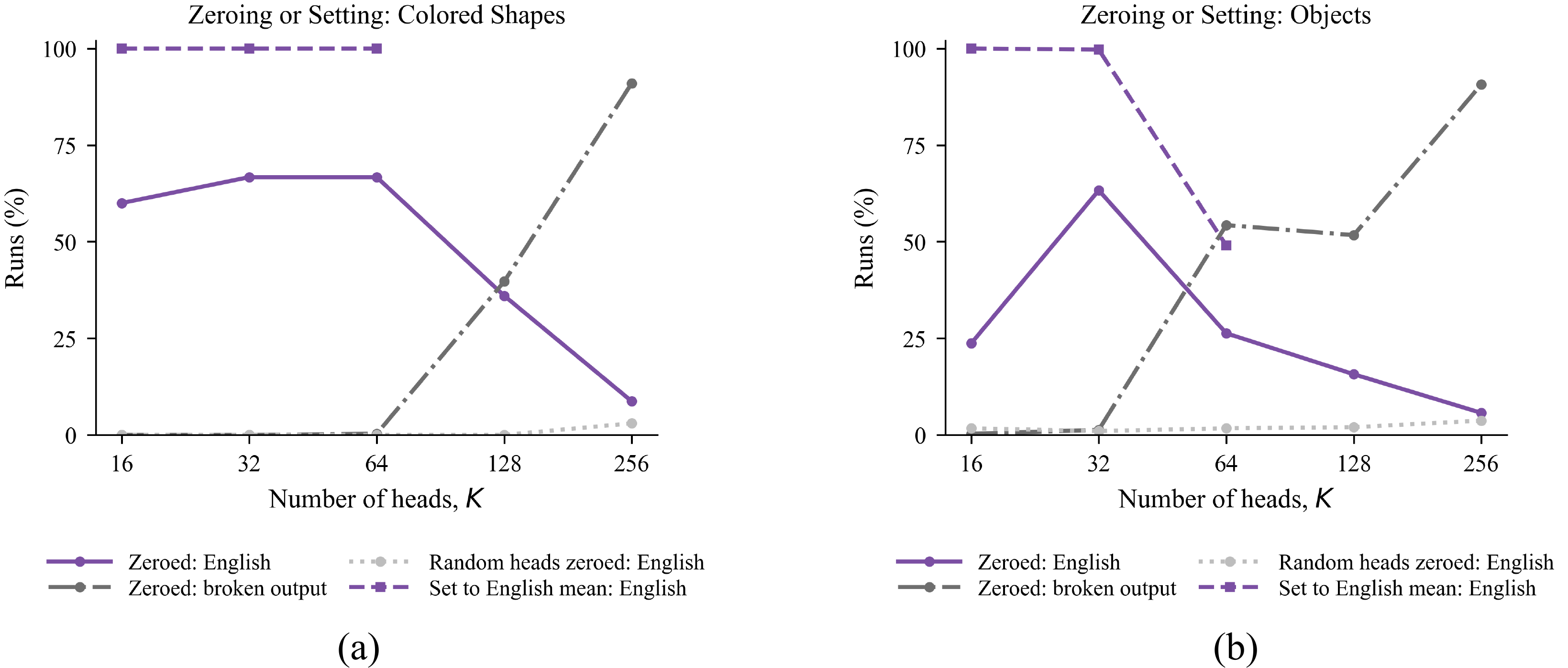}
\caption{\textbf{Zeroing the language heads, against setting them.} We patch the top $K$ heads of the language search at every generated token, in runs asked in German, French, or Chinese on the held-out images. Solid purple: we zero the heads and the answer turns English. Dash-dot gray: zeroing breaks the output. Dotted light gray: we zero random heads. Dashed purple: we set the heads to their English average and the answer turns English, run up to 64 heads. (a)~Colored shapes: 16 heads set to English give English in 100.0\% of runs, and 16 heads zeroed give it in 62.7\pmv{2.4}\%. (b)~Objects: 100.0\% and 23.7\%.}
\label{fig:appendix_zeroing}
\end{figure*}

%% file: figures/appendix/switch.tex
\begin{figure*}[t]
\centering
\includegraphics[width=\textwidth]{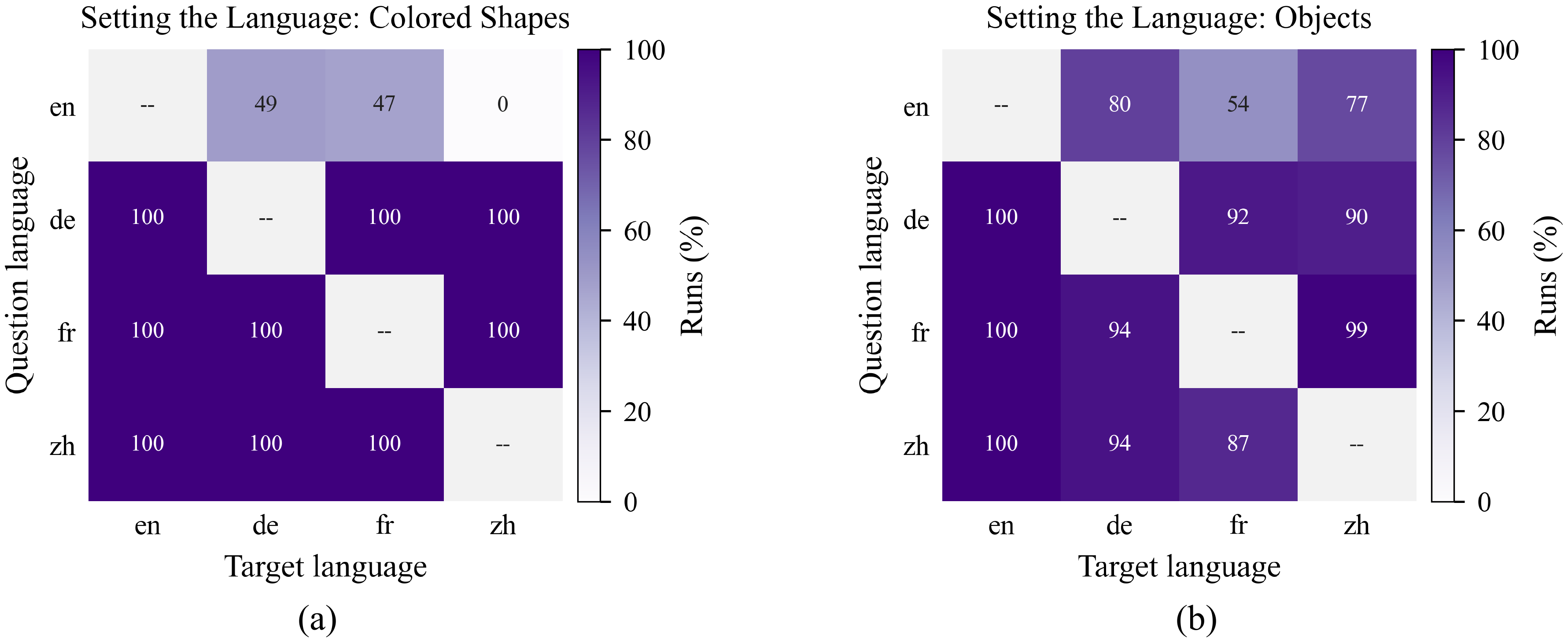}
\caption{\textbf{Choosing the language with no source.} We set the top 16 language heads to their average output under the target language (columns), in runs asked in the question language (rows). Each cell is the percent of held-out images whose answer names its own concept in the target language. (a)~Colored shapes. (b)~Objects. Leaving English is the hardest direction.}
\label{fig:appendix_switch}
\end{figure*}

%% file: tables/appendix/chinese.tex
\begin{table}[ht]
\centering
\small
\caption{\textbf{What keeps Chinese runs Chinese.} Percent of runs that answer in English, on all held-out images of the colored-shape task. We patch the heads at every generated token, and the residual at the last prompt token. We search the extra heads separately for each language.}
\label{tab:chinese}
\fittable{%
\begin{tabular}{@{}lccc@{}}
\toprule
Patch & German & French & Chinese \\
\midrule
16 language heads zeroed & 99.6 & 88.4 & 0 \\
\quad and 16 more heads found with those zeroed & 100 & 100 & 0 \\
\quad and 64 more heads found with those zeroed & 100 & 100 & 90 \\
\quad and 64 random heads & 92 & 85 & 0 \\
The 64 more heads zeroed alone & 0 & 12 & 0 \\
16 heads zeroed, residual at layer 4 set to the English average & 100 & 100 & 31 \\
\quad at layer 8 & 100 & 100 & 89 \\
\quad at layer 16 & 100 & 100 & 100 \\
Residual at layer 8 set to the English average alone & 0 & 0 & 0 \\
\bottomrule
\end{tabular}}
\end{table}

%% file: secs/appendix/G_generalization.tex
\section{Generalization}
\label{app:generalization}

The main paper studies one model, on rows of panels, with one prompt wording. In this appendix, we ask whether the mechanism holds beyond this setup. We test it on other sizes of the same model, on other model families, and on other images and prompts.

\subsection{Model Sizes}
\label{app:model_size}

The main paper studies the 8B model. To test whether the mechanism depends on model size, we run the same pipeline on the 2B, 4B, and 32B models of the Qwen3-VL-Instruct family. We carry no head over between models. Instead, we search every head set again on each model with the rule of Appendix~\ref{app:one-implementation}, and test it on that model's own held-out pairs.

\subsubsection{Implementation Details}
The 2B language model has 28 layers of 16 query heads, with 2 query heads per key-value group. The 4B and 8B have 36 layers of 32 heads, with 4 per group. The 32B has 64 layers of 64 heads, with 8 per group. The images, prompts, pairs, splits, patches, and controls are those of Appendix~\ref{app:task}, and the layer bands of the language protocol scale with the depth of the model. The 2B, 4B, and 8B run in 32-bit floating point on one GPU. The 32B does not fit one GPU in 32-bit precision, so we split its weights across two 80 GB GPUs and let its matrix products use TensorFloat-32. This changes the letter scores of an unedited run by less than the $7\times10^{-5}$ of Appendix~\ref{app:task-implementation}, and everything else is unchanged. We do not run the language tasks on the 32B. With no patch, the next token after \texttt{Answer:} is right in 93.4\% of single-answer questions on the 2B, 96.7\% on the 4B, and 98.9\pmv{0.2}\% on the 8B. On the list task, both runs of a pair are right on the second letter in 48.0\% of held-out pairs on the 2B, 92.5\% on the 4B, and 100.0\% on the 32B (8B: 97.4\pmv{1.0}\%). So the 2B's list results rest on fewer cases.

\subsubsection{Binding and Relay}
Table~\ref{tab:sizes-one} repeats the single-answer patches of Section~\ref{sec:binding} at every size. At every size, the key patch, the output patch, and the relay values move the pointer, and restoring the target's values at the relay heads restores the answer. The query patch is weaker below 8B, with 37.5\% on the 2B and 59.0\% on the 4B, and strong on the 32B, with 82.0\%. Two blocking experiments differ by size. On the 2B, the binding heads are necessary: when we block them from the image at the shape token, only 12.0\% of answers are kept. The relay heads are also necessary there, since blocking them keeps only 3.0\%. On the 32B, blocking its 16 relay heads leaves every answer intact. So its relay set is sufficient (84.0\%) but not the only route. Random heads never exceed 1.5\% at any size.

\input{tables/appendix/sizes_one}

\subsubsection{Gaze Heads}
Table~\ref{tab:sizes-gaze} repeats the gaze head tests of Section~\ref{sec:reading}. The query-patch search selects 16 heads on the 2B, where no $K$ passes the rule and the rule takes the best. It selects 128 heads on the 4B, which plateaus from 16 heads on, 32 on the 8B, and 64 on the 32B, which passes at 0.95 on the validation pairs. At every size these heads sit in the last third of the layers. At every size, patching their queries moves the answer to the source's panel, holding their queries blocks a moved pointer, and their image values bring the source's letter. On the 2B, the 16 gaze heads are also necessary: blocking them from the image loses 74.5\% of the answers, while the 4B, 8B, and 32B keep theirs. The published 8B Gaze ranking, applied to the other sizes by head index, moves the answer in 56.0\% of cases on the 4B and 10.0\% on the 2B. So the ranking is a property of the 8B, and each size needs its own search.

\input{tables/appendix/sizes_gaze}

\subsubsection{Progress}
Table~\ref{tab:sizes-progress} repeats the three experiments of Sections~\ref{subsec:progress} to~\ref{subsec:progress-source}, with the residual copied at every layer of each model. The order of Section~\ref{subsec:progress-cursor} reappears at every size, scaled by depth. First comes a ``one letter was written'' phase, then a count, then a cursor after the last written panel, and then the source's letter. On the 32B, the cursor reaches 100\% of pairs at layer 48 of 64, and the shared-panel skip peaks at 99\% at layer 47, which is layer 26 in 8B units. The 32B shows no early skip from the comma-token embedding (10\% or less through layer 40), and it never names the visited panel itself (1\% or less). On the 4B, the cursor reaches 53\% at layer 23 and the count 65.5\% at layer 20, but the visited panel itself is named in 49\% of pairs at layer 22. On the 2B, the visited panel is named in up to 41\% of pairs (layer 18), against a cursor of at most 22\%. So the 8B and 32B hold a clean ``done up to here'' state, while the smaller models hold the count and the visited panel about as often as the position after it. Rebinding the last answer token to panel 4 with each model's own binding heads moves the cursor in 23.5\% of pairs on the 4B, 10.5\% on the 2B, and 23.0\% on the 32B, against 84.2\pmv{3.2}\% on the 8B. Random heads give 0.0\%, and the first answer token gives at most 7.0\% at every size. On the list task, a comma token that cannot attend to the answer token repeats the written letter in 100.0\% of cases on the 32B and in 41.6\% on the 4B, which answers with a letter of another panel in most of the rest. The 2B breaks instead of repeating (6.3\% repeats). All three models keep 81\% or more of their answers when the answer token or the comma token alone is blocked from the image (32B: 90.0\% and 72.0\%). They lose them when both are blocked from the image, keeping 12.5\%, 4.9\%, and 14.0\% on the 2B, 4B, and 32B. So the two routes of Appendix~\ref{app:progress-relay} exist at every size.

\input{tables/appendix/sizes_progress}

\paragraph{The answer-token search on the 32B.} On the 32B, the 16 binding heads alone move progress in only 23.0\% of pairs. We therefore repeat the answer-token search of Appendix~\ref{app:many-mark} on the 32B itself. We start by letting one head at a time read the source's image keys at the answer token, for all 4,096 heads, and rank the heads on the discovery pairs. We then patch the top $K$ heads together on the validation pairs, for every $K$ from 1 to 4,096, and test the chosen set once on held-out pairs. The selection rule of Appendix~\ref{app:one-implementation} selects 64 heads, which is 1.6\% of the heads of the model. Larger sets add little: on the validation pairs, all 4,096 heads move the answer in 76.0\% of cases, against 68.0\% for the 64 heads. The 64 heads sit in layers 33-61, led by L45H11, and 53 of them sit in layers 40-52, around the binding heads in layers 42-48. Of the 16 binding heads, 9 are among the 64. With one letter written, we let these 64 heads read the source's image keys at the answer token. The next answer then moves to the letter after the source's panel in 68.0\% of held-out cases, against 0.0\% for random heads from the same layers (Table~\ref{tab:sizes-answer-token}). The 16 binding heads alone give 22.0\% on the same cases. So the 32B resolves the pointer at the answer token as the 8B does, with the binding heads and the heads around them, but it spreads this over more heads.

\input{tables/appendix/sizes_answer_token}

\subsubsection{Language}
Table~\ref{tab:sizes-language} repeats the colored-shape task on the 2B and 4B. On the 4B, the picture matches the 8B. We find 16 language heads in the same way, and patching their outputs moves the language in 93.0\% of cases, with 2.6\% copied words. Holding MLPs 33-35 leaves 10.5\%. Zeroing the heads returns 54\% of German and French runs to English and no Chinese run, and the language vector switches runs in 32\%, 48\%, and 92\% (German, French, Chinese). The gaze heads' summed output is less separated on the smaller models. Its cosine is 0.88 across languages of one image against 0.71 across images on the 4B, and 0.87 against 0.78 on the 2B, where the 8B gives 0.95 against 0.50. On the 2B, the 16 heads move the language in 23.7\% of cases and copy the source's word in 55.7\%. So concept and language are not yet separate there.

\input{tables/appendix/sizes_language}

\subsection{Other Model Families}
\label{app:model-families}

Every result in the main paper is on Qwen3-VL. Here we ask which parts of the mechanism appear in five other open VLMs, whose language models come from four other families. We run only the tests that need no head search, so that every model is tested the same way and nothing is chosen on it.

\subsubsection{Implementation Details}
The models are LLaVA-OneVision-7B (Qwen2 language model), InternVL3-8B (Qwen2.5), Gemma-3-12B (Gemma 3), Pixtral-12B (Mistral), and Idefics3-8B (Llama 3). We also run Qwen3-VL-8B through the same code as a reference. All models run in 32-bit floating point. Each model sees the images and prompts of Appendix~\ref{app:task} through its own chat template, with the assistant turn prefilled with ``\texttt{Answer:}'', and we find and check the token positions for each model. A patch either replaces the residual vector at one token and one layer with the source's, or blocks attention in the target's own run. The single-answer patches run in both directions, and we keep a pair when both unedited runs are right. We run the list tests only on models that answer the list task correctly in at least 70\% of questions.

Table~\ref{tab:model-families} gives one row per mechanism. We mark a mechanism present ($\surd$) when its effect is above 40\%, and absent ($\times$) when it is at or below 40\%. On the four-panel tests, a model that has lost the pointer answers from a random panel, so 25\% is chance. Row (1) tests whether the colon token holds a pointer: we patch the residual at the colon token as in Section~\ref{sec:binding}, and report the percent of cases that answer from the source's panel at the best layer. Row (2) tests whether the shape token holds the pointer, with the same patch at the shape token. Row (3) tests whether the shape token is read: no later token can attend to the shape token, and we report the percent of answers lost. Row (4) tests whether the comma token skips a written panel, with the design of Section~\ref{subsec:progress}. Row (5) tests the cursor with the design of Section~\ref{subsec:progress-cursor}: we report the first matching panel after the source's last written panel, with the set reading at the same layer in parentheses. Row (6) tests whether the comma token needs the answer token: with one letter written, the last comma token cannot attend to the answer token, and we report the percent that repeat it. Row (7) tests whether select falls back: with nothing written, we hide the first matching panel from every text token at every layer, and report the percent that answer from the second matching panel.

\input{tables/appendix/model_families}

\subsubsection{Results}
Every control is near zero. Patching the token before the colon token into the colon token gives the source's panel in 0.0\% of cases in every model. Patching the token before the shape token into the shape token gives at most chance, because it erases the request and the model then answers from a random panel. A source that has written nothing gives the skip in at most 3\% and the cursor in 0\%. Blocking the attention to the token before the shape token keeps every answer. Blocking the comma token's attention to the colon token gives no repeats. Hiding a non-matching panel moves the answer in at most 1\%.

All six models resolve the pointer at the shape token and use it later. In early layers, the patch at the shape token moves the request in 92.5 to 100\% of cases in every model. Every model loses 69 to 80\% of its answers when later tokens cannot attend to the shape token. In four of the six models, a single residual vector at the colon token or at the shape token carries the pointer, in 65.5 to 84.0\% of cases at the colon token. In Gemma 3 and Idefics3, the single-token patch gives the pointer in only 30.5 to 38.5\% of cases. So in those two models, the pointer is not held in one token's residual in a way the patch can move.

The cursor appears in all four models that can do the list task. The copied comma state sends the target past the source's last written panel in 86 to 99\% of pairs, while the set reading stays at 0 to 2\%. The states also come in the same order along depth as in Qwen3-VL: skip one, then a count, then the cursor, then the source's choice, and then the source's letter. In all four models, the comma token needs the last answer token. When it cannot attend to it, the model writes the same letter again in 56 to 97\% of cases. Hiding the first matching panel moves the answer to the second in three of the four models, and in Pixtral it does so in 21\%. LLaVA-OneVision and Idefics3 do not solve the list task with our prompt (18.5\% and 3.5\%), and LLaVA-OneVision lists the letters of all six panels. So we do not test their list rows.

\subsection{Other Images and Prompts}
\label{app:other-images}

Every image in the main paper is a row of panels, with one prompt wording. Here we ask whether the heads found there do the same job when the image or the prompt changes. In every experiment below, we use the binding, relay, gaze, and language heads \emph{exactly as found} on the panel images, with no new search unless we say so.

\subsubsection{Comic Images With a Longer Prompt}
\label{app:one-comic}
We first ran the single-answer study on a different version of the task, and its held-out results remain valid as a test of robustness. In that version, each panel held a comic drawing under a white header that carried the marker, the letters used all 26 capitals, and the prompt was \texttt{Each panel has one shape marker containing one letter. What is the letter inside the \{shape\} shape?}, with the same \texttt{Answer:} prefix. We ranked the heads, fixed the number of heads, and measured the results on three separate sets of pairs.

The same steps as in Section~\ref{sec:binding} appear. At the colon token, the pointer outcome peaks at 80.2\% at layer 24 (Figure~\ref{fig:comic-sweep}). At the shape token, 8 query heads move the request in 97.5\% of cases, 8 key heads move the layout in 90.0\%, and 32 output heads move the pointer in 85.5\%. The published Gaze Heads follow in 99.0\%, 88.0\%, and 84.5\%. Sixteen relay heads carry the pointer to the colon token. Values from the run with the binding-head patch (Appendix~\ref{app:one-sources}) move it in 85.5\% of cases, and the source's own values in 91.5\%. Restoring the target's values brings back the target's answer in 100.0\%. The first MLP moves the request in 99.0\%. The strongest matched control reaches 0.0\% for any attention patch. The heads selected on the comic images overlap heavily with the heads selected on the blank panels (Table~\ref{tab:selected-heads}). Used unchanged on the blank panels, the comic heads still move each variable: the queries in 67.0\% of cases, the keys in 81.0\%, the outputs in 85.0\%, and the relay values in 87.5\%.

\input{figures/appendix/comic_sweep}

\paragraph{Language heads on comic images.}\label{app:read-comic} We first ran the language experiments with the colored markers drawn over comic strips, and fixed there the numbers of heads used in Section~\ref{subsec:language}. Those held-out results agree with the plain-panel results. Replacing the 16 heads moved the language in 99.2\% of cases, with 0\% copied words. Random and lowest-ranked heads gave 0\%, and holding MLPs 33-35 left 0.3\%. The language vector switched 95\%, 100\%, and 49\% of held-out images to French, Chinese, and German. Zeroing the heads returned English in 100\% of German and French runs, and Chinese runs kept their language there too (Appendix~\ref{app:read-zeroing}).

\subsubsection{Images With No Panels}
\label{app:no-panels}
The panels might do some of the work: they give the model a small number of slots, and a place could simply be a slot number. To test this, we remove them. The shapes sit at random places on one white square canvas with no borders (Figure~\ref{fig:no-panels}), and we run the experiments of Sections~\ref{sec:binding} to~\ref{sec:reading} again, with the head sets of the paper unchanged.

\paragraph{Implementation details.} Single-answer and language images are $768\times768$ pixels with four shapes, and list images are $960\times960$ with six. Both sizes give the same number of image tokens as the matching panel image, or close to it. We draw the centers at random with a minimum distance of 250 pixels, so shapes never touch. The shapes, letters, prompts, pairs, splits, and outcome letters are those of the panel task. A source and its target share the centers, as they share the panels in the main task. For the Gaze count, a place is the box around a shape with a margin of 24 pixels, and image tokens outside every box count for no place. We test the head sets on the held-out sets.

\paragraph{Order of a list.} With no panels, ``left to right'' no longer defines the order of a list. On images with random centers, the model names the right set of letters in 98.3\% of them, but in left-to-right order in only 8.3\%. It reads the image like text: rows from the top, and left to right inside a row. A simple rule reproduces 79.7\% of those lists: take the topmost remaining shape, and everything within 160 pixels below it, as a row. So the list pairs use only layouts on which this rule gives the same order for row heights of 60, 160, and 260 pixels, for the six shapes and for every subset of the four asked places. On such layouts, the rule matches 100.0\% of lists, and $r_1$ to $r_4$ are the places in this order. With these pairs, the unedited runs are as reliable as on panels: 97.0\%, 92.5\%, and 98.0\% of held-out pairs are usable for the next answer, the first answer, and stopping.

\paragraph{Size does not change the order.} A larger shape could stand out more and be named first. To test this, we draw random layouts with the asked shape at four or five of the six places, and give the asked shapes clearly different sizes. Each shape has at least 1.5 times the area of the next smaller one. We assign the sizes at random, so that size does not depend on the place. We draw every layout in four versions: shapes and letters of different sizes, only shapes of different sizes, only letters of different sizes, and one size for all. The model names the right set of letters in 98.0\% to 100.0\% of images. In all four versions, the reading-order rule gives the order of 88.9\% to 90.6\% of pairs of asked shapes. The larger shape comes first in 49.6\% to 49.9\% of these pairs, and the larger letter in 49.4\%, which is chance. Against the same layouts drawn with one size, different sizes change the share of pairs with the larger shape first by only 1.1 to 1.4 points. So on these images, the order of a list follows the places of the shapes, not their sizes.

\input{figures/appendix/no_panels}

\paragraph{The heads carry over unchanged.} Figure~\ref{fig:appendix_no_panels_tests} puts the two kinds of image side by side. Every patch of the single-answer task works with the heads of the paper, and most work better than on panels. The residual patch at the colon token gives the pointer in 93.5\% of cases. The binding heads' outputs move the pointer in 97.5\%, the relay heads in 95.5\%, and the gaze heads' queries in 90.0\%, against 74.0\pmv{2.9}\% on panels. When only the shape token and the gaze heads see the image, the target still answers correctly in 98.0\% of cases. When we keep the gaze heads' queries at the target's values, a patched pointer leaves the answer unchanged in 99.0\%. The list task tells the same story. The gaze heads read the next place at the comma token (86.6\%), and the relay heads deliver the first place (83.2\%). A comma token that cannot attend to the answer token repeats its letter (78.4\%), and a run whose shape token is blocked from the image never stops (99.5\%). Gaze heads pointed at a written place are followed in 0.0\% of cases, and at a place that holds another shape in 37.7\%. So a place is not a panel number. The same heads compute, carry, and read places on a continuous canvas.

\input{figures/appendix/no_panels_tests}

\paragraph{The cursor does not need the layout filter.} The list pairs above use only layouts with a clear reading order. To test whether progress depends on this, we repeat the test of Section~\ref{subsec:progress-source} on canvases, and take the order from the model's own list, not from the rule. The target has written its first two letters. The source is the same image with the last written letter moved to the place of the next answer. We let the 16 binding heads at the last answer token read the source image's keys. On layouts that pass the filter, the next answer then moves one place further in the model's own order in 70.0\% of pairs. On layouts that the filter rejects, where the rule gives different orders at different row heights, it does so in 82.0\%. Random heads from the same layers give 0.0\% and 2.0\%. The same patch at the first answer token leaves 94.0\% and 82.0\% of answers unchanged, and the model repeats a written letter in at most 8.0\% of pairs. So on a canvas the binding heads also set the cursor at the last answer token, with or without the filter.

A new search agrees. A new search for gaze heads on these images selects 16 heads, and all 16 are among the 32 gaze heads of the paper. On the language task, a new search selects 16 heads, and 12 of them are language heads of the paper. Replacing their outputs moves the language in 100.0\% of cases, with 0.0\% copied words (random heads 0.0\%). Holding the late MLPs leaves 8.3\%. Setting the heads to the average of another language switches 77.8\% of runs with 16 heads and 100.0\% with 32. Patching the 32 gaze heads of the paper moves the concept in 71.5\% of cases and the language in 0.0\%.

\paragraph{What the panels give us.} One tool of the paper needs the panels. Appendix~\ref{app:select-sweep} hides one panel from the text tokens and watches the model fall to the next one. On a blank canvas, this does not hide anything. With the box around $r_2$ hidden from every text token in every layer, the model still names the letter of $r_2$ in 100.0\% of cases. When we hide every image token closer to $r_2$ than to any other shape, it still names it in 84.2\%. When nothing separates the shapes, the vision encoder spreads what a shape holds over image tokens far from it. A panel keeps a shape's content inside its own image tokens, and this is what makes a place something we can hide.

\subsubsection{Six Separate Images}
\label{app:multi-image}
A panel could be a region of one image or an image of its own. To test this, we give the model six separate 384 by 384 images in one user turn, each holding one shape and one letter, and ask the paper's list question. Each image becomes 144 image tokens, 864 in all, the same count as one strip. The model lists the letters across the images in order in 97.0\% of items, and exactly in 92.0\%. The cursor experiment of Section~\ref{subsec:progress-cursor}, with images in place of panels, gives the same profile as the strip. The target writes the letter of its sixth image in 83.0\% of pairs at layer 20 and 82.0\% at layer 21. Before that, it holds the count reading (87.0\% at layer 19), and from layer 30 it takes the source's letter. A source that has written nothing leaves the target's own first letter in 100.0\% through layer 22. So the cursor indexes images as it indexes panels. Whether it counts images or positions in the token sequence is not separated here, since the six images have equal size.

\subsubsection{Other Prompt Wordings and Layouts}
\label{app:variants}
We change the prompt or the layout, one at a time, and run the held-out patches of Sections~\ref{sec:binding} to~\ref{sec:reading} with the heads of the paper. We test two wordings. \emph{Prompt A} asks \texttt{Which letter is inside the \{shape\}? Reply with one letter only.} for one answer, and \texttt{What are the letters inside every \{shape\}?} for a list. \emph{Prompt B} asks \texttt{Look at the \{shape\}. What letter is written in it? Answer with the letter only.} and \texttt{List the letter inside each \{shape\}.} We also test two layouts with the paper's prompt. The \emph{grid} puts four panels in a $2\times2$ grid of 768 by 768 pixels and six panels in a $2\times3$ grid of 1,152 by 768, read in row-major order. The \emph{column} stacks the panels top to bottom (384 by 1,536 and 384 by 2,304 pixels). A fifth variant draws the four markers with a white fill and a black outline over a COCO photograph resized to 768 by 768, and defines a place as the 248-pixel box around a marker. We run the single-answer patches on held-out pairs for each variant. We run the list task on the same pairs, counting the cases whose unedited runs are right.

\input{tables/appendix/variants_one}

Table~\ref{tab:variants-one} gives the single-answer patches. The heads of the paper were found on the row images with one wording. Still, they move the pointer on every layout and on photographs in 85\% to 98\% of cases with the outputs, the joint query and key patch, and the relay values, against at most 1.2\% for random heads. The relay heads remain the necessary channel: blocking them keeps only 10\% to 20\% of answers. Prompt B is the weak case. The model's own accuracy drops to 86.7\%, and every patch lands in 48\% to 62\% of cases. The photographs transfer best of all (97\% to 98\%).

\paragraph{A longer question spreads binding over more tokens.} The patches are weakest when the question goes on for two sentences after the shape token (\texttt{Look at the \{shape\}. What letter is written in it? Answer with the letter only.}). In the paper's prompt, only the question mark follows the shape token, and our binding patches act on the shape token alone. So we ask whether the binding heads also act on the tokens that follow it. We patch the heads of the paper at the shape token, at the tokens of the question after it, and at both (Table~\ref{tab:variants-span}). With the two-sentence wording, the outputs of the 16 binding heads at the shape token move the pointer in 52.0\% of cases. At the shape token and the tokens after it, they move the pointer in 87.0\%, against 0.0\% for random heads at the same tokens. The joint query and key patch rises from 55.0\% to 91.0\%. With the paper's prompt, the wider patch adds little, from 84.0\% to 87.0\%. So under a different wording, the same binding heads resolve the pointer, on the shape token and on the tokens after it.

The relay heads then carry this pointer to the colon. We take the relay heads' values from the run with the wider binding patch and install them at the colon token of an unedited target run. The answer then moves to the source's panel in 88.0\% of cases, against 58.0\% with values from the run patched at the shape token alone, and 0.0\% for random heads. Conversely, when we patch the binding outputs but hold the relay heads' values at the target's own, the target keeps its own answer in 97.0\% of cases. With this wording, the relay heads read the pointer from the tokens after the shape token. When we install their values only at these tokens, the answer moves in 67.0\% of cases. When we install them only at the shape token, it moves in 0.0\%, against 47.3\% with the paper's prompt.

\input{tables/appendix/variants_span}

\input{tables/appendix/variants_gaze}

Table~\ref{tab:variants-gaze} gives the gaze heads and the list task. On every variant, patching the gaze heads' queries moves the answer to the source's panel, and holding their queries blocks a moved pointer. Their image values bring the source's letter, and the target still answers correctly when only they and the shape token see the image. Random heads reach at most 3.8\% on these tests. On the lists, the relay values move the first answer, and the gaze heads' queries at the comma token move the next answer. A comma token that cannot attend to the answer token repeats its letter (control 0.0\%), and blocking both the answer token and the comma token from the image loses the answer. The column reads as reliably as a row. With no patch, the model gets the second letter right in 98.5\% of held-out pairs, the first in 99.0\%, and stops in 86.0\%.

\subsubsection{A List From Right to Left}
\label{app:right-to-left}
With no order in the prompt, the model lists the letters from left to right. To test whether this order is fixed, we add \texttt{List them from right to left.} to the end of the list question. The model then lists the letters from right to left. We repeat the answer-token patch of Section~\ref{subsec:progress-source} with this prompt, using the head sets of the main text.

\paragraph{Progress follows the requested order.} With one letter written, we let the 32 heads at the answer token read the source's image keys, where the written letter sits one matching panel further on. The next answer then moves one matching panel further in the requested direction in 66.0\% of pairs, against 73.8\% with the original prompt and 0.0\% for random heads. The model is also less reliable at the list itself when the order is requested: 77.0\% of pairs pass every check with no patch, against 87.9\% with the original prompt. One hypothesis is that the requested order adds a step, carried by the 16 heads below. We leave this as an open question.

\paragraph{The direction comes from the instruction.} We run each image with both prompts and the same letters written. In the right-to-left run, we block the attention from every token after the instruction to the tokens of the instruction. The model then gives the answer of the original prompt in 86.9\% to 100.0\% of cases, for the first answer and with one or two letters written. Blocking the attention to seven other tokens of the question instead gives this answer in at most 5.2\% of cases.

\paragraph{The heads that read the instruction.} We start by blocking the attention of one head at a time to the instruction, with one letter written, for all 1,152 heads. We also score each head when it is the only head that can attend to the instruction. We fix the number of heads on validation images as the smallest set that reaches 90\% of the effect of blocking all heads, and test the set once on held-out images. We find that the top 16 attention heads, in layers 18-23, read the instruction: L18H11, L18H17, L18H19, L19H0, L19H2, L19H12, L19H23, L20H0, L20H1, L20H28, L21H16, L21H17, L21H26, L22H2, L22H9, and L23H24, led by L18H17. When we block these 16 heads from the instruction, the model returns to left to right in 98.4\% of held-out lists. Conversely, when we block every other head from the instruction, the model stays right to left in 98.4\% (Table~\ref{tab:right-to-left}). Random heads from the same layers change the order in at most 1.6\%. So left to right is the order the model uses when the prompt gives none. A small set of heads reads a requested order from the instruction, and progress then follows that order.

\input{tables/appendix/right_to_left}

\subsubsection{Natural Images and Natural Questions}
\label{app:coco-natural}
The list task of the paper has one shape token and one panel per answer. Here the image is a photograph, and the question names a category. We take COCO val2014 with its instance annotations and ask, after the image, \texttt{What animals are in this image? Answer with the names only, separated by a comma and a space.} We ask the same for vehicles, for furniture (chair, couch, bed, dining table), and for objects (all 80 categories, on images with at least three present), with the assistant turn starting \texttt{Answer:}. A category counts if its boxes cover at least 1\% of the image. We resize images to multiples of 32 pixels, so that each image token is a 32 by 32 cell, and take a category's region as the union of its boxes. A case is valid when the model's first two items name two different annotated categories, $A$ then $B$, and the rebuilt prefix \texttt{Answer: A,} gives $B$'s first token again. Of the images, 64.4\% are valid. We lose 26.3\% because the first two items were not two annotated categories, 8.5\% because the list had fewer than two items, 0.7\% because a region was smaller than one token, and 0.1\% because the rebuilt prefix changed the next token. The written answer $A$ is now several tokens, and ``the answer token'' means all of them.

\input{tables/appendix/coco_natural}

Table~\ref{tab:coco-natural} gives the results with the paper's heads. On natural images, the written word carries progress, and the comma token reads it. When the comma token cannot attend to the tokens of $A$, the model names $A$ again in 79.6\% of lists. When it cannot attend to the colon token instead, 98.0\% are unchanged. The written word's own look at the image is not needed, because the comma token finds the object again by itself, as in Appendix~\ref{app:progress-relay}. With $A$'s tokens blocked from their own region, 94.2\% of answers are kept, and with them blocked from the whole image, 85.7\%. With the comma token blocked from the image, 68.6\% are kept, and with $A$ and the comma token both blocked, 62.7\%. The gaze heads at the comma token look at the object named next: their attention per image token is higher on $B$'s region than on $A$'s in 83.6\% of lists, by a median factor of 2.6. The binding heads at $A$'s last token look at $A$ more than at $B$ in 85.5\% of lists, by a median factor of 2.6, against 69.4\% and a factor of 1.4 for random heads from the same layers. The language heads carry the output language here too. With a German question, the first token is German in 44.0\% of runs, and the rest are names spelled the same in both languages, such as \emph{Giraffe} or \emph{Bus}. Setting the 16 language heads to their English averages turns 96.5\% of those runs English, while 16 random heads set the same way leave 97.0\% German. Hiding $B$'s region from every text token does not remove $B$ on a photograph: 70\% of lists are unchanged even when it is hidden from the first layer, for the reason given in Appendix~\ref{app:no-panels}. So we cannot read the layer window of select off natural images with region masking.

\paragraph{Larger objects come first.} On the strips, the model lists from left to right. To see what sets the order on a photograph, we take every pair of categories that a list names, and ask which rule puts the earlier one first. A pair that a rule cannot order, such as two objects of nearly the same size, is left out for that rule. We fix the analysis on one set of lists and run it once on the rest, which we report here. The larger object comes first in 70.8\% of pairs, and the first item of a list is the largest annotated category in 63.2\% of lists, against 40.6\% by chance. The more central object comes first in 62.9\% of pairs. The left object comes first in 53.1\% and the upper one in 49.4\%, which is close to chance. Where size and left to right disagree, the larger object comes first in 68.7\% of pairs. Mirroring the photograph tests position directly. It reverses 13.4\% of pairs, against 3.9\% when we only darken the image slightly. So position plays a small part, and a photograph is not read from left to right. We do not separate size from what goes with it in a photograph, such as the category, the foreground, and how easy an object is to recognize. A score that multiplies size by closeness to the center does no better than size alone (71.4\%), and we did not use a saliency model.

\paragraph{A requested order on photographs.} We add one sentence to the question and compare the new list with the list from the original question. Each pair of named categories either keeps its order or reverses it. With the neutral sentence \texttt{List them in any order.}, 4.3\% of pairs reverse. With \texttt{List them starting with the most salient one.}, 15.2\% reverse, and with \texttt{Start with the most prominent one.}, 16.1\% reverse. With the most prominent one first, the share of pairs with the larger object first rises by 9.9 points over the neutral sentence. With \texttt{List them from left to right.}, 21.0\% of pairs reverse, and the share of pairs in left-to-right order rises by 14.7 points. So the order of a list on a photograph is not fixed. The model changes it when the prompt asks for another order, as it does on the strips (Appendix~\ref{app:right-to-left}).

\subsubsection{A List With No Image}
\label{app:text-only}
The Qwen3-VL language model starts from the Qwen3 text model. To ask whether the mechanism needs the image, we describe the six panels in text (``\texttt{Panel 1: a star with the letter G. Panel 2: a heart with the letter D.}'' and so on), ask the paper's list question with no image, and repeat three tests on Qwen3-VL-8B. The model lists the letters correctly in 89\% of items. When the last comma token cannot attend to the answer token, it repeats the written letter in 72\% of cases (3\% correct), against 99 to 100\% correct when it cannot attend to the colon token. The skip experiment of Section~\ref{subsec:progress} gives 86\% at layer 21 (84\% at 20), against 91.0\pmv{2.4}\% with images, with a no-write control at 0\%. A single-head search at the answer token, ranked by the same key patch on the panel descriptions, selects 16 heads. These heads move the next answer past the source's panel in 61\% of held-out cases, against 0\% for random heads. Of the 16, 7 are binding heads, among them L19H24 and L19H27, two of its top three. So the cursor and the heads that set it do not depend on the image. The text-only Qwen3-8B model lists the letters correctly in only 32\% of items with this prompt, so whether this machinery predates the vision fine-tuning is not settled here.

\subsubsection{Cursor Directions on Other Layouts}
\label{app:cursor-layouts}
Appendix~\ref{app:cursor-subspace} writes the cursor on strips, with one direction per panel. The strip directions do not carry over to a 2 by 3 grid or to shapes scattered on a blank canvas (at or below the random null), and only partly to a column. Directions fitted on images of each layout do work there: grid 98.8\%, column 95.0\%, and scatter 81.3\% for $p=3$, against 36.2 to 63.8\% for random directions. So every layout has a cursor, and each has its own code (cosines with the strip directions from 0.18 to 0.93).

%% file: tables/appendix/sizes_one.tex
\begin{table}[ht]
\centering
\small
\caption{\textbf{The binding and relay heads move the pointer at every size.} We use each model's own heads, found with the rule of Appendix~\ref{app:one-implementation}. Each cell is the percent of cases whose next token is the expected letter. The blocking rows give the percent of cases that keep the target's own answer. We use held-out pairs.}
\label{tab:sizes-one}
\fittable{%
\begin{tabular}{@{}lcccc@{}}
\toprule
Patch & 2B & 4B & 8B & 32B \\
\midrule
Binding queries: source's request & 37.5\% & 59.0\% & 81.0\pmv{2.5}\% & 82.0\% \\
Binding image keys: source's layout & 71.0\% & 82.0\% & 83.1\pmv{2.3}\% & 94.0\% \\
Queries and keys: source's panel & 84.0\% & 96.0\% & 95.0\pmv{1.5}\% & 99.0\% \\
Binding outputs: source's panel & 86.0\% & 93.5\% & 85.9\pmv{2.4}\% & 87.0\% \\
Relay values from the run with the binding-head patch: source's panel & 84.5\% & 95.5\% & 89.0\pmv{2.1}\% & 84.0\% \\
\quad target's values restored: own answer & 91.0\% & 98.0\% & 99.8\pmv{0.2}\% & 100.0\% \\
Random heads, any patch & $\le$1.0\% & $\le$1.5\% & $\le$0.1\pmv{0.1}\% & 0.0\% \\
\midrule
Nothing blocked: own answer & 93.0\% & 97.0\% & 100.0\pmv{0.0}\% & 100.0\% \\
Binding heads blocked from the image at the shape token: own kept & 12.0\% & 57.5\% & 84.9\pmv{2.1}\% & 100.0\% \\
Later tokens cannot attend to the shape token: own kept & 10.5\% & 24.5\% & 25.3\pmv{2.5}\% & 31.0\% \\
Relay heads cannot attend to the text at the colon token: own kept & 3.0\% & 28.0\% & 7.8\pmv{1.6}\% & 100.0\% \\
\quad random heads: own kept & 93.5\% & 97.0\% & 99.8\pmv{0.2}\% & 100.0\% \\
\bottomrule
\end{tabular}}
\end{table}

%% file: tables/appendix/sizes_gaze.tex
\begin{table}[ht]
\centering
\small
\caption{\textbf{The gaze heads read the pointer at every size.} We use the heads found by the query-patch search on each model. Each cell is a percent of cases on held-out pairs.}
\label{tab:sizes-gaze}
\fittable{%
\begin{tabular}{@{}lcccc@{}}
\toprule
Test & 2B (16 heads) & 4B (128 heads) & 8B (32 heads) & 32B (64 heads) \\
\midrule
Queries patched: source's panel & 63.5\% & 75.5\% & 74.0\pmv{2.9}\% & 90.0\% \\
Pointer moved at the shape token, queries held: own kept & 82.0\% & 92.0\% & 98.4\pmv{0.8}\% & 100.0\% \\
Source's image values: source's letter & 44.5\% & 63.0\% & 93.7\pmv{1.6}\% & 97.5\% \\
Only these heads and the shape token see the image: own kept & 84.5\% & 75.5\% & 95.1\pmv{1.3}\% & 95.0\% \\
These heads blocked from the image at the colon token: own kept & 25.5\% & 95.5\% & 88.2\pmv{2.0}\% & 100.0\% \\
Random heads, queries patched & 0.0\% & 0.0\% & 0.0\pmv{0.0}\% & 0.0\% \\
\bottomrule
\end{tabular}}
\end{table}

%% file: tables/appendix/sizes_progress.tex
\begin{table}[ht]
\centering
\small
\caption{\textbf{Progress at every size.} We repeat the experiments of Section~\ref{sec:progress-select} with the residual copied at every layer of each model, and the rebind with each model's own binding heads. Each cell is a percent of pairs, with the layer of the peak in parentheses. We use held-out pairs.}
\label{tab:sizes-progress}
\fittable{%
\begin{tabular}{@{}lcccc@{}}
\toprule
Reading of the copied comma state & 2B (28 layers) & 4B (36 layers) & 8B (36 layers) & 32B (64 layers) \\
\midrule
Skip experiment: skips the shared first matching panel & 78 (17) & 74 (20) & 91.0\pmv{2.4} (20) & 99 (47) \\
Cursor experiment: ``one written'', skip one & 87 to 99 (3 to 7) & 97 (7) & 92.4 to 98.4 (0 to 7) & 75 to 91 (34 to 40) \\
Cursor experiment: count of two & 100 (8 to 13) & 65.5 (20) & 90.8\pmv{2.5} (15) & 93 (42) \\
Cursor experiment: the visited panel itself & 41 (18) & 49 (22) & $\le$1.6\pmv{1.1} (20 to 24) & $\le$1 \\
Cursor experiment: first matching panel after the visited panel & 22 (20) & 53 (23) & 98.4\pmv{1.1} (20) & 100 (48) \\
Source's letter, from layer & 22 & 30 & 30 & 53 \\
\midrule
Rebind, last answer token, binding heads: past panel 4 & 10.5 & 23.5 & 84.2\pmv{3.2} & 23.0 \\
\quad all heads & 13.5 & 23.5 & 62.2\pmv{4.4} & 39.0 \\
\quad random heads & 0.0 & 0.0 & 0.0\pmv{0.0} & 0.0 \\
Rebind, first answer token, binding heads / all heads & 0.0 / -- & 0.0 / 7.0 & 1.2\pmv{0.9} / 2.5 & 0.0 / 5.0 \\
\bottomrule
\end{tabular}}
\end{table}

%% file: tables/appendix/sizes_answer_token.tex
\begin{table}[ht]
\centering
\small
\caption{\textbf{The answer-token heads of the 32B.} Each row lets a set of heads at the last answer token read the source's image keys, in a list with one letter written. Percent of held-out lists in which the next answer moves to the letter after the source's panel.}
\label{tab:sizes-answer-token}
\begin{tabular}{@{}lc@{}}
\toprule
Heads that read the source's image keys & Next answer follows the source \\
\midrule
The 64 heads found on the 32B & 68.0\% \\
64 random heads from the same layers & 0.0\% \\
The top 256 heads & 68.0\% \\
All 4,096 heads & 68.0\% \\
The 16 binding heads & 22.0\% \\
The 64 heads without the binding heads & 0.0\% \\
\bottomrule
\end{tabular}
\end{table}

%% file: tables/appendix/sizes_language.tex
\begin{table}[ht]
\centering
\small
\caption{\textbf{The colored-shape language task on the 2B, 4B, and 8B.} We use held-out images, with cross-language cases for the head patches.}
\label{tab:sizes-language}
\fittable{%
\begin{tabular}{@{}lccc@{}}
\toprule
Experiment & 2B & 4B & 8B \\
\midrule
Answers in the question's language, no patch & 97.5\% & 98.2\% & 100.0\pmv{0.0}\% \\
Gaze heads' summed output: cosine across languages / across images & 0.87 / 0.78 & 0.88 / 0.71 & 0.95 / 0.50 \\
16 language heads replaced from the source: own concept, source's language & 23.7\% & 93.0\% & 95.3\pmv{0.5}\% \\
\quad copied source words & 55.7\% & 2.6\% & 0.0\pmv{0.0}\% \\
\quad MLPs of the last three layers held & 0.4\% & 10.5\% & 0.9\pmv{0.2}\% \\
Heads zeroed: German and French runs to English & 76\% & 54\% & 94.0\pmv{1.5}\% \\
Heads zeroed: Chinese runs kept Chinese & 7\% & 100\% & 100.0\pmv{0.0}\% \\
Language vector: to German / French / Chinese & 75 / 75 / 100\% & 32 / 48 / 92\% & 75.0\pmv{3.8} / 100.0\pmv{0.0} / 100.0\pmv{0.0}\% \\
\bottomrule
\end{tabular}}
\end{table}

%% file: tables/appendix/model_families.tex
\begin{table}[ht]
\centering
\small
\caption{\textbf{Most parts of the mechanism appear across model families.} Each cell is a percent of cases, with the layer of the peak and the number of layers. $\surd$ marks an effect above 40\%, and $\times$ an effect at or below 40\%. ``Not solvable'' means the model answers the list task correctly in under 70\% of questions. We rerun Qwen3-VL-8B with the same code as the other models.}
\label{tab:model-families}
\fittable{%
\begin{tabular}{@{}lcccccc@{}}
\toprule
 & Qwen3-VL-8B & LLaVA-OV-7B & InternVL3-8B & Gemma-3-12B & Pixtral-12B & Idefics3-8B \\
Language model & Qwen3 & Qwen2 & Qwen2.5 & Gemma 3 & Mistral & Llama 3 \\
No patch: single answer / list & 98.5 / 96.5 & 99.5 / 18.5 & 99.5 / 80.0 & 100.0 / 99.0 & 100.0 / 98.0 & 88.5 / 3.5 \\
\midrule
(1) Pointer at the colon token & 80.0 (24/36) $\surd$ & 84.0 (22/28) $\surd$ & 65.5 (22/28) $\surd$ & 30.5 (34/48) $\times$ & 68.0 (25/40) $\surd$ & 38.5 (17/32) $\times$ \\
(2) Pointer at the shape token & 42.5 (20/36) $\surd$ & 69.5 (17/28) $\surd$ & 68.0 (17/28) $\surd$ & 32.0 (20/48) $\times$ & 79.0 (16/40) $\surd$ & 31.0 (12/32) $\times$ \\
(3) Shape token is read: answers lost & 77 $\surd$ & 70 $\surd$ & 80 $\surd$ & 73 $\surd$ & 75 $\surd$ & 69 $\surd$ \\
(4) Comma token skips a written panel & 96 (20/36) $\surd$ & not solvable & 79 (19/28) $\surd$ & 96 (23/48) $\surd$ & 87 (18/40) $\surd$ & not solvable \\
(5) Cursor (set reading) & 99 (0) (21/36) $\surd$ & not solvable & 94 (0) (19/28) $\surd$ & 86 (0) (26/48) $\surd$ & 89 (2) (18/40) $\surd$ & not solvable \\
(6) Comma token needs the answer token: repeats & 82 $\surd$ & not solvable & 56 $\surd$ & 97 $\surd$ & 90 $\surd$ & not solvable \\
(7) Select falls back to the next matching panel & 48 $\surd$ & not solvable & 70 $\surd$ & 88 $\surd$ & 21 $\times$ & not solvable \\
\bottomrule
\end{tabular}}
\end{table}

%% file: figures/appendix/comic_sweep.tex
\begin{figure*}[t]
\centering
\includegraphics[width=\textwidth]{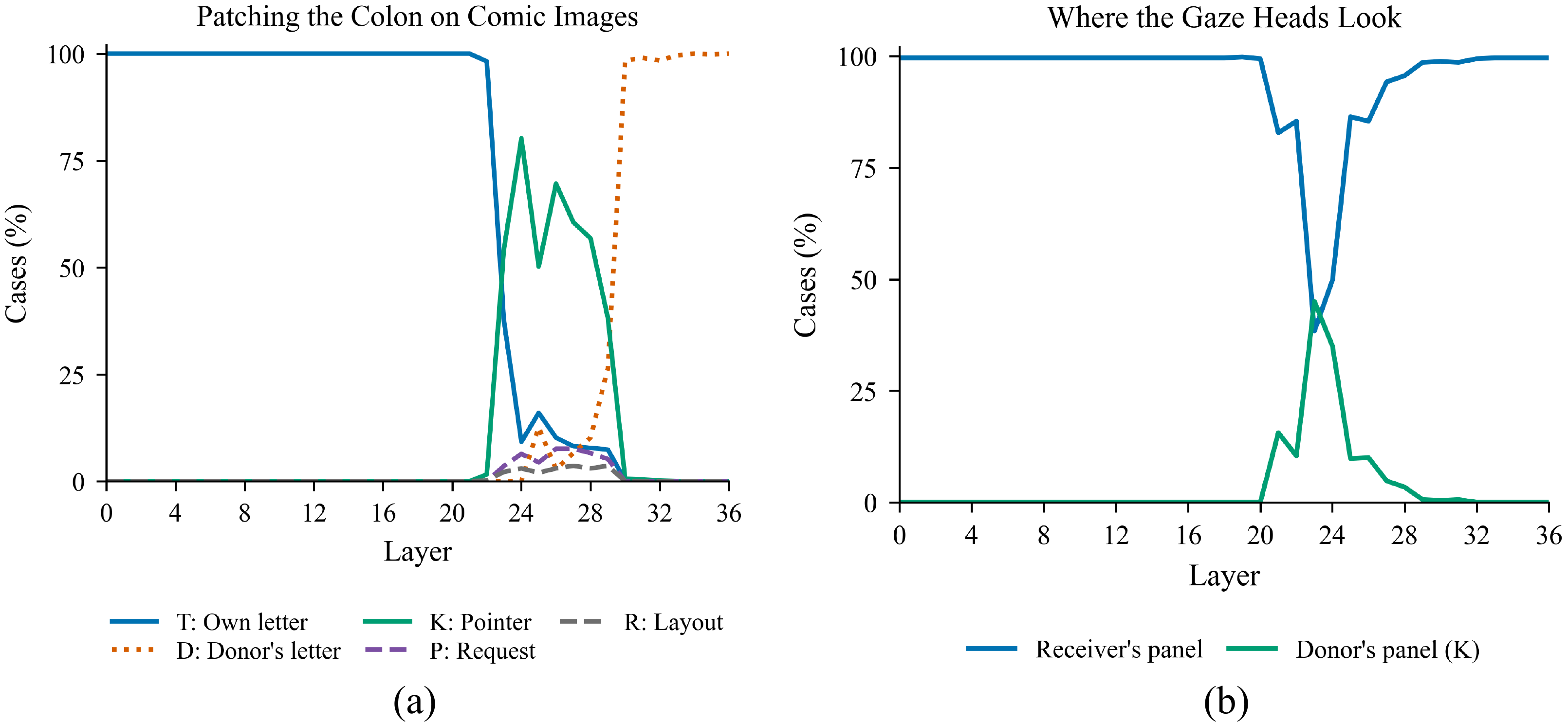}
\caption{\textbf{On comic images, the colon token holds the pointer at the same layers.} We replace the target's residual vector at the colon token with the source's at one layer, as in Figure~\ref{fig:binding}a, on cases with comic drawings under the markers and the longer prompt. (a)~The target's next letter. The pointer, \texttt{K}, peaks at 80.2\% at layer 24, and the source's letter, \texttt{D}, comes through from layer 30. (b)~The panel the published Gaze Heads look at, with no head patched. At layer 23 they look at the source's panel in 45.0\% of cases.}
\label{fig:comic-sweep}
\end{figure*}

%% file: figures/appendix/no_panels.tex
\begin{figure}[ht]
\centering
\includegraphics[width=0.34\textwidth]{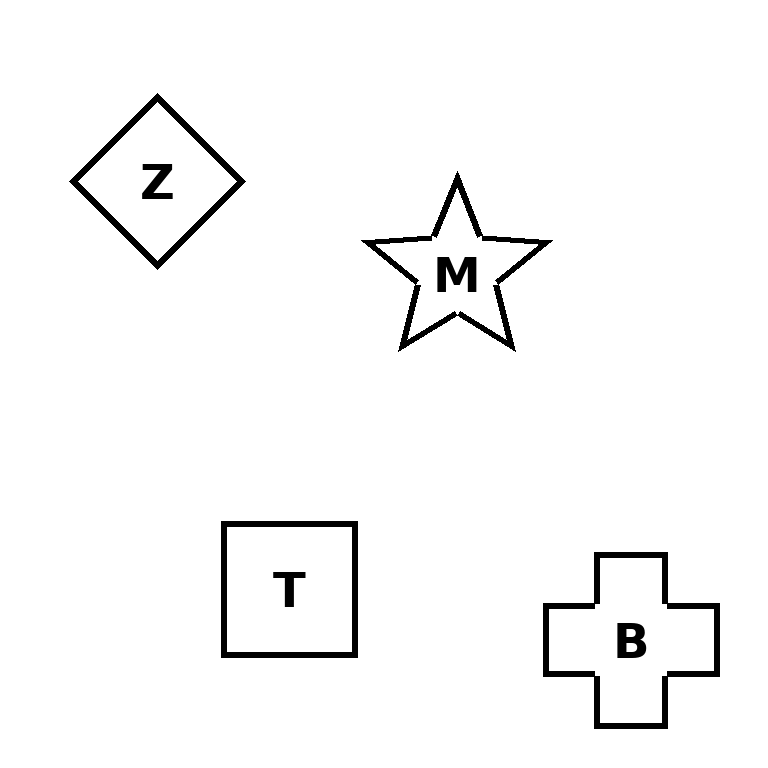}\hspace{0.05\textwidth}
\includegraphics[width=0.34\textwidth]{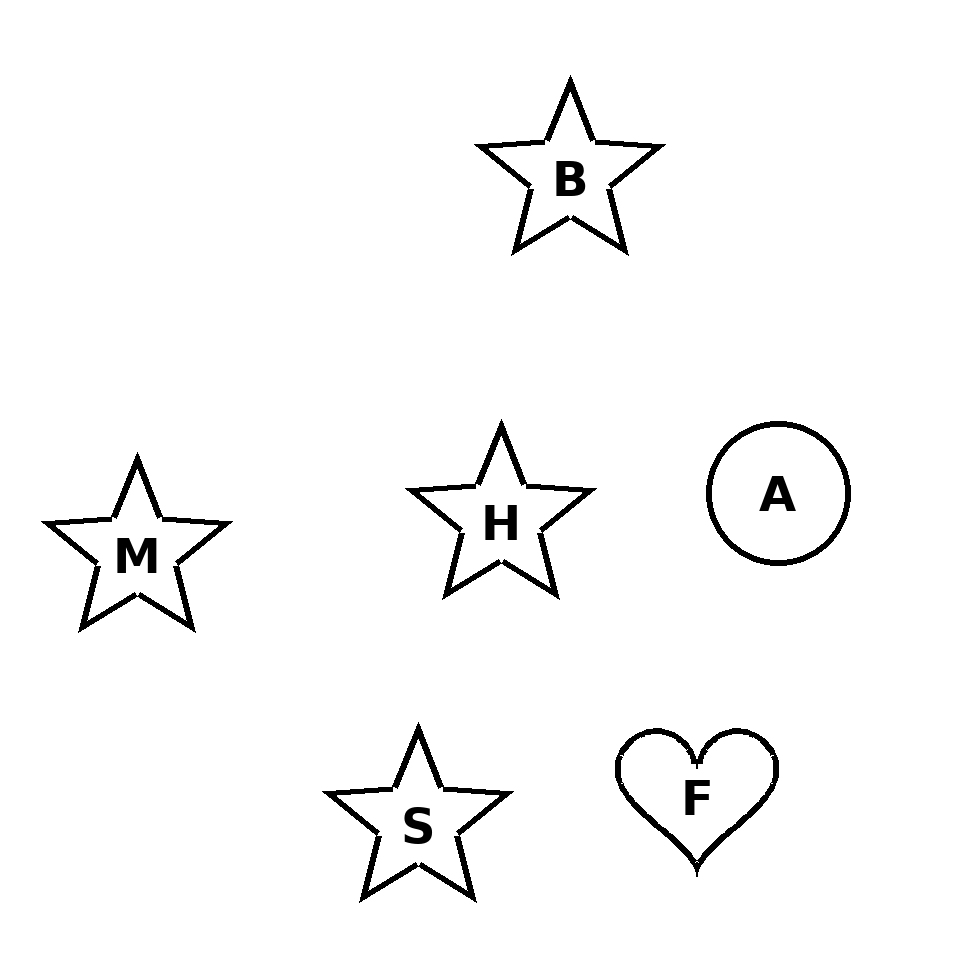}
\caption{\textbf{Images with no panels.} Left: a single-answer image. Right: a list image in which the asked shape, a star, sits at four places.}
\label{fig:no-panels}
\end{figure}

%% file: figures/appendix/no_panels_tests.tex
\begin{figure*}[t]
\centering
\includegraphics[width=\textwidth]{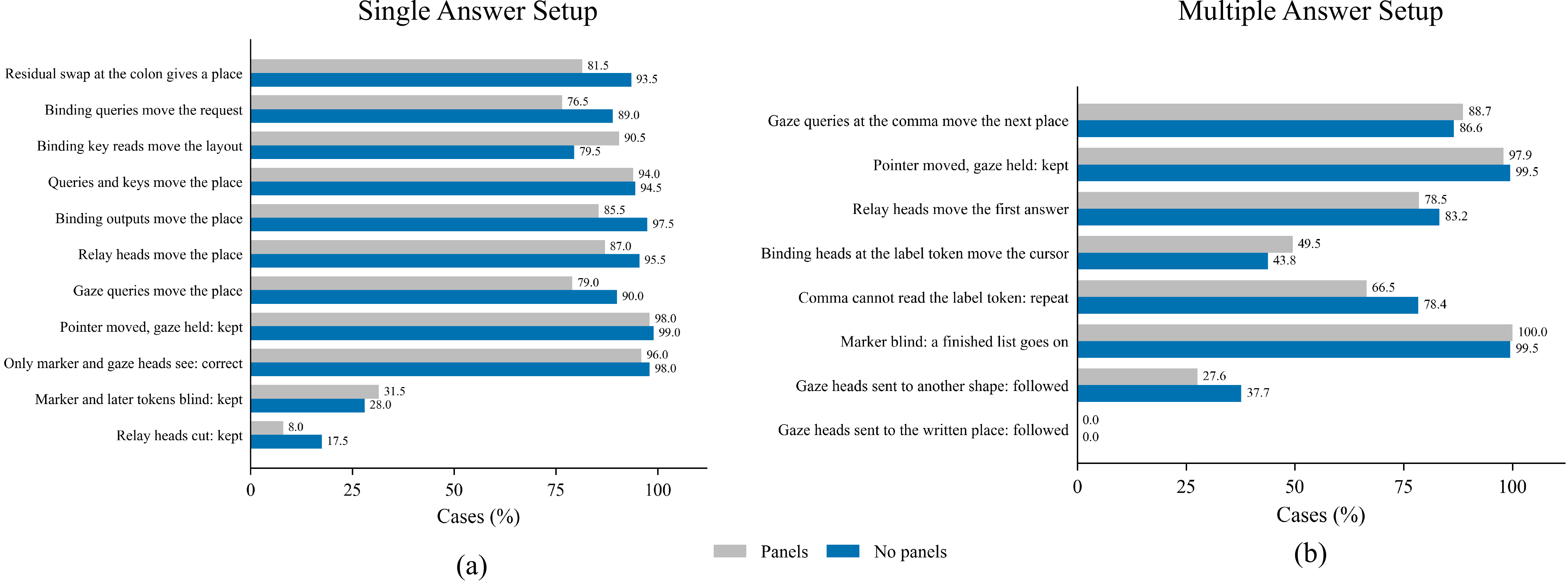}
\caption{\textbf{The same heads work on images with no panels.} Every bar uses the head sets selected on panel images, with no new search. (a)~The single-answer tests, on the held-out pairs with panels (gray) and with no panels (blue). (b)~The list tests, on the held-out list cases whose unedited runs are right. Random heads from the same layers move the answer in 0.0\% of single-answer and list cases.}
\label{fig:appendix_no_panels_tests}
\end{figure*}

%% file: tables/appendix/variants_one.tex
\begin{table}[ht]
\centering
\small
\caption{\textbf{The paper's heads move the pointer on other prompts and layouts.} Each cell is the percent of cases whose next token is the expected letter. The blocking rows give the percent of cases that keep the target's own answer. The first column is the paper's held-out set on the row images.}
\label{tab:variants-one}
\fittable{%
\begin{tabular}{@{}lcccccc@{}}
\toprule
Patch & Row, paper & Prompt A & Prompt B & Grid & Column & Photographs \\
\midrule
Binding queries: source's request & 81.0\pmv{2.5}\% & 73.3\% & 48.3\% & 67.5\% & 72.5\% & 90.8\% \\
Binding image keys: source's layout & 83.1\pmv{2.3}\% & 83.3\% & 61.7\% & 75.0\% & 75.0\% & 98.3\% \\
Queries and keys: source's panel & 95.0\pmv{1.5}\% & 90.0\% & 56.7\% & 93.8\% & 92.5\% & 96.7\% \\
Binding outputs: source's panel & 85.9\pmv{2.4}\% & 80.0\% & 53.3\% & 93.8\% & 85.0\% & 96.7\% \\
Relay values from the run with the binding-head patch: source's panel & 89.0\pmv{2.1}\% & 79.2\% & 58.3\% & 93.8\% & 86.2\% & 98.3\% \\
\quad target's values restored: own answer & 99.8\pmv{0.2}\% & 99.2\% & 93.3\% & 98.8\% & 96.2\% & 100.0\% \\
Random heads, any patch & 0.1\pmv{0.1}\% & 0.0\% & 0.0\% & 1.2\% & 1.2\% & 0.0\% \\
\midrule
Nothing blocked: own answer & 100.0\pmv{0.0}\% & 99.2\% & 86.7\% & 97.5\% & 97.5\% & 100.0\% \\
Relay heads cannot attend to the text at the colon token: own kept & 7.8\pmv{1.6}\% & 9.2\% & 10.0\% & 15.0\% & 21.2\% & 20.0\% \\
Later tokens cannot attend to the shape token: own kept & 25.3\pmv{2.5}\% & 30.8\% & 25.0\% & 25.0\% & 23.8\% & 18.3\% \\
Binding heads blocked from the image at the shape token: own kept & 84.9\pmv{2.1}\% & 89.2\% & 93.3\% & 77.5\% & 82.5\% & 78.3\% \\
Gaze heads blocked from the image at the colon token: own kept & 88.2\pmv{2.0}\% & 97.5\% & 91.7\% & 96.2\% & 95.0\% & 100.0\% \\
\bottomrule
\end{tabular}}
\end{table}

%% file: tables/appendix/variants_span.tex
\begin{table}[ht]
\centering
\small
\caption{\textbf{A longer question spreads binding over more tokens.} Each cell is the percent of cases whose next token is the letter at the source's panel, except the last row, which gives the percent of cases that keep the target's own answer. We patch the heads of the paper at the shape token, at the tokens of the question after it, or at both. In the relay rows, the relay heads' values come from a run with the binding outputs patched at the tokens named.}
\label{tab:variants-span}
\fittable{%
\begin{tabular}{@{}llccc@{}}
\toprule
 & & Paper's & One sentence after & Two sentences after \\
Patch & Tokens patched & prompt & the shape token & the shape token \\
\midrule
Binding outputs & The shape token & 84.0\% & 79.0\% & 52.0\% \\
 & The tokens after it & 4.0\% & 8.0\% & 26.0\% \\
 & Both & 87.0\% & 88.0\% & 87.0\% \\
 & Both, random heads & 1.0\% & 0.0\% & 0.0\% \\
\midrule
Queries and keys & The shape token & 93.0\% & 88.0\% & 55.0\% \\
 & The tokens after it & 2.0\% & 2.0\% & 11.0\% \\
 & Both & 97.0\% & 98.0\% & 91.0\% \\
 & Both, random heads & 1.0\% & 0.0\% & 0.0\% \\
\midrule
Relay values & The shape token & 86.0\% & 80.0\% & 58.0\% \\
 & Both & 86.0\% & 88.0\% & 88.0\% \\
 & Both, random heads & 0.0\% & 0.0\% & 0.0\% \\
 & Both, target's values restored: own answer & 99.0\% & 99.0\% & 97.0\% \\
\bottomrule
\end{tabular}}
\end{table}

%% file: tables/appendix/variants_gaze.tex
\begin{table}[ht]
\centering
\small
\caption{\textbf{The gaze heads and the list task carry over to other prompts and layouts.} Each cell is a percent of cases, with the paper's heads. The top rows test the 32 gaze heads. The bottom rows test the list task, on the cases whose unedited runs are right. A dash marks a test we did not run on that variant.}
\label{tab:variants-gaze}
\fittable{%
\begin{tabular}{@{}lccccc@{}}
\toprule
Test & Row, paper & Prompt A & Prompt B & Grid & Column \\
\midrule
Queries patched: source's panel & 74.0\pmv{2.9}\% & 83.3\% & 76.7\% & 82.5\% & 73.8\% \\
Pointer moved, gaze queries held: own kept & 98.4\pmv{0.8}\% & 98.3\% & 93.3\% & 97.5\% & 96.2\% \\
Source's image values: source's letter & 93.7\pmv{1.6}\% & 94.2\% & 95.0\% & 90.0\% & 90.0\% \\
Only these heads see the image: own kept & 95.1\pmv{1.3}\% & 90.0\% & 91.7\% & 96.2\% & 96.2\% \\
Random heads: queries patched / only these see & 0.0\pmv{0.0}\% / 4.8\pmv{1.2}\% & 0.0\% / 3.3\% & 0.0\% / 1.7\% & 1.2\% / 3.8\% & 0.0\% / 3.8\% \\
\midrule
First answer: binding outputs at the shape token: source's first panel & 53.4\% & 58.6\% & -- & 71.1\% & 51.3\% \\
First answer: relay values at the colon token: source's first panel & 78.5\% & 93.1\% & -- & 97.4\% & 76.9\% \\
First answer: gaze queries at the colon token: source's first panel & 46.1\% & 53.4\% & -- & 60.5\% & 51.3\% \\
Second answer: gaze queries at the comma token: source's next panel & 88.2\pmv{2.8}\% & -- & 70.0\% & -- & 82.1\% \\
Random heads, every patch & $\le$1.0\% & 0.0\% & $\le$3.3\% & 0.0\% & $\le$2.6\% \\
Comma token cannot attend to the answer token: repeats & 67.8\pmv{4.0}\% & 72.4\% & 83.3\% & 72.7\% & 82.1\% \\
\quad control: comma token cannot attend to the colon token: repeats & 0.0\pmv{0.0}\% & 0.0\% & 0.0\% & 0.0\% & 0.0\% \\
Relay heads cannot attend to the text at the comma token: not own & 34.4\pmv{4.0}\% & 41.4\% & 43.3\% & 30.3\% & 46.2\% \\
Answer token and comma token both blocked from the image: not own & 81.8\pmv{3.4}\% & 62.1\% & 70.0\% & 39.4\% & 56.4\% \\
\bottomrule
\end{tabular}}
\end{table}

%% file: tables/appendix/right_to_left.tex
\begin{table}[ht]
\centering
\small
\caption{\textbf{A small set of heads reads the order instruction.} Each row blocks the attention of a set of heads to the instruction \texttt{List them from right to left.}, in a list with one letter written. Each cell is the percent of held-out lists in which the model then gives the next answer of the left-to-right order.}
\label{tab:right-to-left}
\begin{tabular}{@{}lc@{}}
\toprule
Heads blocked from the instruction & Left-to-right answer \\
\midrule
None & 0.0\% \\
All 1,152 heads & 100.0\% \\
The 16 heads found here & 98.4\% \\
All heads except these 16 & 0.0\% \\
16 random heads from the same layers & $\le$1.6\% \\
The 16 heads, blocked from seven other question tokens & 0.0\% \\
\bottomrule
\end{tabular}
\end{table}

%% file: tables/appendix/coco_natural.tex
\begin{table}[ht]
\centering
\small
\caption{\textbf{On natural images, the comma token reads progress from the written word.} We use the paper's heads on the valid lists of COCO val2014. $A$ is the first item the model wrote, and $B$ the second. Each cell is the percent of valid lists in that group.}
\label{tab:coco-natural}
\fittable{%
\begin{tabular}{@{}lccccc@{}}
\toprule
Test & Animals & Vehicles & Furniture & Objects & All \\
\midrule
Comma token cannot attend to $A$'s tokens: names $A$ again & 89.2\% & 70.6\% & 82.2\% & 75.2\% & 79.6\% \\
\quad control: comma token cannot attend to the colon token: unchanged & 99.5\% & 99.5\% & 95.8\% & 97.1\% & 98.0\% \\
$A$'s tokens blocked from $A$'s region: unchanged & 98.6\% & 95.9\% & 93.9\% & 88.1\% & 94.2\% \\
$A$'s tokens blocked from the image: unchanged & 93.7\% & 91.8\% & 89.7\% & 67.6\% & 85.7\% \\
Comma token blocked from the image: unchanged & 73.0\% & 70.1\% & 91.1\% & 39.5\% & 68.6\% \\
$A$'s tokens and the comma token both blocked from the image: unchanged & 69.8\% & 67.5\% & 82.2\% & 31.0\% & 62.7\% \\
Gaze heads at the comma token: more attention per token on $B$ than on $A$ & 95.5\% & 86.1\% & 76.6\% & 75.7\% & 83.6\% \\
Binding heads at $A$: more attention per token on $A$ than on $B$ & 77.9\% & 80.9\% & 89.3\% & 93.8\% & 85.5\% \\
\quad control: random heads from the same layers & 53.2\% & 61.9\% & 76.2\% & 86.7\% & 69.4\% \\
\bottomrule
\end{tabular}}
\end{table}